# A modified RIME algorithm with covariance learning and diversity enhancement for numerical optimization

Shangqing Shi [1, +], Luoxiao Zhang [2, +], Yuchen Yin [3], Xiong Yang [4,*] and Hoileong Lee [5]

[1] School of Information Science and Engineering, Southeast University, Nanjing 210096, China;
[2] Faculty of Natural, Mathematical & Engineering Sciences, Department of Engineering, King's College of London, 57 Waterloo Road London, SE1 8WA, British;
[3] Teachers College, Columbia University, 525 West 120th Street, New York, NY 10027, USA;
[4] Zhicheng College, Fuzhou University, Fuzhou, 350002, China;
[5] Faculty of Electronic Engineering & Technology, Universiti Malaysia Perlis, 02600 Arau, Perlis, Malaysia;
* Correspondence: 02116828@fdzcxy.edu.cn
+ These authors contributed equally to this work

**Abstract:** Metaheuristics are widely applied for their ability to provide more efficient solutions. The RIME algorithm is a recently proposed physical-based metaheuristic algorithm with certain advantages. However, it suffers from rapid loss of population diversity during optimization and is prone to fall into local optima, leading to unbalanced exploitation and exploration. To address the shortcomings of RIME, this paper proposes a modified RIME with covariance learning and diversity enhancement (MRIME-CD). The algorithm applies three strategies to improve the optimization capability. First, a covariance learning strategy is introduced in the soft-rime search stage to increase the population diversity and balance the over-exploitation ability of RIME through the bootstrapping effect of dominant populations. Second, in order to moderate the tendency of RIME population to approach the optimal individual in the early search stage, an average bootstrapping strategy is introduced into the hard-rime puncture mechanism, which guides the population search through the weighted position of the dominant populations, thus enhancing the global search ability of RIME in the early stage. Finally, a new stagnation indicator is proposed, and a stochastic covariance learning strategy is used to update the stagnant individuals in the population when the algorithm gets stagnant, thus enhancing the ability to jump out of the local optimal solution. The proposed MRIME-CD algorithm is subjected to a series of validations on the CEC2017 test set, the CEC2022 test set, and the experimental results are analyzed using the Friedman test, the Wilcoxon rank sum test, and the Kruskal Wallis test. The results show that MRIME-CD can effectively improve the performance of basic RIME and has obvious superiorities in terms of solution accuracy, convergence speed and stability.

## 1. Introduction

The rapid development of artificial intelligence technology complements the continuous progress of society. On the one hand, the real world emerges more complex needs after continuous development, which span many fields such as natural sciences, medicine, engineering, economics, etc., presenting unprecedented complexity and challenge [1]. On the other hand, science and technology have emerged after continuous innovation with various artificial intelligence technologies, which provide powerful tools for solving complex real-world problems. Artificial intelligence technology, with its powerful data processing ability, learning ability and reasoning ability, shows great potential for application in various fields [2, 3]. Optimization problems are an important research direction in a wide range of scientific fields and engineering areas. Its main goal is to find the best global solution for problems with certain constraints or limited resources. In practice, many optimization problems usually involve nonlinear, nonconvex and large-scale dimensions with high complexity and uncertainty. In general, for practical optimization problems, traditional deterministic methods and metaheuristic algorithms are two widely used approaches [4]. Since traditional deterministic methods are very time consuming and energy intensive when tackling complex problems with a large number of constraints and variables, especially in real-life engineering applications, they are prone to converge to locally optimal solutions [5]. With the increase in computational power and the continuous development of optimization techniques, a category of stochastic methods known as metaheuristic algorithms have emerged, which have attracted much attention due to their simple structure, high local optimum avoidance ability, and no need for gradient information requirement [6, 7]. Meta-heuristic algorithms have been developed over a long period of time and different kinds of branches have been formed. Xie and Xue et al. classified meta-heuristic algorithms into three categories: evolution-based algorithms, physics-based algorithms, and swarm-based algorithms [8, 9]. There is also a category of human-based meta-heuristic algorithms according to Jia and Abualigah et al [10, 11]. In this paper, meta-heuristic algorithms are categorized into five classes: evolutionary based, swarm based, human based, mathematical based and physical based algorithms.

Evolutionary-based algorithms primarily emulate biological phenomena like Darwin's evolution, genetic inheritance, and mating. Some of the well-known algorithms in this category include Genetic Algorithm (GA) [12], Differential Evolution (DE) [13], Evolutionary Strategies (ES) [14], Alpha evolution (AE) [15] and Evolutionary Mating Algorithm (EMA) [16]. Swarm-based algorithms simulate the collective social behavior of foraging, reproduction, and avoidance of natural enemies in groups of organisms. Particle Swarm Optimization (PSO) [17] is one of the classical instances in which the foraging behavioral patterns of bird flocks are simulated to locate the ideal solution. Other prominent algorithms in this category comprise Ant Colony Optimization (ACO) [18], Grey Wolf Optimizer (GWO) [19], Eurasian Lynx Optimizer (ELO) [20], Dwarf Mongoose Optimization (DMO) [21], Draco Lizard Optimizer (DLO) [22], Slime Mould Algorithm (SMA) [23], Artificial Meerkat Algorithm (AMA) [24], Crayfish Optimization Algorithm (COA) [25], Genghis Khan Shark Optimizer (GKSO) [26], Prairie Dog Optimization Algorithm (PDOA) [27] and Gazelle Optimization Algorithm (GOA) [28]. The third group involved human-based algorithms that are based on human interaction and cooperation within a community. Teaching Learning Based Optimization (TLBO) [29], Football Team Training Algorithm (FTTA) [30], Political Optimizer (PO) [31], Catch Fish Optimization Algorithm (CFOA) [32], Poor and Rich Optimization Algorithm (PROA) [33], Educational Competition Optimizer (ECO) [34] and Hybrid Leader Based Optimization (HLBO) [35] are some of the widely employed algorithms. The fourth category comprises mathematical based algorithms. Algorithms based on mathematics are inspired by mathematical theories, functions, and formulas, which have demonstrated significant promise in enhancing the computing efficacy of optimization approaches. One of the noteworthy methods in this category is the Sine Cosine Algorithm (SCA) [36], which applies the concept of trigonometric functions to create an algorithmic model. Arithmetic Optimization Algorithm (AOA) [37], Quadratic Interpolation Optimization (QIO) [38], Tangent Search Algorithm (TSA) [39], Triangulation Topology Aggregation Optimizer (TTAO) [40] and Sinh Cosh optimizer (SCO) [41] are other instances in this category. The final group encompasses physical based algorithms that mimic the principles of physical phenomena in natural. One of an exceptional physical based approaches is Simulate Annealing (SA) [42], which is influenced by the metallurgical annealing principle. Other meta-heuristic algorithms in this category are Gravitational Search Algorithm (GSA) [43], Snow Ablation Optimizer (SAO) [44], PID-based search algorithm (PSA) [45], Equilibrium Optimizer (EO) [46], Multi-Verse Optimizer (MVO) [47] and Polar Lights Optimization (PLO) [48]. These meta-heuristic algorithms have been widely used in various fields such as image segmentation [49], task planning [50], structural optimization [51], engineering design problem [52, 53], energy system optimization [54], hyperparameter optimization [55], fault diagnosis [56], feature selection [57] and portfolio optimization [58], which is considered as an emerging field in artificial intelligence.

As a new physical based meta-heuristic algorithm, RIME algorithm [15], was proposed by Su et al. in 2023. The representation model of the algorithm is inspired from the growth process of ice fog. The algorithm mathematically models two different growth mechanisms of freezing fog. The RIME algorithm has been successfully applied to optimization problems in different domains.

Although the RIME algorithm has the advantages of few control parameters and fast convergence speed, it faces the shortcomings of weakening population diversity, low convergence accuracy, and easy to fall into local optimum when solving complex optimization problems. In recent years, many researchers have conducted studies around improving the performance and application of RIME. Although these developments have improved the efficiency of RIME, the limited performance enhancement of RIME when faced with complex optimization problems has motivated us to seek to further improve the performance of RIME. Furthermore, according to the No Free Lunch (NFL) theorem, no meta-heuristic technique can be guaranteed to be appropriate and feasible for every application.

In view of the above motivation, this paper proposes a RIME variant combining covariance learning and diversity enhancement techniques called MRIME-CD. First, a covariance learning strategy is introduced in the soft-rime search stage to increase the population diversity and balance the over-exploitation ability of RIME through the bootstrapping effect of dominant populations. Second, in order to moderate the tendency of RIME population to approach the optimal individual in the early search stage, an average bootstrapping strategy is introduced into the hard-rime puncture mechanism, which guides the population search through the weighted position of the dominant populations, thus enhancing the global search ability of RIME in the early stage. Finally, a new stagnation indicator is proposed, and a stochastic covariance learning strategy is used to update the stagnant individuals in the population when the algorithm gets stagnant, thus enhancing the ability to jump out of the local optimal solution. The key contribution of this study is briefly stated as below.

(1) We proposed Gaussian-based covariance learning strategies, average bootstrapping strategy and stochastic covariance learning-based population diversity mechanism to improve the performance of basic RIME.

(2) We examined the efficiency of the proposed MRIME-CD on the 29 CEC2017 and 12 CEC2022 benchmark test functions and compared the obtained results with other comparative algorithms to ensure the superiority and stability in solving complex optimization problems.

(3) We applied various statistical tests including the Friedman test, the Wilcoxon rank sum test, and the Kruskal Wallis test to analyze the differences between MRIME-CD and other competitors.

The rest of the paper is organized in the following way: Section 2 presents a review of the principles and mathematical framework of the basic RIME algorithm. Section 3 clarifies the core concepts of the proposed MRIME-CD approach. In addition, pseudo-code and flowcharts are provided and their time complexity is discussed in depth. Section 4 demonstrates the optimization performance of MRIME-CD on the CEC2017 and CEC2022 test sets while also discussing the impact of each improvement strategy on MRIME-CD. Section 5 summarizes in detail the conclusions of this study and the scope of future research. Figure 1 provides a complete outline of this paper.

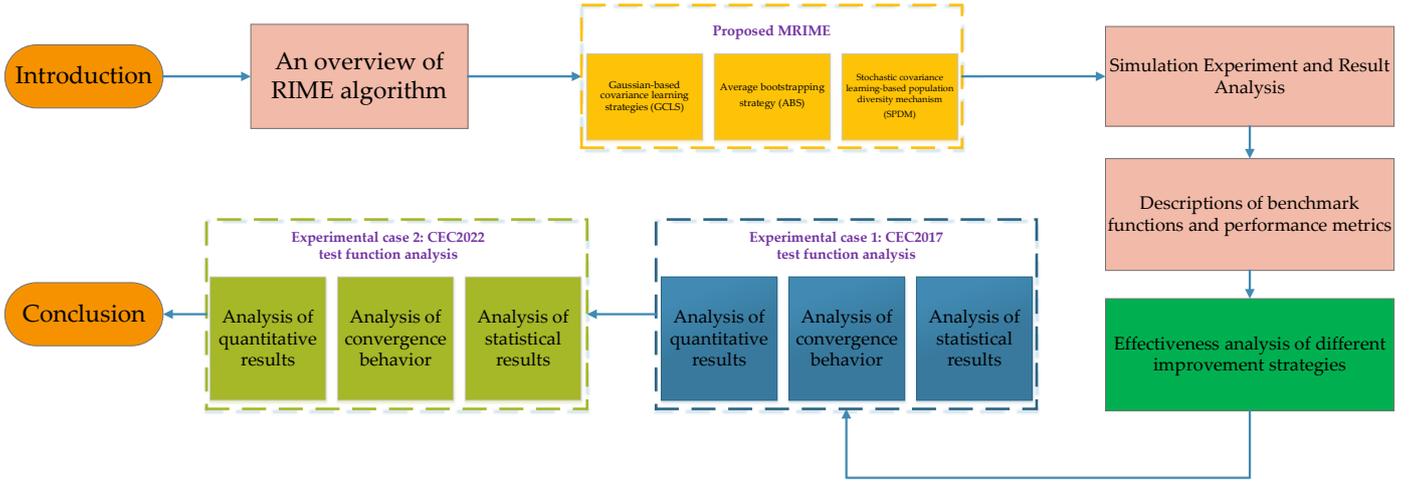

**Figure 1.** A complete outline of this work.

## 2. An overview of RIME algorithm

This section will provide a detailed explanation of the basic RIME algorithm.

### 2.1. Population initialization

In RIME, each rime ice agent represents a proposed solution to a problem. Mathematically, each population member is a vector $X_i$. These vectors together form the population matrix of the RIME algorithm $X$. Like other metaheuristic algorithms, the initial population of the RIME algorithm is generated by random initialization in the search space. Eq. (1) denotes a rime agent and Eq. (2) denotes the rime population.

$$X_i = [x_{i,1}, x_{i,2}, \ldots, x_{i,D}] = LB + rand \times (UB - LB), i = 1, 2, \ldots, NP \tag{1}$$

$$X = \begin{bmatrix} x_{1,1} & x_{1,2} & \cdots & x_{1,D} \\ x_{2,1} & x_{2,2} & \cdots & x_{2,D} \\ \vdots & \vdots & \ddots & \vdots \\ x_{NP,1} & x_{NP,2} & \cdots & x_{NP,D} \end{bmatrix}_{NP \times D} \tag{2}$$

where $D$ denotes the number of problem variables. $LB$ and $UB$ denote the upper and lower boundaries of the problem. $NP$ represents the number of population members. $x_{NP,D}$ is the value of the $D^{th}$ variable specified by the $NP^{th}$ proposed solution. As mentioned earlier, each ice agent is a proposed solution to the problem. Therefore, the objective function of the problem can be evaluated based on each member. These values obtained for the objective function can be represented as vectors using Eq. (3).

$$F(X) = \begin{bmatrix} Fit_1 = F(X_1) \\ \vdots \\ Fit_i = F(X_i) \\ \vdots \\ Fit_{NP} = F(X_{NP}) \end{bmatrix}_{NP \times 1} \tag{3}$$

where $F$ denotes the vector of objective function values and $Fit_i$ denotes the objective function value of the $i^{th}$ candidate solution.

### 2.2. Soft-rime search strategy

The RIME algorithm performs a comprehensive sweep of the problem space using the soft-rime search strategy in the exploration phase. This search strategy simulates the random growth of soft rime ice under breezy wind environment, which can cover a wide area. In other words, the soft-rime search strategy enables the searching individuals to search the problem space extensively for potential solutions. The concept expressed in the soft-rime search strategy is mathematically modeled using Eq. (4) to Eq. (7).

$$X_{i,j}^{new} = X_{best,j} + \alpha \times \cos(\theta) \times \beta \times \left(rand \times (UB_{i,j} - LB_{i,j}) + LB_{i,j}\right), r_1 < E \tag{4}$$

$$\theta = \frac{FEs \times \pi}{10 \times FEs_{max}} \tag{5}$$

$$\beta = 1 - round(w \times FEs / FEs_{max}) / w \tag{6}$$

$$E = \sqrt{FEs / FEs_{max}} \tag{7}$$

where $X_{i,j}^{new}$ represents the $j^{th}$ dimensional position of the $i^{th}$ agent after updating. $X_{best,j}$ represents the $j^{th}$ dimensional position of the optimal agent. $\alpha$ is a randomly generated value that matches the range between -1 and 1. $FEs$ and $FEs_{max}$ represent the current evaluation number and the maximum number of evaluations, respectively. $w$ is a constant, which is set to 5 in this paper. $round(\cdot)$ denotes the rounding operation.

2.3. Hard-rime puncture mechanism

The hard-rime puncture mechanism of RIME enhances the ability of rime agents to converge to the optimal agent. This mechanism simulates the formation process of hard-rime ice under strong wind conditions. Eq. (8) summarizes this puncture mechanism.

$$X_{i,j}^{new} = X_{best,j}, rand < nrom(Fit_i) \tag{8}$$

where $nrom(Fit_i)$ denotes the normalized value of the current rime agent fitness value.

2.4. Positive greedy selection mechanism

RIME selects offspring individuals after each updating by a positive greedy selection mechanism. By comparing the fitness of parent and offspring individuals, RIME ensures that the population moves to better agents and accelerates search convergence.

## 3. Proposed MRIME

RIME achieves better convergence through both search mechanisms. However, RIME faces the challenges of insufficient population diversity and ease of falling into local optimum traps in the later search phase. To cope with these challenges, this paper proposes the MRIME-CD algorithm, which contains three improved techniques: Gaussian-based covariance learning strategies (GCLS), average bootstrapping strategy (ABS) and stochastic covariance learning-based population diversity mechanism (SPDM). In this section, we first elaborate the framework of the proposed MRIME-CD algorithm. Then the time complexity of the MRIME-CD algorithm proposed in this paper is analyzed, and finally the pseudo-code and flowchart are provided.

3.1. Gaussian-based covariance learning strategy (GCLS)

RIME updates each individual around the optimal rime particle in the soft-rime search phase. This approach guarantees fast convergence and is extremely efficient when dealing with simple problems. However, as time progresses, the challenges we face are sophisticated, which leads to the inability of RIME to efficiently solve complex optimization problems when faced with them. This is due to over exploitation that weakens the global exploration capability. When the population is stuck in a local optimum, RIME lacks the ability to get rid of it. In view of this, this paper proposes a Gaussian-based covariance learning strategy, which is a method that relies on the evolutionary trend of the population. The strategy is divided into two parts, sampling and generation. GCLS uses maximum likelihood estimation methods on the dominant population to model the Gaussian distribution. The specific equations are as follows.

$$X_{mean} = \sum_{i=1}^{|S|} \omega_i \times X_i \tag{9}$$

$$\omega_i = \ln(|S|+1) / \left(\sum_{i=1}^{|S|} (\ln(|S|+1) - \ln(i))\right) \tag{10}$$

$$C = \frac{1}{|S|} \sum_{i=1}^{|S|} (X_i - X_{mean}) \times (X_i - X_{mean})^T \tag{11}$$

where $S$ denotes the ensemble of dominant groups and $|S|$ denotes the number of dominant groups. $\omega_i$ is the weight coefficient and $C$ is the weighted covariance matrix. By introducing weighting coefficients, it is ensured that the better individuals have a greater impact on population evolution while retaining effective information from more individuals. For GCLS strategy, choosing a suitable dominant group is essential to perform its performance. In this paper, we propose a roulette domain selection mechanism to construct the dominant group. The mechanism selects the optimal individual $X_r$ through the adaptive distance balance mechanism based on roulette, and then calculates the Euclidean distance between $X_r$ and the other individuals in the group, and selects the part of the individuals with closer distance to form the dominant population. The benefits of this mechanism are twofold. One is that the roulette domain selection mechanism ensures that the more superior individuals can be selected as $X_r$, and at the same time can select the rest of the individuals to expand the search range. The second is that the use of distance rather than fitness as a selection metric enables the learning of valid information around $X_r$. Since individuals close to $X_r$ are not necessarily well-adapted, this can enrich the diversity of the dominant group. In addition, as the optimization proceeds, the number of dominant populations retained by multiple iterations keeps increasing, which can lead to some old individuals being overused and stagnate the evolution, and thus it is necessary to discard them. In this paper, we adopt the first-in-first-out method, when the number of dominant populations is too large, the newly selected individuals will replace the earliest retained dominant individuals. After the sampling is completed, new candidate solutions are generated according to the method shown in Eq. (12).

$$X_i^{new} = Gaussian(X_{mean}, C) + rand \times (X_{mean} - X_i) \tag{12}$$

The coordinated transition between global exploration and local exploitation has important role in leveraging the performance of RIME. Therefore, in this paper, we use Gaussian-based covariance learning strategy in the early stage, and gradually transition to soft-rime search strategy in the later stage. This switching is judged using Eq. (13).

$$X_i^{new} = \begin{cases} X_{best} + \alpha \times \cos(\theta) \times \beta \times (rand \times (UB - LB) + LB), r_1 < E \\ Gaussian(X_{mean}, C) + rand \times (X_{mean} - X_i), r_1 \geq E \end{cases} \tag{13}$$

3.2. Average bootstrapping strategy (ABS)

RIME achieved a speed-up convergence using the hard-rime puncture mechanism. However, this direct movement of a certain dimension to the location of the optimal individual greatly diminishes the population diversity. This tends to cause the algorithm to concentrate on the locally optimal individual at a later stage without escape. Therefore, this paper proposes an average bootstrapping strategy. Unlike the hard-rime puncture mechanism, the ABS mechanism allows rime particles to refer to the weighted average position of the optimal agent and the dominant population at the same time in the early stage. This can promote each individual to explore in more directions while ensuring the convergence efficiency. Similar to Eq. (13), the ABS mechanism works in the pre-search phase, as in Eq. (14).

$$X_{i,j}^{new} = \begin{cases} X_{best,j}, rand < nrom(Fit_i) \cup r_1 < E \\ (X_{best,j} + X_{mean,j})/2, rand < nrom(Fit_i) \cup r_1 \geq E \end{cases} \tag{14}$$

3.3. Stochastic covariance learning-based population diversity mechanism (SPDM)

As the search progresses, the population inevitably clusters around an optimal individual or multiple localized optimal individuals, which leads to a decrease in population diversity. When the population diversity is seriously insufficient, the search of the population will be stagnant and unable to move to other promising areas. Therefore, it is necessary to determine whether the algorithm is stagnant or not and take measures to avoid stagnation. In this paper, we propose a diversity metric that considers population distribution and fitness changes to determine whether the algorithm is stagnant or not. The method consists of two key computations: one is to compute the boundaries of the search space, and the other is to compute the spatial distribution of the population during the iteration process. The diversity indicator is represented as follows.

$$V_{\lim} = \sqrt{\prod_{j=1}^{D} |UB_j - LB_j|} \tag{15}$$

$$V_{pop} = \sqrt{\prod_{j=1}^{D} |(UB_{x_j} - LB_{x_j})/2|} \tag{16}$$

$$nVOL = \sqrt{V_{pop} / V_{\lim}} \tag{17}$$

where $UB_{x_j}$ and $LB_{x_j}$ are the upper and bound of $j^{th}$ dimension of the current rime population. On the other hand, we introduce a $Count$ to record whether each individual gets better at each updating. If the $i^{th}$ offspring is better than the corresponding parent individual, Count accumulates 1. When $nVOL < 0.01$ and $Count > 2 \times D$ are satisfied, MRIME-CD recognizes that it is stagnant at this point, and the stochastic covariance learning strategy is executed to update the stagnant individuals, as shown below.

$$X_i^{new} = Gaussian(X_{mean}, C) + rand \times (X_{r1} - X_i) + rand \times (X_{r2} - X_i) \tag{18}$$

where $X_{r1}$ and $X_{r2}$ are two different individuals randomly selected from the population.

### 3.4. The framework of MRIME-CD algorithm

The proposed MRIME-CD is obtained by combining the basic RIME with the three improved techniques described above. To elucidate the proposed MRIME-CD, pseudocode is provided in Algorithm 1. Furthermore, Figure 2 presents a flowchart for the detailed description of the MRIME-CD.

---

**Algorithm 1.** Pseudocode of MRIME-CD

Initialize $NP$, $D$, $FEs_{max}$

Create an initial population generated by using Eq.(1)

**While** $FEs < FEs_{max}$

Calculate the fitness value of each rime individual

Calculate coefficient $E$ using Eq.(7)

Calculate $C$, $X_{mean}$ using Eq.(9) and Eq.(10) **%% GCLS**

**For each rime agent**

**If** $r_1 < E$

    Compute the rime agent's position using Eq.(4)

**Else**

    Compute the rime agent's position using Eq.(11) **%% GCLS**

**End If**

**If** $rand < F_{norm}$

    Compute the rime agent's position using Eq.(13) **%% ABS**

**End If**

**If** $F(R_i^{new}) < F(R_i)$

    Update the rime agent's position using greedy strategy

**Else**

    $Count_i = Count_i + 1$

**End If**

Calculate $nVOL$ using Eq.(16) **%% SPDM**

**If** $nVOL < 0.1\ \&\&\ Count_i > 2 \times D$

Compute the rime agent's position using Eq.(16) **%% SPDM**

**End**

**End for**

$FEs = FEs + NP$

**End While**

---

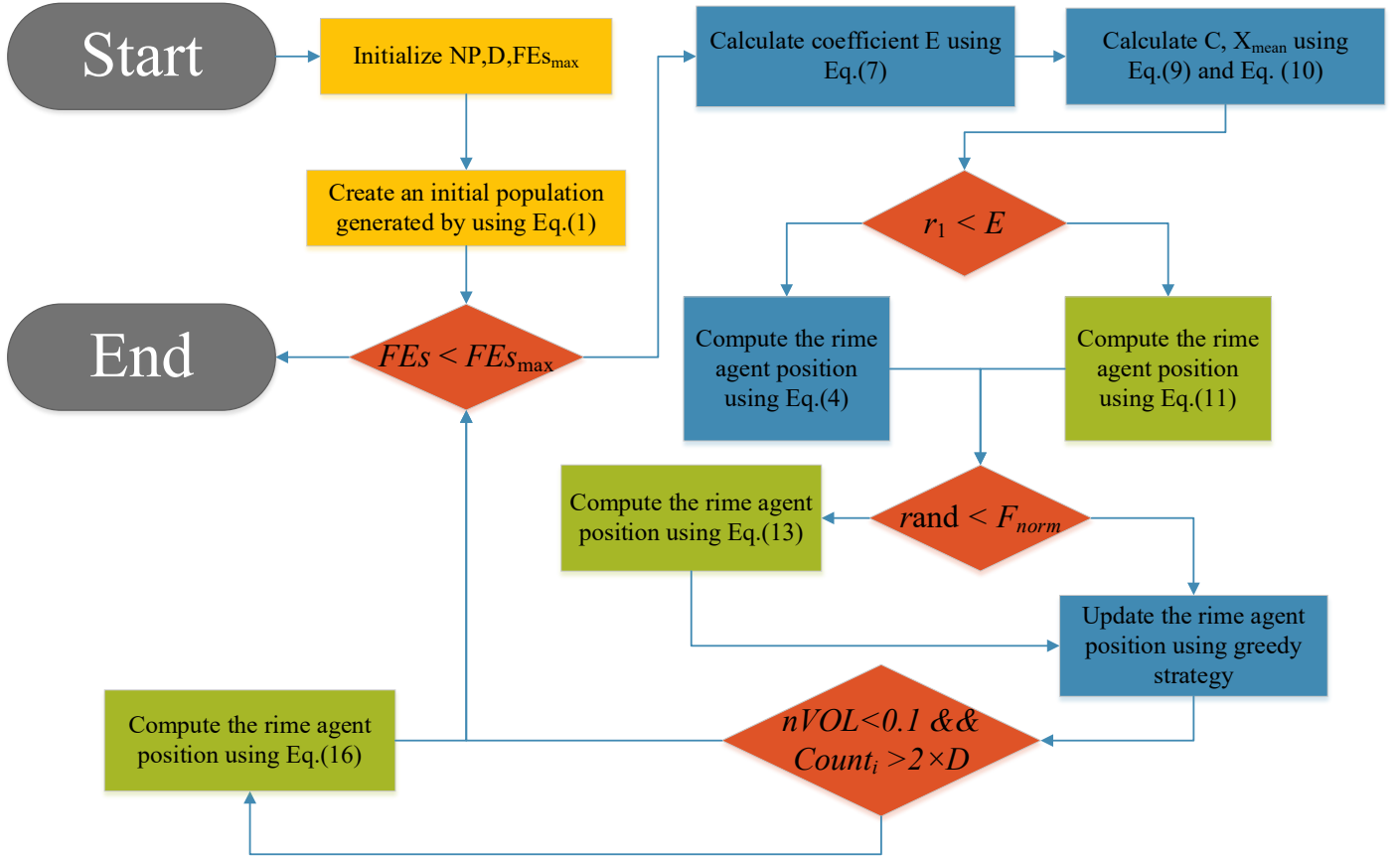

**Figure 2.** The flowchart of the proposed MRIME-CD algorithm

Complexity analysis is a measure of the system as a whole and evaluates the efficiency of the algorithm. We set the number of populations as $NP$, the maximum number of iterations as $T$, and the problems number of variables as $D$. The complexity of the basic RIME is $O(T \times NP \times D)$. As for MRIME-CD, at each updating, each rime particle randomly selects one from GCLS and soft-rime search strategy to execute, and one from ABS and hard-rime puncture mechanism to execute. Therefore, these two strategies do not increase the time complexity of the algorithm. We assume that $N_{re}$ agents use SPDM per iteration, then this strategy increases the time complexity of $O(T \times N_{re} \times D)$. Therefore, the time complexity of the proposed MRIME-CD is $O(T \times (NP + N_{re}) \times D)$. In this paper, we use the maximum number of evaluations as the algorithm stopping criterion, so that MRIME-CD and RIME can be compared under fair conditions.

The space complexity is mainly related to population size $NP$ and problem dimension $D$. The space complexity of RIME is $O(NP \times D)$. According to the pseudo-code and flowchart of MRIME-CD, it can be seen that GCLS and ABS do not introduce additional computation of population fitness and therefore do not increase the space complexity of the algorithm. SPDM makes the size of the population to $(NP + N_{re})$, and the dimension of the problem does not change. Therefore, the space complexity of SPDM is $O((NP + N_{re}) \times D)$. Since the order of magnitude does not change, the space complexity of MRIME-CD remains $O(NP \times D)$, which is the same as the basic RIME.

## 4. Simulation Experiment and Result Analysis

In order to evaluate the performance of the proposed MRIME-CD, this section presents a detailed report about its performance on 29 CEC2017 test functions and 12 CEC2022 test functions. To ensure a fair comparison and considering the stochastic character of the algorithms, all algorithms will be solved independently 51 times for each function, with a maximum number of evaluations of 3000xDim. All the experiments in this paper are conducted on the same platform, with the platform specifications shown in Table 1. In addition, all the competitors for comparison will set their respective parameters according to the original literature, as shown in Table 2.

**Table 1.** Platform specifications utilized for experiments

| Hardware | |
|---|---|
| CPU | AMD R9 9950X (4.3GHz) |
| RAM | 64GB |
| Operating system | 64-bit Windows 11 |
| **Software** | |
| Programming language | Matlab R2022a |

**Table 2.** Parameters setting of MRIME-CD and other competing algorithms.

| **Common setting** | |
|---|---|
| Maximum number of function evaluations | $FEs_{max} = 3000 \times Dim$ |
| Dimension of the Problem | $Dim = 10/30/50/100 (CEC2017)$ |
|  | $Dim = 10/20 (CEC2022)$ |
| Number of independent runs | 51times |
| **Algorithm setting** | |
| MRIME-CD | $w = 5$ |
| RIME | $w = 5$ |
| EO | $a_1 = 2, a_2 = 1, GP = 0.5$ |
| SAO | $k = 1$ |
| ACGRIME | $a = 4$ |
| IRIME | $F_1 = 1, F_2 = 0.8, F_3 = 1, Cr_1 = 0.1, Cr_2 = 0.2, Cr_3 = 0.9, w = 5$ |
| TERIME | $w_{max} = 0.3, w_{min} = 0.0625$ |
| EOSMA | $V = 1, a_1 = 2, a_2 = 1, GP = 0.5, z = 0.3, q = 0.2$ |
| RDGMVO | $W_{max} = 1, W_{min} = 0.2, p = 0.6$ |
| MTVSCA | $\lambda = 0.25, c = 0.7, a = 2$ |

4.1. Descriptions of benchmark functions and performance metrics

In general, the effectiveness of an algorithm is determined by various factors. One of the most important considerations is the results obtained by applying the proposed methodology to different classes of problems. In this regard, this section aims to evaluate the feasibility, applicability and reliability of the proposed MRIME-CD algorithm.

Therefore, we use two benchmark test suites from the IEEE Congress on Evolutionary Computation to explore MRIME-CD comprehensively. The CEC2017 test suite contains unimodal functions (F1-F2), multimodal functions (F3-F9), hybrid functions (F10-F19), and composite functions (F20-F29). The CEC2022 test suite contains unimodal functions (F1), basic functions (F2-F5), hybrid functions (F6-F8) and composite functions (F9-F12). These functions are typically considered nonlinear, non-derivable, non-convex, and extremely complex. A broad summary of these test functions is presented in Table 3 and Table 4.

**Table 3.** Detailed description of CEC2017 test functions

| Type | ID | Functions name | fmin |
|---|---|---|---|
| Unimodal functions | F1 | Shifted and Rotated Bent Cigar Function | 100 |
|  | F2 | Shifted and Rotated Zakharov Function | 300 |
| Multimodal functions | F3 | Shifted and Rotated Rosenbrock's Function | 400 |
|  | F4 | Shifted and Rotated Rastrigin's Function | 500 |
|  | F5 | Shifted and Rotated Expanded Scaffer's F6 Function | 600 |
|  | F6 | Shifted and Rotated Lunacek Bi_Rastrigin Function | 700 |
|  | F7 | Shifted and Rotated Non-Continuous Rastrigin's Function | 800 |
|  | F8 | Shifted and Rotated Levy Function | 900 |
|  | F9 | Shifted and Rotated Schwefel's Function | 1000 |
| Hybrid functions | F10 | Hybrid Function 1 (N=3) | 1100 |
|  | F11 | Hybrid Function 2 (N=3) | 1200 |

| | F12 | Hybrid Function 3 (N=3) | 1300 |
|---|---|---|---|
| | F13 | Hybrid Function 4 (N=4) | 1400 |
| | F14 | Hybrid Function 5 (N=4) | 1500 |
| | F15 | Hybrid Function 6 (N=4) | 1600 |
| | F16 | Hybrid Function 6 (N=5) | 1700 |
| | F17 | Hybrid Function 6 (N=5) | 1800 |
| | F18 | Hybrid Function 6 (N=5) | 1900 |
| | F19 | Hybrid Function 6 (N=6) | 2000 |
| | F20 | Composition Function 1 (N=3) | 2100 |
| | F21 | Composition Function 2 (N=3) | 2200 |
| | F22 | Composition Function 3 (N=4) | 2300 |
| | F23 | Composition Function 4 (N=4) | 2400 |
| Composition functions | F24 | Composition Function 5 (N=5) | 2500 |
| | F25 | Composition Function 6 (N=5) | 2600 |
| | F26 | Composition Function 7 (N=6) | 2700 |
| | F27 | Composition Function 8 (N=6) | 2800 |
| | F28 | Composition Function 9 (N=3) | 2900 |
| | F29 | Composition Function 10 (N=3) | 3000 |

Search Range: $[-100, 100]^D$

**Table 4.** Detailed description of CEC2022 test functions

| Type | ID | Functions name | fmin |
|---|---|---|---|
| Unimodal functions | F1 | Shifted and full Rotated Zakharov Function | 300 |
| | F2 | Shifted and full Rotated Rosenbrock's Function | 400 |
| Basic functions | F3 | Shifted and full Rotated Expanded Schaffer's f6 Function | 600 |
| | F4 | Shifted and full Rotated Non-Continuous Rastrigin's Function | 800 |
| | F5 | Shifted and full Rotated Levy Function | 900 |
| | F6 | Hybrid Function 1 (N = 3) | 1800 |
| Hybrid functions | F7 | Hybrid Function 2 (N = 6) | 2000 |
| | F8 | Hybrid Function 3 (N = 5) | 2200 |
| | F9 | Composition Function 1 (N = 5) | 2300 |
| Composition functions | F10 | Composition Function 2 (N = 4) | 2400 |
| | F11 | Composition Function 3 (N = 5) | 2600 |
| | F12 | Composition Function 4 (N = 6) | 2700 |

Search Range: $[-100, 100]^D$

## 4.2. Effectiveness analysis of different improvement strategies

In this paper, three improvement strategies are implemented to enhance the performance of the basic RIME, so it is necessary to evaluate the impact of each improvement technique on the performance of MRIME-CD. To fully demonstrate the effect of each improvement strategy on the test functions, the CEC2017 test suite with 10/30/50/100 dimensions and the CEC2022 test suite with 10/20 dimensions will be applied to evaluate the performance of these improvement strategies. In this section, six MRIME-CD variants were involved in the experiment. They combine either a single improvement technique or two improvement techniques, details of which are displayed in Table 5. In Table 5, "Y" indicates that the variant contains this strategy, and "N" indicates that it does not contain this strategy.

**Table 5.** Description of MRIME-CD variants

| Algorithms | GCLS | ABS | SPDM |
|---|---|---|---|
| RIME-G | Y | N | N |
| RIME-A | N | Y | N |
| RIME-S | N | N | Y |
| RIME-GA | Y | Y | N |
| RIME-GS | Y | N | Y |
| RIME-AS | N | Y | Y |
| MRIME-CD | Y | Y | Y |

**Table 6.** Description of MRIME-CD variants

| Test suite | Dimension | RIME | RIME-G | RIME-A | RIME-S | RIME-GA | RIME-GS | RIME-AS | MRIME-CD | P-value |
|---|---|---|---|---|---|---|---|---|---|---|
| CEC 2017 | 10 | 6.97 | 4.79 | 5.45 | 5.38 | 3.86 | 3.66 | 3.90 | **2.00** | 1.27E-13 |
| | 30 | 7.66 | 2.97 | 6.07 | 6.86 | 2.83 | 2.90 | 4.97 | **1.76** | 1.42E-31 |
| | 50 | 7.55 | 3.14 | 5.76 | 6.83 | 2.83 | 3.66 | 4.97 | **1.28** | 2.14E-30 |
| | 100 | 7.55 | 3.14 | 5.41 | 7.24 | 2.55 | 3.93 | 5.17 | **1.00** | 1.62E-34 |
| | Mean rank | 7.43 | 3.51 | 5.67 | 6.58 | 3.02 | 3.53 | 4.75 | **1.51** | N/A |
| CEC 2022 | 10 | 6.92 | 4.00 | 5.92 | 5.17 | 4.17 | 3.25 | 4.25 | **2.33** | 9.72E-05 |
| | 20 | 7.50 | 3.67 | 6.08 | 6.08 | 3.00 | 2.92 | 5.08 | **1.67** | 1.10E-09 |
| | Mean rank | 7.21 | 3.83 | 6.00 | 5.63 | 3.58 | 3.08 | 4.67 | **2.00** | N/A |

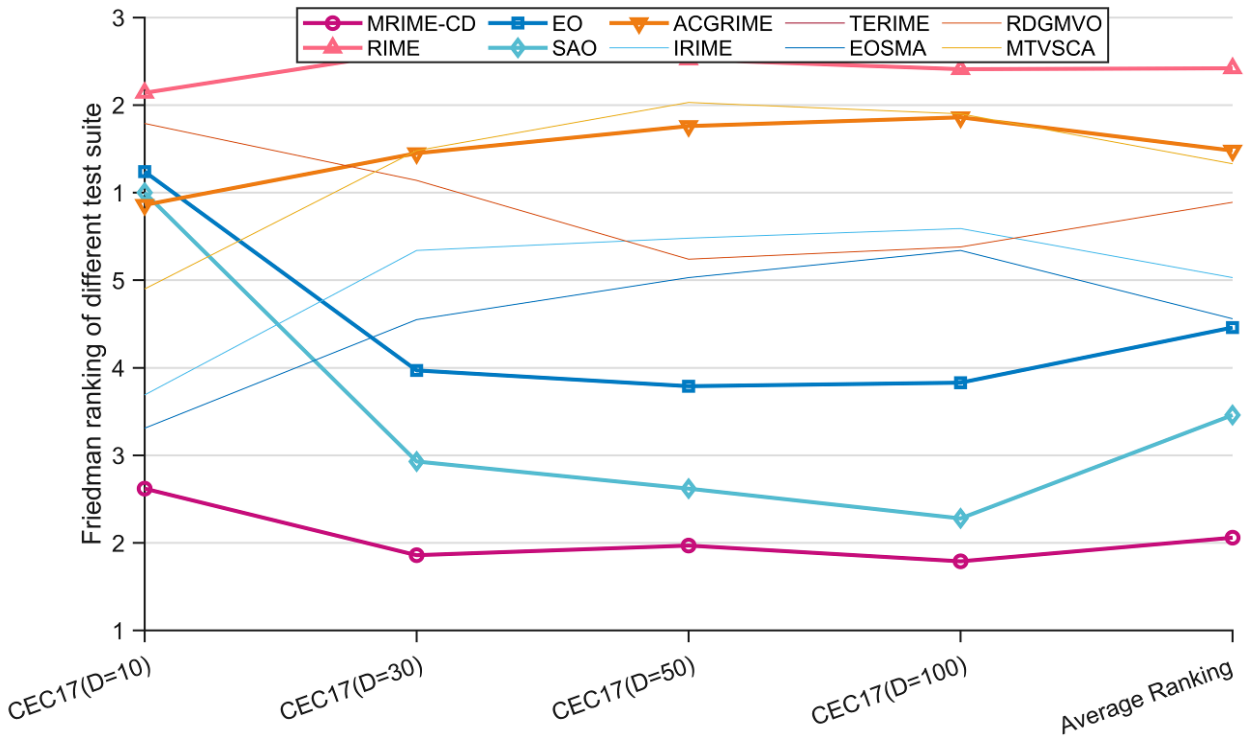

**Figure 3.** Friedman ranking of MRIME-CD and other variants

Due to the large amount of experimental data, this paper uses three statistical tests to analyze the experimental data. The results of the Friedman test for MRIME-CD and other variants are summarized in Table 6. The Friedman test, also known as the multiple comparison test, is used to determine significant differences between multiple algorithmic approaches. Based on the Friedman test p-values in Table 6, we can learn that there are significant differences between these algorithms. The MRIME-CD proposed in this paper has the best overall performance, achieving an average ranking of 1.51 and 1.67 on the two test sets, respectively. In contrast, basic RIME ranks last in the tests. Figure 3 plots line graphs based on the Friedman test rankings of these algorithms to visualize the trend of each algorithm's ranking. The performance of MRIME-CD is stable in different dimensions of the same test set, and there is little difference in performance across different test sets, which clearly demonstrates the superior performance and stability of MRIME-CD across different dimensions. Evaluation rankings of RIME variants combining a single or two improvement techniques also outperform the basic RIME, indicating that all improvement strategies enhance the performance of RIME. Specifically, the GCLS strategy has the greatest impact on MRIME-CD, and ABS and SPDM have more impact on MRIME-CD than each other on the CEC2017 and CEC2022 test sets, respectively. Moreover, except for RIME-GS which is inferior to RIME-G on CEC2017 50/100D, all the other RIME variants combining the two strategies are better than those combining a single strategy, which suggests that the three improvement strategies proposed in this paper are mutually reinforcing and enhancing each other.

**Table 7.** The Wilcoxon rank sum test results (w/ e /l) of MRIME-CD and other variants

| vs. RIME | CEC 2017 test suite | | | | CEC 2022 test suite | |
|---|---|---|---|---|---|---|
| w/e/l | D=10 | D=30 | D=50 | D=100 | D=10 | D=20D |
| RIME-G | 21/5/3 | 27/2/0 | 27/2/0 | 27/2/0 | 9/1/2 | 10/1/1 |
| RIME-A | 8/21/0 | 14/15/0 | 17/12/0 | 21/8/0 | 4/7/1 | 6/5/1 |
| RIME-S | 16/13/0 | 13/16/0 | 7/22/0 | 2/27/0 | 5/7/0 | 8/4/0 |
| RIME-GA | 22/4/3 | 28/1/0 | 26/3/0 | 27/2/0 | 9/2/1 | 11/1/0 |
| RIME-GS | 21/5/3 | 27/2/0 | 26/2/1 | 26/3/0 | 9/1/2 | 11/1/0 |
| RIME-AS | 22/6/1 | 24/5/0 | 23/6/0 | 22/7/0 | 9/3/0 | 10/2/0 |
| MRIME-CD | 25/3/1 | 27/2/0 | 28/1/0 | 29/0/0 | 8/2/2 | 12/0/0 |

Table 7 summarizes the results of the Wilcoxon rank sum test at the 5% significance level for MRIME-CD, the derived algorithms and RIME. The nonparametric Wilcoxon value sum test allows analyzing the significant differences between MRIME-CD and other variants. The symbols "w/e/l" in Table 7 indicate that MRIME-CD performs "better", "similar" and "worse" than the other RIME variants, respectively. "poor". According to Table 7, the number of "w" obtained by all the improved algorithms is more than the number of "l", which indicates that all the RIME variants are superior to the basic RIME. The performance of MRIME-CD obtained in different cases of the total number of "w" is the highest for MRIME-CD in different cases, which indicates that the difference between MRIME-CD and the basic RIME is the highest. The Wilcoxon sum test and Friedman test results are the same, which again confirms the effectiveness of the proposed improvement strategy.

Furthermore, the Kruskal Wallis test further analyzes the differences between MRIME-CD and the derived algorithms as well as the basic RIME. The Kruskal Wallis test evaluates the overall ranking of the different meta-heuristic algorithms by comparing the average of the algorithms' rankings. Figure 4 visualizes the results of the Kruskal Wallis test in different dimensions, and the smallest rank value in the figure indicates that the algorithm outperforms the other algorithms. According to Figure 4, MRIME-CD is able to outperform the other algorithms in all the dimensions considered, providing lower rank values. In conclusion, the three improvement strategies proposed in this paper are complementary and can synergistically enhance the performance of MRIME-CD to adapt it to different types of optimization problems.

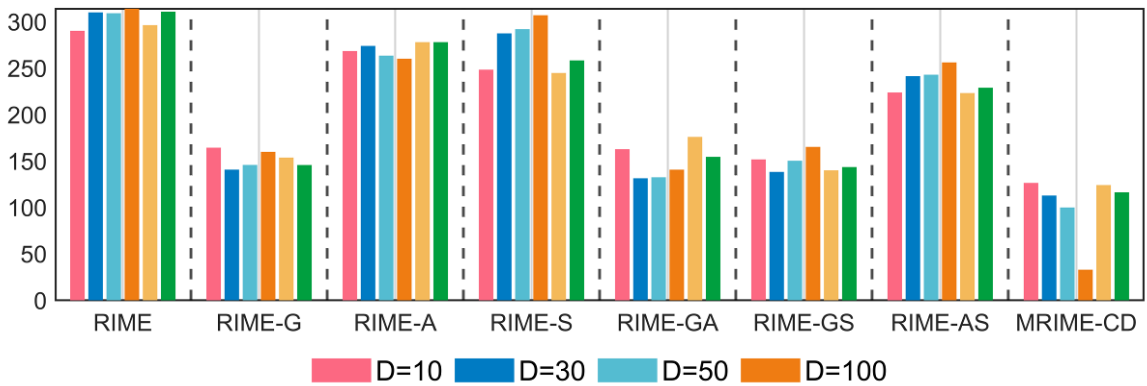

**Figure 4.** Kruskal Wallis ranking of MRIME-CD and other variants

4.3. Experimental case 1: CEC2017 test function analysis

In this subsection, the performance of the proposed MRIME-CD is further validated using the four dimensional functions of the CEC2017 test set. Apart from the basic RIME, two basic algorithms EO [46], SAO [44], three RIME variants ACGRIME [59], IRIME [60], TERIME [61], and three advance improved algorithms EOSMA [62], RDGMVO [63], and MTVSCA [64] are involved in the experiments in this subsection. The experimental environment and parameter settings are presented in Table 1 and Table 2. The specific analysis results are shown next.

### 4.3.1. Analysis of quantitative results

In order to fully demonstrate the performance of MRIME-CD and to avoid stacking of data, this subsection only shows the ranking based on the "mean" of each algorithm in each test function. The records of the best value, mean, standard deviation and ranking are available in Appendix A, Table A1-Table A4. Figure 5 visualizes the ranking of each algorithm on each test function.

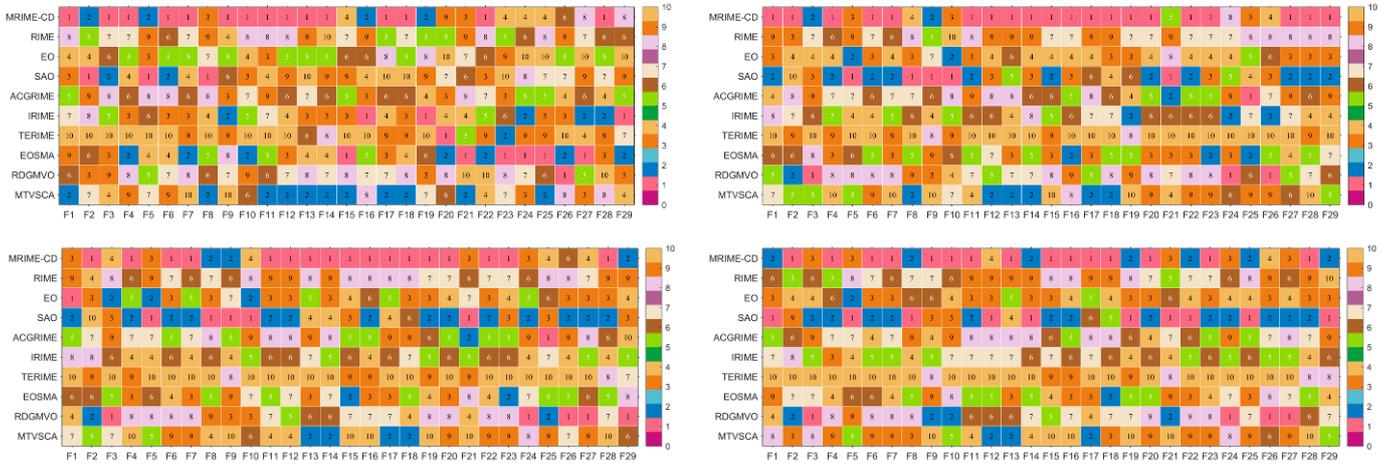

**Figure 5.** Ranking of MRIME-CD and other competitors based on mean values

According to Figure 5, MRIME-CD performs the best on most of the 29 functions in CEC2017, while we can observe that there is no first-ranking occurrence for basic RIME. MRIME-CD ranks first on 15/20/17/15 functions in each of the four test suites, which accounts for half of the entire test suite. Moreover, we can observe that the performance of MRIME-CD is stable in different dimensions, maintaining excellent accuracy and superior robustness regardless of the increase in function dimensions. The overall dominance of MRIME-CD is noticeable in the CE2017 tests. This is due to the fact that GCLS enhances population diversity and balances the exploitation of RIME. ABS mechanism improves the global search of RIME. SPDM mechanism enables to assist the algorithm to get rid of the local optima when it gets stuck in stopping.

### 4.3.2. Analysis of convergence behavior

This subsection provides the convergence curve plots of MRIME-CD and other competitors to visualize the specific evolution of each algorithm during the optimization process. Again, considering the excessive number of convergence graphs for the four cases, the convergence curves for some functions are selected for presentation in Figure 6. The complete convergence graphs are available in Figure A1-Figure A4 in Appendix A.

By analyzing Figure 6, we can realize that MRIME-CD effectively avoids stagnation at local optima and premature convergence, and exhibits better convergence speeds, which outperforms both the basic RIME and the other compared algorithms. Specifically, MRIME-CD demonstrates the fastest convergence speed and higher convergence accuracy on the single-peak functions F1 and F2, which shows that its exploitation ability is not weakened but rather stronger. MRIME-CD performs best on the multi-peak function F6, and outperforms the other comparison algorithms in convergence speed on F8. For hybrid and combinatorial functions, MRIMR-CD provides the best solution in fewer evaluations. In particular, MRIME-CD outperforms the basic RIME in convergence on all these functions. These results further confirm the extraordinary capability of MRIME-CD for global exploration and local exploitation, and also show that the proposed improved method improves the performance of RIME.

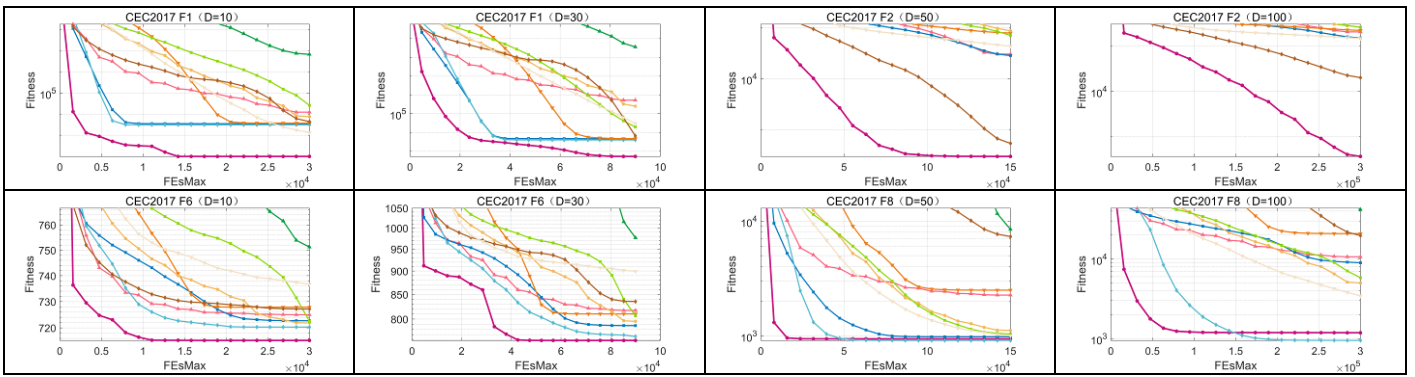

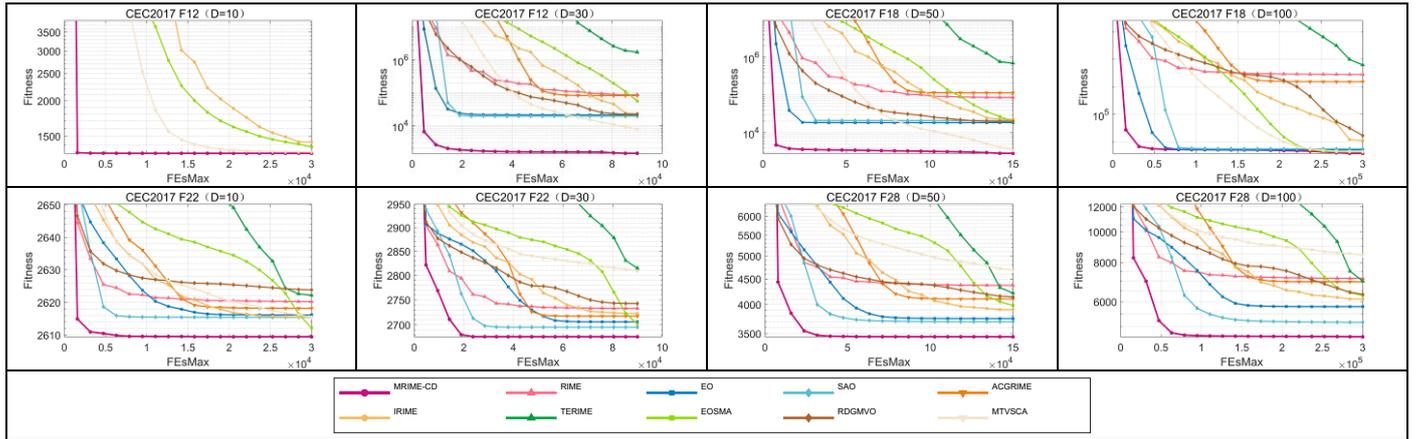

**Figure 6.** Convergence curves of MRIME-CD and other competitors based on CEC2017

4.3.3. Analysis of statistical results

In order to eliminate the randomness and chance introduced by the experiment, the results are further analyzed in this section using statistical tests based on multiple runs. The Wilcoxon rank sum test is first used to determine the performance difference between MRIME-CD and other algorithms. Table 8 summarizes the Wilcoxon rank sum test results of MRIME-CD and the comparison algorithms on the CEC2017 test set. Figure 7 visualizes the number of MRIME-CD better/similar/worse than competitors. According to the Wilcoxon rank sum test results, it is clear that MRIME-CD obtains more number of "w" than "l". This indicates that the performance of MRIME-CD is better than the comparison algorithm in general. It is worth noting that MRIMR-CD only has one function that do not perform as well as the basic RIME on 10D, and there is no weaker performance than the basic RIME on the other three dimensions. The specific wins and losses between MRIME-CD and the comparison algorithms are shown below.

For the 10D functions, MRIME-CD wins (loses) to RIME on 26(1) functions, EO on 24(0) functions, SAO on 22(3) functions, ACGRIME on 23(3) functions, IRIME on 18(3) functions, TERIME on 25(3) functions, EOSMA on 19(7) functions, RDGMVO on 25(2) functions and MTVSCA on 21(5) functions.

For the 30D functions, MRIME-CD wins (loses) to RIME on 28(0) functions, EO on 23(3) functions, SAO on 20(5) functions, ACGRIME on 27(0) functions, IRIME on 27(1) functions, TERIME on 29(0) functions, EOSMA on 25(1) functions, RDGMVO on 25(3) functions and MTVSCA on 26(2) functions.

For the 50D functions, MRIME-CD wins (loses) to RIME on 28(0) functions, EO on 21(4) functions, SAO on 16(8) functions, ACGRIME on 27(1) functions, IRIME on 25(0) functions, TERIME on 29(0) functions, EOSMA on 27(0) functions, RDGMVO on 19(6) functions and MTVSCA on 29(0) functions.

For the 100D functions, MRIME-CD wins (loses) to RIME on 29(0) functions, EO on 22(3) functions, SAO on 13(9) functions, ACGRIME on 29(0) functions, IRIME on 28(0) functions, TERIME on 29(0) functions, EOSMA on 28(0) functions, RDGMVO on 21(4) functions and MTVSCA on 28(0) functions.

**Table 8.** The Wilcoxon rank sum test results (w/ e /l) of MRIME-CD and other competitors based on CEC2017

| MRIME-CD | CEC-2017 test suite | | | |
|---|---|---|---|---|
| vs. w/e/l | D=10 | D=30 | D=50 | D=100 |
| RIME | 26/2/1 | 28/1/0 | 28/1/0 | 29/0/0 |
| EO | 24/5/0 | 23/3/3 | 21/4/4 | 22/4/3 |
| SAO | 22/4/3 | 20/4/5 | 16/5/8 | 13/7/9 |
| ACGRIME | 23/3/3 | 27/2/0 | 27/1/1 | 29/0/0 |
| IRIME | 18/8/3 | 27/1/1 | 25/4/0 | 28/1/0 |
| TERIME | 25/1/3 | 29/0/0 | 29/0/0 | 29/0/0 |
| EOSMA | 19/3/7 | 25/3/1 | 27/2/0 | 28/1/0 |
| RDGMVO | 25/2/2 | 25/1/3 | 19/4/6 | 21/4/4 |
| MTVSCA | 21/3/5 | 26/1/2 | 29/0/0 | 28/0/1 |

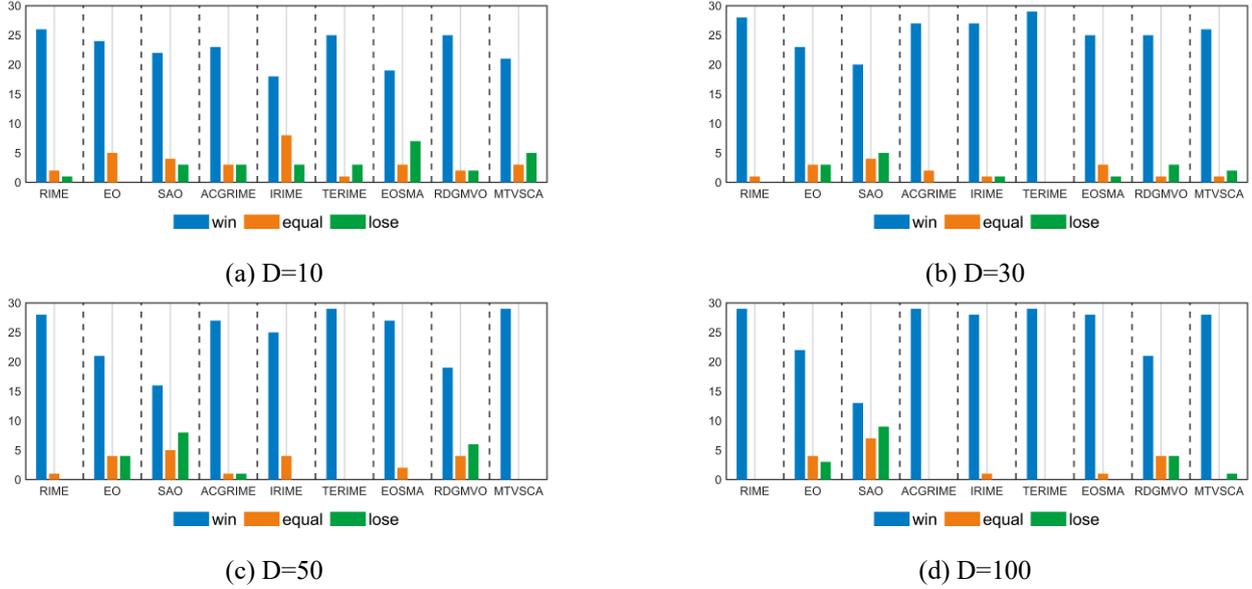

**Figure 7.** The visualization of Wilcoxon rank sum test results of MRIME-CD and other competitors based on CEC2017.

Table 9 shows the results of the Friedman test for MRIME-CD and competitors in the four CEC2017 scenarios. The ranking trends of each algorithm in the four dimensions are visualized in Figure 8. According to Table 9, the p-values for all cases are less than 0.05, which indicates that there are significant differences between the algorithms involved in the experiment. MRIME-CD achieves the best Friedman rankings of 2.62, 1.86, 1.97, and 1.79 in all four cases, corresponding to 7.14, 7.62, 7.52, and 7.41 for basic RIME, which indicates that there is a large gap between the performance of MRIME-CD and basic RIME. Moreover, the Friedman ranking of MRIME-CD fluctuates little among the four cases according to Figure 8, which indicates that MRIME-CD is insensitive to changes in dimensionality and has greater scalability. The Friedman rankings for the MRIME-CD and the basic RIME are specified below. The MRIME-CD is ranked 1st at 10D, the EOSMA is ranked 2nd, and the RIME is ranked 9th. For 30D, the MRIME-CD and SAO are in the top two, and the RIME is ranked 9th. For 50D and 100D, the MRIME-CD is in first place, with the SAO followed by a smaller gap, with RIME both ranked 9ths.

**Table 9.** The Friedman test results (w/ e /l) of MRIME-CD and other competitors based on CEC2017.

| Algorithm | CEC-2017 test suite | | | | |
|---|---|---|---|---|---|
| | D=10 | D=30 | D=50 | D=100 | Average ranking |
| MRIME-CD | **2.62** | **1.86** | **1.97** | **1.79** | **2.06** |
| RIME | 7.14 | 7.62 | 7.52 | 7.41 | 7.42 |
| EO | 6.24 | 3.97 | 3.79 | 3.83 | 4.46 |
| SAO | 6.00 | 2.93 | 2.62 | 2.28 | 3.46 |
| ACGRIME | 5.86 | 6.45 | 6.76 | 6.86 | 6.48 |
| IRIME | 3.69 | 5.34 | 5.48 | 5.59 | 5.03 |
| TERIME | 8.45 | 9.66 | 9.55 | 9.62 | 9.32 |
| EOSMA | 3.31 | 4.55 | 5.03 | 5.34 | 4.56 |
| RDGMVO | 6.79 | 6.14 | 5.24 | 5.38 | 5.89 |
| MTVSCA | 4.90 | 6.48 | 7.03 | 6.90 | 6.33 |
| P-value | 5.96E-17 | 1.32E-27 | 1.98E-28 | 2.04E-30 | N/A |

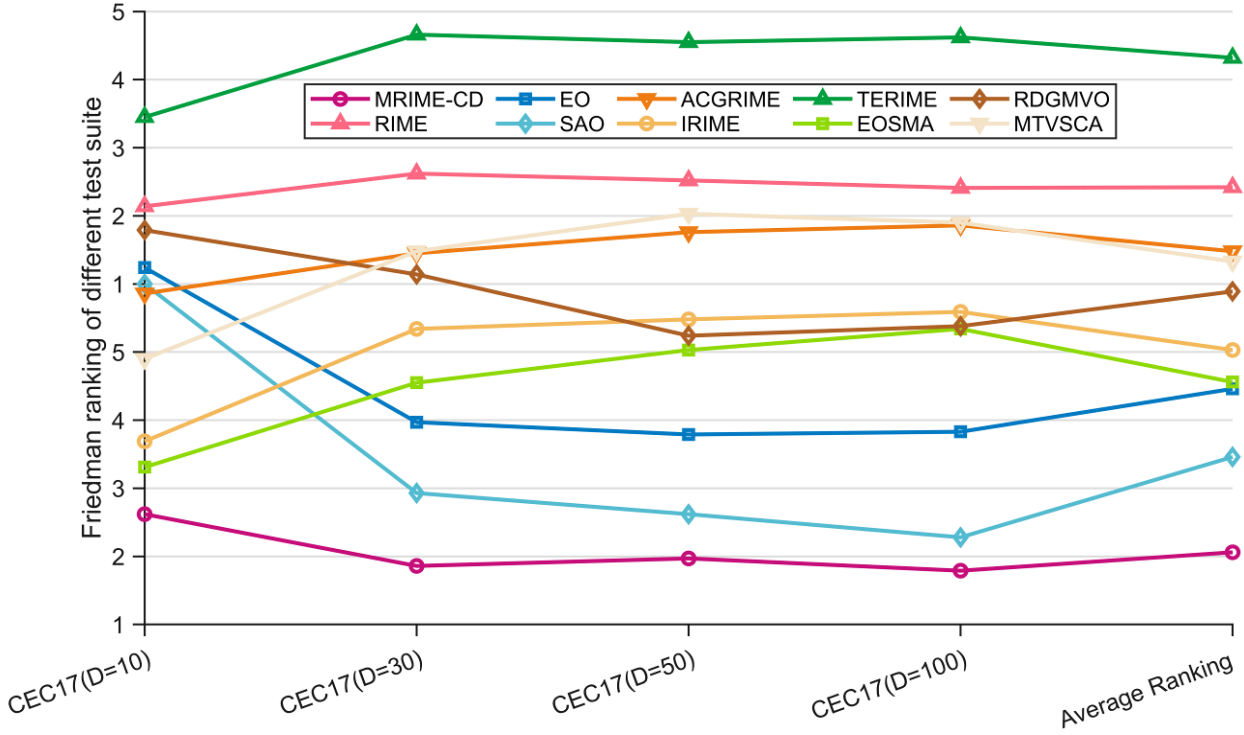

**Figure 8.** The visualization of Friedman test results of MRIME-CD and other competitors based on CEC2017.

4.4. Experimental case 2: CEC2022 test function analysis

In this section, we further validate the performance of MRIME-CD on the CEC2022 test set to illustrate the broad adaptability of the proposed MRIME-CD. In this experimentation, the best value, mean, standard deviation and ranking have been recorded for functions CEC2022 F1-F12, with the dimension of each approach set to 10/20.

4.4.1. Analysis of quantitative results

The experimental results in Table 10 show that MRIMR-CD consistently provides excellent results for most functions, including F1-F2, F4-F8, on the CEC2022 test suite at 10 D. Overall, the advantages of MRIMR-CD at 10 D are obvious. Table 11 records the results of MRIME-CD and the competitors on the CEC2022 20D functions. MRIME-CD provides the best solutions for F1, F3-F4, F6-F8 on 20D.MRIME-CD is weaker than the basic RIME only on F10 and F11 for 10D, and outperforms it across the board on 20D.

**Table 10.** Quantitative results of MRIME-CD and other competitors based on CEC2022 (D=10)

| No. | Metric | MRIME-CD | RIME | EO | SAO | ACGRIME | IRIME | TERIME | EOSMA | RDGMVO | MTVSCA |
|---|---|---|---|---|---|---|---|---|---|---|---|
| F1 | Best | 3.0000E+02 | 3.0013E+02 | 3.0000E+02 | 3.0000E+02 | 3.7168E+02 | 3.0388E+02 | 3.1806E+02 | 3.1861E+02 | 3.0000E+02 | 3.1968E+02 |
|  | Ave | **3.0000E+02** | 3.0095E+02 | 3.0013E+02 | 3.0000E+02 | 8.0730E+02 | 3.5191E+02 | 5.6737E+02 | 3.8907E+02 | 3.0000E+02 | 4.0005E+02 |
|  | Std | 1.3995E-08 | 6.9397E-01 | 1.8308E-01 | 3.7133E-05 | 3.4327E+02 | 5.5789E+01 | 2.0828E+02 | 4.6010E+01 | 6.9695E-04 | 5.8201E+01 |
|  | Rank | 1 | 5 | 4 | 2 | 10 | 6 | 9 | 7 | 3 | 8 |
| F2 | Best | 4.0000E+02 | 4.0001E+02 | 4.0008E+02 | 4.0025E+02 | 4.0041E+02 | 4.0048E+02 | 4.0265E+02 | 4.0026E+02 | 4.0000E+02 | 4.0068E+02 |
|  | Ave | **4.0354E+02** | 4.1211E+02 | 4.0910E+02 | 4.0745E+02 | 4.1101E+02 | 4.0721E+02 | 4.2128E+02 | 4.0488E+02 | 4.1232E+02 | 4.0713E+02 |
|  | Std | 3.7300E+00 | 2.2006E+01 | 1.0394E+01 | 2.5485E+00 | 1.3354E+01 | 2.5429E+00 | 2.7305E+01 | 2.8210E+00 | 2.0849E+01 | 3.7488E+00 |
|  | Rank | 1 | 8 | 6 | 5 | 7 | 4 | 10 | 2 | 9 | 3 |
| F3 | Best | 6.0000E+02 | 6.0009E+02 | 6.0000E+02 | 6.0000E+02 | 6.0000E+02 | 6.0002E+02 | 6.0111E+02 | 6.0001E+02 | 6.0000E+02 | 6.0063E+02 |
|  | Ave | 6.0000E+02 | 6.0022E+02 | 6.0000E+02 | **6.0000E+02** | 6.0009E+02 | 6.0009E+02 | 6.0408E+02 | 6.0007E+02 | 6.0004E+02 | 6.0117E+02 |
|  | Std | 1.5801E-05 | 9.6180E-02 | 1.5969E-04 | 8.3674E-07 | 1.4919E-01 | 4.7217E-02 | 2.3186E+00 | 5.4200E-02 | 3.3757E-02 | 3.1946E-01 |
|  | Rank | 2 | 8 | 3 | 1 | 7 | 6 | 10 | 5 | 4 | 9 |

|  |  | MRIME-CD | RIME | EO | SAO | ACGRIME | IRIME | TERIME | EOSMA | RDGMVO | MTVSCA |
|---|---|---|---|---|---|---|---|---|---|---|---|
| F4 | Best | 8.0298E+02 | 8.0301E+02 | 8.0199E+02 | 8.0298E+02 | 8.0796E+02 | 8.0299E+02 | 8.1319E+02 | 8.0261E+02 | 8.0796E+02 | 8.1824E+02 |
|  | Ave | **8.0607E+02** | 8.1822E+02 | 8.0860E+02 | 8.1356E+02 | 8.1915E+02 | 8.1671E+02 | 8.2915E+02 | 8.0895E+02 | 8.2468E+02 | 8.3139E+02 |
|  | Std | 2.2002E+00 | 8.6447E+00 | 3.3347E+00 | 7.9983E+00 | 7.4159E+00 | 7.1460E+00 | 1.0001E+01 | 3.9265E+00 | 1.0226E+01 | 5.3779E+00 |
|  | Rank | 1 | 6 | 2 | 4 | 7 | 5 | 9 | 3 | 8 | 10 |
| F5 | Best | 9.0000E+02 | 9.0000E+02 | 9.0000E+02 | 9.0000E+02 | 9.0000E+02 | 9.0000E+02 | 9.0092E+02 | 9.0000E+02 | 9.0000E+02 | 9.0032E+02 |
|  | Ave | **9.0001E+02** | 9.0036E+02 | 9.0002E+02 | 9.0002E+02 | 9.0070E+02 | 9.0011E+02 | 9.5721E+02 | 9.0010E+02 | 9.0075E+02 | 9.0161E+02 |
|  | Std | 2.4309E-02 | 4.4590E-01 | 9.0073E-02 | 9.0073E-02 | 1.6102E+00 | 1.2827E-01 | 8.2113E+01 | 1.5315E-01 | 1.5015E+00 | 8.7867E-01 |
|  | Rank | 1 | 6 | 3 | 2 | 7 | 5 | 10 | 4 | 8 | 9 |
| F6 | Best | 1.8000E+03 | 1.8902E+03 | 1.8806E+03 | 1.8189E+03 | 1.8607E+03 | 1.8360E+03 | 3.3092E+03 | 2.2645E+03 | 1.8047E+03 | 1.8575E+03 |
|  | Ave | **1.8010E+03** | 3.9654E+03 | 4.9403E+03 | 4.0890E+03 | 4.6432E+03 | 4.9950E+03 | 4.0046E+04 | 4.0477E+03 | 4.2531E+03 | 2.0239E+03 |
|  | Std | 6.7846E-01 | 2.0225E+03 | 2.3356E+03 | 2.0323E+03 | 2.0239E+03 | 2.1048E+03 | 4.5141E+04 | 1.6832E+03 | 2.7322E+03 | 1.6330E+02 |
|  | Rank | 1 | 3 | 8 | 5 | 7 | 9 | 10 | 4 | 6 | 2 |
| F7 | Best | 2.0000E+03 | 2.0008E+03 | 2.0000E+03 | 2.0000E+03 | 2.0010E+03 | 2.0002E+03 | 2.0086E+03 | 2.0068E+03 | 2.0000E+03 | 2.0189E+03 |
|  | Ave | **2.0103E+03** | 2.0177E+03 | 2.0184E+03 | 2.0193E+03 | 2.0183E+03 | 2.0182E+03 | 2.0285E+03 | 2.0219E+03 | 2.0152E+03 | 2.0320E+03 |
|  | Std | 9.9962E+00 | 7.5903E+00 | 7.3314E+00 | 6.1570E+00 | 6.4919E+00 | 6.3674E+00 | 6.0073E+00 | 5.9007E+00 | 8.5269E+00 | 4.3118E+00 |
|  | Rank | 1 | 3 | 6 | 7 | 5 | 4 | 9 | 8 | 2 | 10 |
| F8 | Best | 2.2000E+03 | 2.2013E+03 | 2.2015E+03 | 2.2002E+03 | 2.2018E+03 | 2.2003E+03 | 2.2023E+03 | 2.2079E+03 | 2.2009E+03 | 2.2165E+03 |
|  | Ave | **2.2055E+03** | 2.2159E+03 | 2.2187E+03 | 2.2172E+03 | 2.2191E+03 | 2.2149E+03 | 2.2226E+03 | 2.2239E+03 | 2.2193E+03 | 2.2249E+03 |
|  | Std | 8.7305E+00 | 8.8465E+00 | 7.5613E+00 | 7.7053E+00 | 6.0998E+00 | 8.6889E+00 | 3.5370E+00 | 5.0540E+00 | 4.5866E+00 | 3.0002E+00 |
|  | Rank | 1 | 3 | 5 | 4 | 6 | 2 | 8 | 9 | 7 | 10 |
| F9 | Best | 2.5293E+03 | 2.5293E+03 | 2.5293E+03 | 2.5293E+03 | 2.5293E+03 | 2.5293E+03 | 2.5294E+03 | 2.4286E+03 | 2.4855E+03 | 2.5300E+03 |
|  | Ave | 2.5293E+03 | 2.5293E+03 | 2.5293E+03 | 2.5293E+03 | 2.5293E+03 | 2.5293E+03 | 2.5334E+03 | 2.5273E+03 | **2.4893E+03** | 2.5309E+03 |
|  | Std | 1.0076E-10 | 3.9055E-03 | 3.5807E-13 | 0.0000E+00 | 1.0600E-01 | 4.6402E-04 | 4.1962E+00 | 1.4100E+01 | 2.6701E+01 | 4.7963E-01 |
|  | Rank | 5 | 7 | 3 | 3 | 8 | 6 | 10 | 2 | 1 | 9 |
| F10 | Best | 2.5000E+03 | 2.5003E+03 | 2.5002E+03 | 2.5002E+03 | 2.5002E+03 | 2.5002E+03 | 2.5004E+03 | 2.5002E+03 | 2.5002E+03 | 2.5002E+03 |
|  | Ave | 2.5252E+03 | 2.5079E+03 | 2.5314E+03 | 2.5132E+03 | 2.5027E+03 | 2.5004E+03 | 2.5007E+03 | **2.5004E+03** | 2.5517E+03 | 2.5005E+03 |
|  | Std | 4.5577E+01 | 3.0434E+01 | 5.1186E+01 | 3.5631E+01 | 1.6294E+01 | 8.5828E-02 | 2.2031E-01 | 7.1320E-02 | 6.2134E+01 | 9.6988E-02 |
|  | Rank | 8 | 6 | 9 | 7 | 5 | 2 | 4 | 1 | 10 | 3 |
| F11 | Best | 2.6000E+03 | 2.6118E+03 | 2.6000E+03 | 2.6000E+03 | 2.6000E+03 | 2.6018E+03 | 2.7641E+03 | 2.6025E+03 | 2.6000E+03 | 2.7077E+03 |
|  | Ave | 2.8967E+03 | 2.8640E+03 | 2.7667E+03 | 2.6853E+03 | 2.6345E+03 | **2.6172E+03** | 3.1735E+03 | 2.6476E+03 | 2.7325E+03 | 2.8641E+03 |
|  | Std | 4.2710E+01 | 1.1684E+02 | 1.5022E+02 | 1.3502E+02 | 6.1786E+01 | 1.6045E+01 | 3.9162E+02 | 8.6930E+01 | 1.5101E+02 | 1.2864E+02 |
|  | Rank | 9 | 7 | 6 | 4 | 2 | 1 | 10 | 3 | 5 | 8 |
| F12 | Best | 2.8594E+03 | 2.8595E+03 | 2.8594E+03 | 2.8586E+03 | 2.8586E+03 | 2.8586E+03 | 2.8615E+03 | 2.8594E+03 | **2.8460E+03** | 2.8641E+03 |
|  | Ave | 2.8634E+03 | 2.8639E+03 | 2.8630E+03 | 2.8632E+03 | 2.8627E+03 | 2.8608E+03 | 2.8656E+03 | 2.8619E+03 | 2.8480E+03 | 2.8658E+03 |
|  | Std | 1.9882E+00 | 1.5781E+00 | 1.0934E+00 | 1.1987E+00 | 1.3671E+00 | 1.2823E+00 | 1.7998E+00 | 1.6649E+00 | 1.5511E+00 | 6.7832E-01 |
|  | Rank | 7 | 8 | 5 | 6 | 4 | 2 | 9 | 3 | 1 | 10 |

**Table 11.** Quantitative results of MRIME-CD and other competitors based on CEC2022 (D=20)

| No. | Metric | MRIME-CD | RIME | EO | SAO | ACGRIME | IRIME | TERIME | EOSMA | RDGMVO | MTVSCA |
|---|---|---|---|---|---|---|---|---|---|---|---|
| F1 | Best | 3.0000E+02 | 3.3694E+02 | 3.3145E+02 | 6.5288E+02 | 2.5208E+03 | 1.4865E+03 | 3.1984E+03 | 1.9600E+03 | 3.0000E+02 | 8.1240E+02 |
|  | Ave | **3.0000E+02** | 3.9913E+02 | 4.6361E+02 | 4.0220E+03 | 8.2326E+03 | 4.9318E+03 | 6.9149E+03 | 4.7544E+03 | 3.0001E+02 | 2.6523E+03 |
|  | Std | 1.3970E-07 | 4.7235E+01 | 1.3846E+02 | 2.5850E+03 | 2.7623E+03 | 2.3652E+03 | 2.3922E+03 | 1.2866E+03 | 7.3468E-03 | 9.7154E+02 |
|  | Rank | 1 | 3 | 4 | 6 | 10 | 8 | 9 | 7 | 2 | 5 |
| F2 | Best | 4.0000E+02 | 4.0574E+02 | 4.0563E+02 | 4.4490E+02 | 4.4490E+02 | 4.4492E+02 | 4.5268E+02 | 4.4497E+02 | 4.0428E+02 | 4.5202E+02 |
|  | Ave | 4.4095E+02 | 4.5231E+02 | 4.5281E+02 | 4.5171E+02 | 4.5898E+02 | 4.5212E+02 | 4.8142E+02 | 4.5427E+02 | **4.3446E+02** | 4.5812E+02 |
|  | Std | 1.8842E+01 | 1.4357E+01 | 1.1447E+01 | 7.5424E+00 | 1.2188E+01 | 7.7930E+00 | 2.5714E+01 | 9.4634E+00 | 2.8818E+01 | 6.2866E+00 |
|  | Rank | 2 | 5 | 6 | 3 | 9 | 4 | 10 | 7 | 1 | 8 |
|  | Best | 6.0000E+02 | 6.0038E+02 | 6.0000E+02 | 6.0000E+02 | 6.0009E+02 | 6.0011E+02 | 6.0333E+02 | 6.0017E+02 | 6.0005E+02 | 6.0119E+02 |
| F3 | Ave | **6.0000E+02** | 6.0117E+02 | 6.0000E+02 | 6.0000E+02 | 6.0067E+02 | 6.0038E+02 | 6.0900E+02 | 6.0044E+02 | 6.0023E+02 | 6.0257E+02 |
|  | Std | 1.9648E-04 | 6.7478E-01 | 1.5093E-03 | 2.0205E-04 | 4.7034E-01 | 1.5191E-01 | 2.9600E+00 | 1.9605E-01 | 1.7429E-01 | 7.5128E-01 |
|  | Rank | 1 | 8 | 3 | 2 | 7 | 5 | 10 | 6 | 4 | 9 |
|  | Best | 8.0597E+02 | 8.2000E+02 | 8.0796E+02 | 8.1094E+02 | 8.2463E+02 | 8.2010E+02 | 8.4646E+02 | 8.1457E+02 | 8.3482E+02 | 8.7821E+02 |
| F4 | Ave | **8.1544E+02** | 8.4283E+02 | 8.2767E+02 | 8.3012E+02 | 8.5107E+02 | 8.4099E+02 | 8.9220E+02 | 8.3408E+02 | 8.6270E+02 | 9.0213E+02 |
|  | Std | 5.1214E+00 | 1.5364E+01 | 1.0898E+01 | 2.1367E+01 | 1.5705E+01 | 1.2447E+01 | 2.3917E+01 | 1.0172E+01 | 1.5941E+01 | 1.0571E+01 |
|  | Rank | 1 | 6 | 2 | 3 | 7 | 5 | 9 | 4 | 8 | 10 |
|  | Best | 9.0000E+02 | 9.0032E+02 | 9.0000E+02 | 9.0000E+02 | 9.0403E+02 | 9.0050E+02 | 1.0256E+03 | 9.0027E+02 | 9.0147E+02 | 9.0330E+02 |
| F5 | Ave | 9.0028E+02 | 9.2067E+02 | 9.0050E+02 | **9.0025E+02** | 1.0014E+03 | 9.0422E+02 | 2.1083E+03 | 9.0298E+02 | 1.0325E+03 | 9.1190E+02 |
|  | Std | 4.2409E-01 | 2.7361E+01 | 7.9206E-01 | 5.2805E-01 | 1.3187E+02 | 3.7051E+00 | 9.3575E+02 | 2.1101E+00 | 2.0710E+02 | 8.1309E+00 |
|  | Rank | 2 | 7 | 3 | 1 | 8 | 5 | 10 | 4 | 9 | 6 |
|  | Best | 1.8002E+03 | 2.7679E+03 | 1.8576E+03 | 1.8226E+03 | 1.9579E+03 | 2.0070E+03 | 1.1621E+05 | 2.9866E+04 | 1.8442E+03 | 7.5503E+03 |
| F6 | Ave | **1.8149E+03** | 1.2959E+04 | 3.8884E+03 | 5.2213E+03 | 4.0638E+03 | 1.5349E+04 | 1.2629E+06 | 1.5125E+05 | 7.7408E+03 | 4.0826E+04 |
|  | Std | 1.9441E+01 | 8.6687E+03 | 2.2553E+03 | 5.5589E+03 | 2.1616E+03 | 9.5942E+03 | 1.0117E+06 | 1.1410E+05 | 9.3796E+03 | 2.7047E+04 |
|  | Rank | 1 | 6 | 2 | 4 | 3 | 7 | 10 | 9 | 5 | 8 |
|  | Best | 2.0013E+03 | 2.0290E+03 | 2.0214E+03 | 2.0210E+03 | 2.0227E+03 | 2.0239E+03 | 2.0394E+03 | 2.0297E+03 | 2.0231E+03 | 2.0621E+03 |
| F7 | Ave | **2.0227E+03** | 2.0517E+03 | 2.0375E+03 | 2.0407E+03 | 2.0396E+03 | 2.0350E+03 | 2.0684E+03 | 2.0475E+03 | 2.0660E+03 | 2.0849E+03 |
|  | Std | 3.4890E+00 | 1.5748E+01 | 1.3170E+01 | 1.5801E+01 | 1.1431E+01 | 8.3373E+00 | 1.8966E+01 | 8.3668E+00 | 3.6937E+01 | 1.1359E+01 |
|  | Rank | 1 | 7 | 3 | 5 | 4 | 2 | 9 | 6 | 8 | 10 |
|  | Best | 2.2204E+03 | 2.2226E+03 | 2.2213E+03 | 2.2205E+03 | 2.2217E+03 | 2.2212E+03 | 2.2242E+03 | 2.2261E+03 | 2.2210E+03 | 2.2319E+03 |
| F8 | Ave | **2.2211E+03** | 2.2279E+03 | 2.2265E+03 | 2.2223E+03 | 2.2256E+03 | 2.2235E+03 | 2.2310E+03 | 2.2319E+03 | 2.2321E+03 | 2.2370E+03 |
|  | Std | 4.9396E-01 | 4.4521E+00 | 2.9446E+00 | 3.0604E+00 | 3.2867E+00 | 1.2811E+00 | 8.5251E+00 | 2.2045E+00 | 3.0405E+01 | 2.6800E+00 |
|  | Rank | 1 | 6 | 5 | 2 | 4 | 3 | 7 | 8 | 9 | 10 |
|  | Best | 2.4808E+03 | 2.4808E+03 | 2.4808E+03 | 2.4808E+03 | 2.4808E+03 | 2.4808E+03 | 2.4822E+03 | 2.4808E+03 | 2.4654E+03 | 2.4823E+03 |
| F9 | Ave | 2.4808E+03 | 2.4811E+03 | 2.4808E+03 | 2.4808E+03 | 2.4821E+03 | 2.4808E+03 | 2.4908E+03 | 2.4809E+03 | **2.4656E+03** | 2.4836E+03 |
|  | Std | 7.8086E-09 | 2.5700E-01 | 2.5251E-04 | 2.1478E-05 | 1.3269E+00 | 2.1841E-02 | 8.2041E+00 | 8.9459E-02 | 1.7220E-01 | 7.0782E-01 |
|  | Rank | 2 | 7 | 4 | 3 | 8 | 5 | 10 | 6 | 1 | 9 |
|  | Best | 2.4169E+03 | 2.4103E+03 | 2.5003E+03 | 2.5004E+03 | 2.4253E+03 | 2.4296E+03 | 2.5005E+03 | 2.5003E+03 | 2.4186E+03 | 2.5005E+03 |
| F10 | Ave | 2.5104E+03 | 2.5521E+03 | 2.8355E+03 | 2.5542E+03 | 2.5044E+03 | 2.5033E+03 | 2.5121E+03 | **2.5005E+03** | 2.5798E+03 | 2.5166E+03 |
|  | Std | 4.8590E+01 | 1.2292E+02 | 5.0251E+02 | 1.7349E+02 | 3.5925E+01 | 2.4321E+01 | 4.3788E+01 | 1.0565E-01 | 1.4484E+02 | 5.4944E+01 |
|  | Rank | 4 | 7 | 10 | 8 | 3 | 2 | 5 | 1 | 9 | 6 |
|  | Best | 2.9000E+03 | 2.9431E+03 | 2.9000E+03 | 2.9000E+03 | 2.6511E+03 | 2.9797E+03 | 3.9660E+03 | 2.9171E+03 | 2.6005E+03 | 3.3020E+03 |
| F11 | Ave | 2.9005E+03 | 2.9999E+03 | **2.9000E+03** | 2.9020E+03 | 2.9824E+03 | 3.2252E+03 | 4.6470E+03 | 2.9358E+03 | 2.9213E+03 | 3.5300E+03 |
|  | Std | 3.3918E+00 | 2.4768E+01 | 2.7987E-04 | 1.4003E+01 | 1.3452E+02 | 9.6320E+01 | 3.6933E+02 | 9.5757E+00 | 1.1686E+02 | 1.1199E+02 |
|  | Rank | 2 | 7 | 1 | 3 | 6 | 8 | 10 | 5 | 4 | 9 |
|  | Best | 2.9329E+03 | 2.9377E+03 | 2.9341E+03 | 2.9329E+03 | 2.9351E+03 | 2.9336E+03 | 2.9478E+03 | 2.9358E+03 | 2.8912E+03 | 2.9645E+03 |
| F12 | Ave | 2.9460E+03 | 2.9483E+03 | 2.9437E+03 | 2.9447E+03 | 2.9440E+03 | 2.9408E+03 | 2.9707E+03 | 2.9448E+03 | **2.8947E+03** | 2.9817E+03 |
|  | Std | 1.0591E+01 | 7.4177E+00 | 5.9420E+00 | 6.7704E+00 | 4.5573E+00 | 3.9188E+00 | 1.4153E+01 | 4.2591E+00 | 2.6186E+00 | 9.1024E+00 |
|  | Rank | 7 | 8 | 3 | 5 | 4 | 2 | 9 | 6 | 1 | 10 |

### 4.4.2. Analysis of convergence behavior

To further check the convergence of MRIME-CD for different functions, Figure 10 shows some convergence curves of MRIMR-CD and the comparison algorithm on the CEC2022 test set. The complete convergence graphs are available in Figure A5-Figure A6 in Appendix A. From Figure 10, it can be observed that the proposed MRIME-CD shows better and faster convergence on most functions. Moreover, MRIME-CD is not sensitive to dimension and can output satisfactory solutions stably. However, MRIMR-CD has half convergence on combinatorial functions. Although it shows very fast convergence early on, the convergence accuracy is not enough to outperform other algorithms.

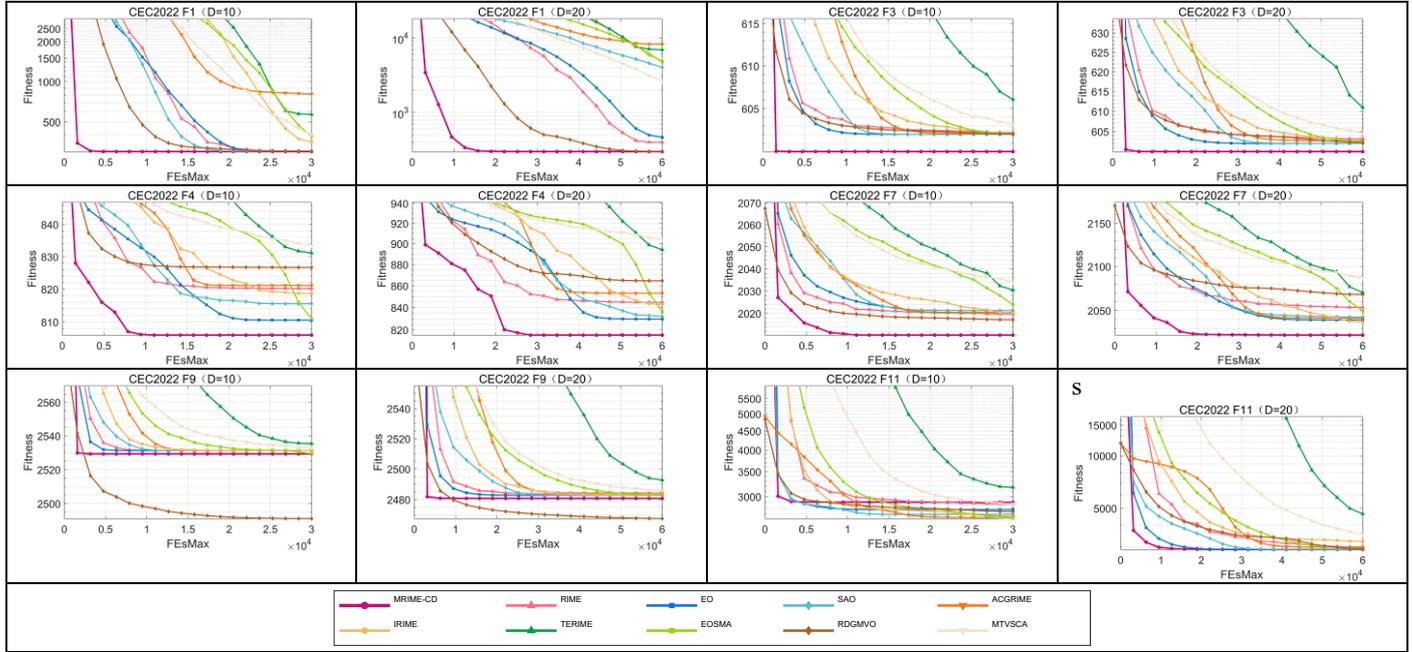

**Figure 10.** Convergence curves of MRIME-CD and other competitors based on CEC2022

### 4.4.3. Analysis of statistical results

The Kruskal Wallis test was used to analyze the experimental results of MRIMR-CD and competitors on the CEC2022 test set. The different dimensional Kruskal Wallis test results are presented in Figure 11. It is clear that MRIMME-CD provides a lower mean of ranks in all of the cases and outranks the other algorithms accordingly. It is worth noting that the average rank of MRIMR-CD decreases with increasing dimensionality, which indicates good scalability. In addition, Table 12 and Table 13 provide the p-values for the Kruskal Wallis test. Similar to the Wilcoxon rank sum test, there was no difference between the MRIMR-CD and the comparison algorithm when the p-value was greater than 0.05. When the p-value is less than 0.05, "+" and "-" indicate that MRIME-CD is superior and inferior to the competitor, respectively. According to Table 12, MRIMR-CD dominates on most functions, and Figure 12 visualizes these results.

The Friedman test is employed to determine the significant differences among multiple algorithmic approaches. The average rankings of each case for MRIME-CD and other competing algorithms are provided in Table 14. This is further supported by Figure 13, a visual depiction of the rankings according to the Friedman test, which clearly shows that MRIME-CD is superior to the other algorithms. According to the test results, the proposed MRIME-CD performs the best and ranks first overall among the ten algorithms, achieving an average ranking of 3.17 and 2.08, followed by SAO and EO.

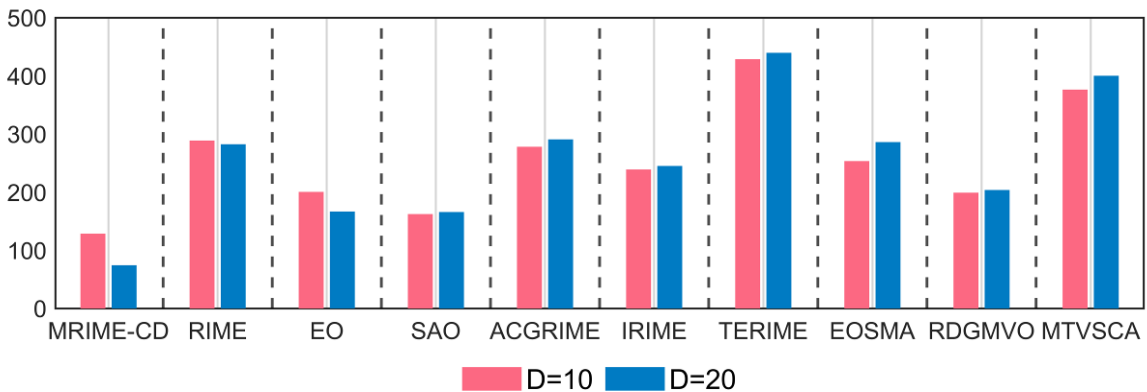

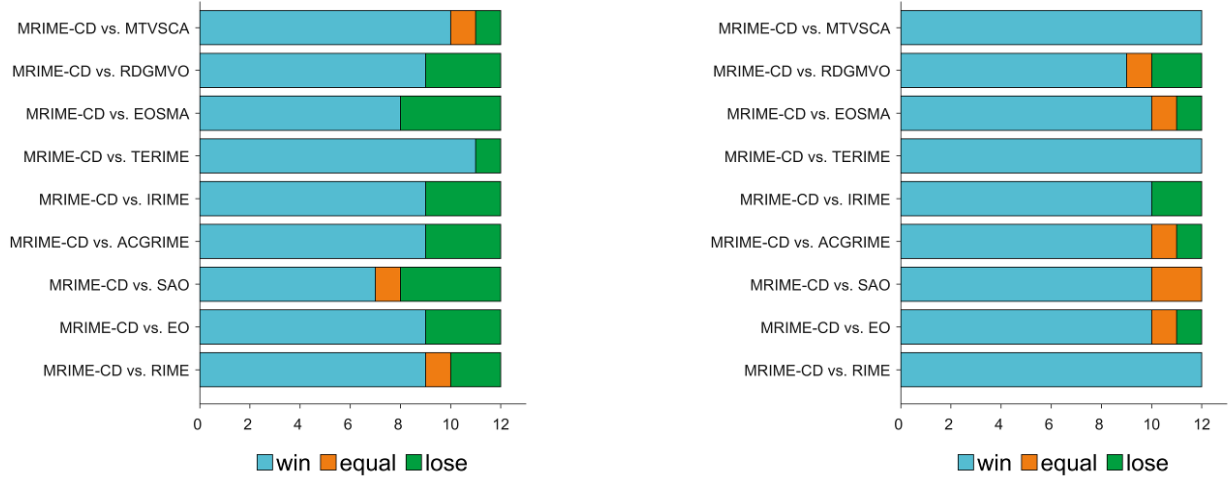

(a) D=10  (b) D=20

**Figure 12.** The visualization of Kruskal Wallis test results of MRIME-CD and other competitors based on CEC2022.

**Table 12.** The p values of Kruskal Wallis test of MRIME-CD and other competitors based on CEC2022 (D=10)

| No | RIME | | EO | | SAO | | ACGRIME | | IRIME | | TERIME | | EOSMA | | RDGMVO | | MTVSCA | |
|---|---|---|---|---|---|---|---|---|---|---|---|---|---|---|---|---|---|---|
| F1 | 3.21E-18 | + | 3.21E-18 | + | 2.82E-11 | + | 3.21E-18 | + | 3.21E-18 | + | 3.21E-18 | + | 3.21E-18 | + | 3.21E-18 | + | 3.21E-18 | + |
| F2 | 7.20E-07 | + | 1.25E-04 | + | 2.04E-04 | + | 5.80E-11 | + | 1.20E-09 | + | 3.08E-11 | + | 2.81E-03 | + | 2.68E-04 | + | 4.62E-06 | + |
| F3 | 3.21E-18 | + | 1.80E-02 | + | 3.21E-18 | - | 3.21E-18 | + | 3.21E-18 | + | 3.21E-18 | + | 3.21E-18 | + | 3.21E-18 | + | 3.21E-18 | + |
| F4 | 1.51E-14 | + | 1.87E-05 | + | 7.55E-10 | + | 2.76E-17 | + | 3.26E-15 | + | 3.21E-18 | + | 5.85E-05 | + | 7.73E-18 | + | 3.21E-18 | + |
| F5 | 2.00E-16 | + | 6.27E-07 | + | 1.44E-13 | + | 1.60E-16 | + | 9.39E-16 | + | 3.21E-18 | + | 1.45E-15 | + | 2.07E-17 | + | 3.21E-18 | + |
| F6 | 3.21E-18 | + | 3.21E-18 | + | 3.21E-18 | + | 3.21E-18 | + | 3.21E-18 | + | 3.21E-18 | + | 3.21E-18 | + | 3.21E-18 | + | 3.21E-18 | + |
| F7 | 6.73E-12 | + | 6.73E-12 | + | 2.46E-11 | + | 2.57E-11 | + | 3.22E-13 | + | 1.47E-17 | + | 1.18E-11 | + | 1.00E-03 | + | 1.47E-17 | + |
| F8 | 8.20E-13 | + | 1.43E-14 | + | 4.94E-10 | + | 8.04E-15 | + | 2.09E-10 | + | 1.47E-17 | + | 4.73E-15 | + | 5.84E-12 | + | 1.51E-16 | + |
| F9 | 3.20E-18 | + | 6.03E-19 | - | 1.34E-20 | - | 3.20E-18 | + | 3.20E-18 | + | 3.20E-18 | + | 6.13E-17 | - | 6.13E-17 | - | 3.20E-18 | + |
| F10 | 1.42E-06 | - | 1.36E-07 | + | 7.71E-07 | - | 4.34E-06 | - | 6.16E-06 | - | 4.07E-06 | - | 9.25E-06 | - | 3.99E-10 | + | 4.48E-06 | + |
| F11 | 1.42E-05 | - | 2.95E-16 | - | 7.73E-18 | - | 5.73E-16 | - | 6.13E-17 | - | 8.09E-04 | + | 3.06E-13 | - | 5.07E-04 | - | 7.40E-01 | = |
| F12 | 6.72E-02 | = | 9.86E-03 | - | 8.35E-02 | = | 1.88E-03 | - | 2.52E-09 | - | 2.43E-08 | + | 3.56E-04 | + | 3.21E-18 | - | 1.11E-15 | - |
| +/=/- | 9/1/2 | | 9/0/3 | | 7/1/4 | | 9/0/3 | | 9/0/3 | | 11/0/1 | | 8/0/4 | | 9/0/3 | | 10/1/1 | |

**Table 13.** The p values of Kruskal Wallis test of MRIME-CD and other competitors based on CEC2022 (D=20)

| No | RIME | EO | SAO | ACGRIME | IRIME | TERIME | EOSMA | RDGMVO | | MTVSCA |
|---|---|---|---|---|---|---|---|---|---|---|
| F1 | 3.21E-18 | 3.21E-18 | 3.21E-18 | 3.21E-18 | 3.21E-18 | 3.21E-18 | 3.21E-18 | 3.21E-18 | | 3.21E-18 |
| F2 | 6.38E-10 | 1.07E-13 | 2.07E-14 | 2.39E-13 | 2.07E-14 | 2.32E-17 | 7.87E-14 | 1.47E-01 | = | 5.73E-16 |
| F3 | 3.21E-18 | 2.64E-16 | 1.36E-09 | 3.21E-18 | 3.21E-18 | 3.21E-18 | 3.21E-18 | 3.21E-18 | | 3.21E-18 |
| F4 | 1.31E-17 | 1.63E-11 | 7.81E-09 | 3.40E-18 | 8.20E-18 | 3.21E-18 | 4.60E-16 | 3.21E-18 | | 3.21E-18 |
| F5 | 8.15E-17 | 9.34E-04 | 5.18E-01 | = | 3.21E-18 | 7.70E-17 | 3.21E-18 | 8.15E-17 | 3.40E-18 | 3.21E-18 |
| F6 | 3.21E-18 | 3.83E-18 | 5.77E-18 | 3.21E-18 | 3.21E-18 | 3.21E-18 | 3.21E-18 | 4.31E-18 | | 3.21E-18 |
| F7 | 3.21E-18 | 8.94E-15 | 8.94E-15 | 4.62E-17 | 4.35E-16 | 3.21E-18 | 3.21E-18 | 4.12E-17 | | 3.21E-18 |
| F8 | 3.40E-18 | 9.20E-18 | 3.41E-03 | 9.20E-18 | 3.89E-17 | 3.20E-18 | 3.20E-18 | 8.28E-11 | | 3.20E-18 |
| F9 | 3.21E-18 | 2.19E-17 | 7.29E-18 | 3.21E-18 | 3.21E-18 | 3.21E-18 | 3.21E-18 | 3.21E-18 | - | 3.21E-18 |
| F10 | 9.90E-11 | 1.68E-13 | 4.01E-12 | 8.22E-10 | - | 1.54E-10 | - | 1.42E-11 | 2.95E-11 | - | 2.28E-10 | 1.12E-11 |

| | | | | | | | | | | |
|---|---|---|---|---|---|---|---|---|---|---|
| F11 | 3.21E-18 | | 3.21E-18 | - | 6.14E-17 | 2.64E-16 | 3.21E-18 | 3.21E-18 | 4.31E-18 | 1.01E-13 | 3.21E-18 |
| F12 | 3.36E-02 | | 5.85E-01 | = | 9.81E-01 | = | 9.33E-01 | = | 1.37E-02 | - | 1.01E-13 | 4.84E-01 | = | 3.21E-18 | - | 2.32E-17 |
| +/=/- | 12/0/0 | | 10/1/1 | | 10/2/0 | | 10/1/1 | | 10/0/2 | | 12/0/0 | 10/1/1 | 9/1/2 | 12/0/0 |

**Table 14.** The Friedman test results of MRIME-CD and other competitors based on CEC2022

| Algorithm | CEC-2022 test suite | | |
|---|---|---|---|
| | 10D | 20D | Average ranking |
| MRIME-CD | **3.17** | **2.08** | **2.63** |
| RIME | 5.83 | 6.42 | 6.13 |
| EO | 5.04 | 3.83 | 4.44 |
| SAO | 4.21 | 3.75 | 3.98 |
| ACGRIME | 6.25 | 6.08 | 6.17 |
| IRIME | 4.33 | 4.67 | 4.50 |
| TERIME | 9.00 | 9.00 | 9.00 |
| EOSMA | 4.25 | 5.75 | 5.00 |
| RDGMVO | 5.33 | 5.08 | 5.21 |
| MTVSCA | 7.58 | 8.33 | 7.96 |
| P-value | 3.86E-05 | 4.08E-08 | N/A |

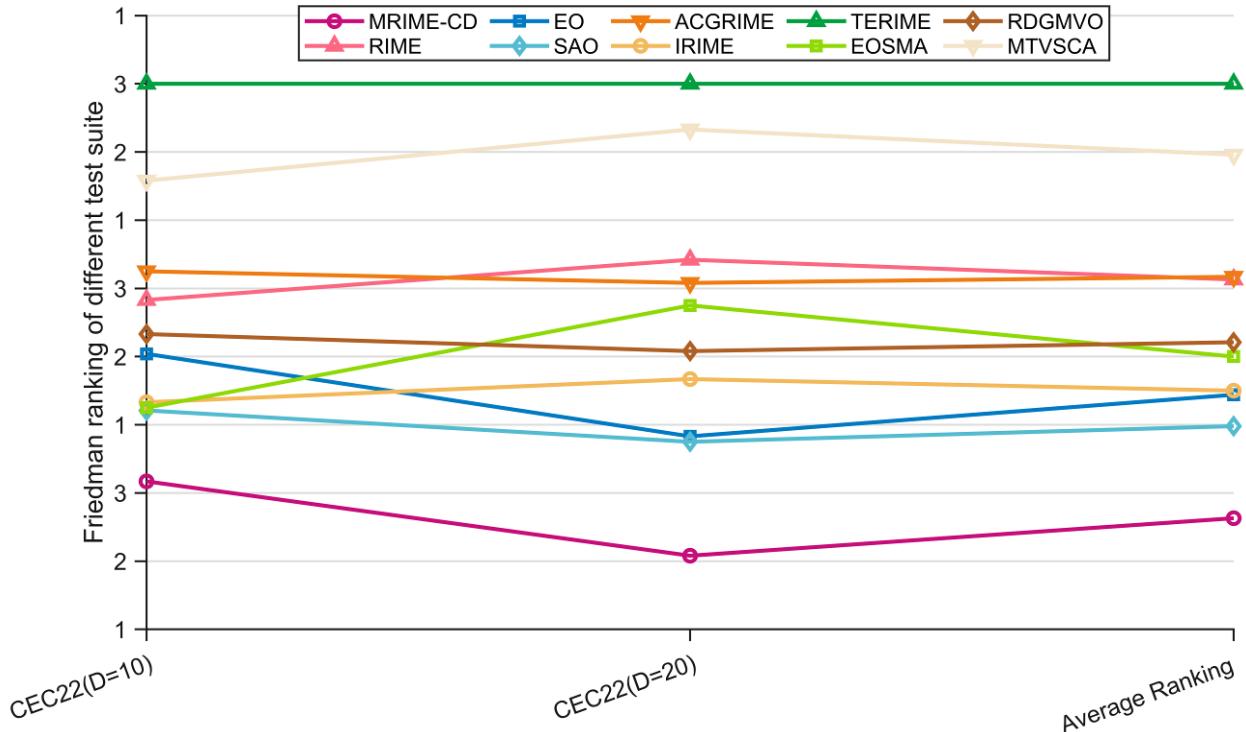

**Figure 13.** Friedman ranking of MRIME-CD and other competitors based on CEC2022

## 5. Engineering Design Optimization

We confirm the superior performance of MRIME-CD on test functions in Section 4. In this subsection, we evaluate the potential of MRIME-CD to solve real-world problems through 10 engineering constrained problems. The comparison algorithms are all the algorithms involved in the function testing. Table 15 shows the details of these engineering constrained optimization problems. In this paper, the penalty function method is used to transform the constrained optimization problems into unconstrained optimization problems. Table 16 shows the results of those 30 independent runs of MRIME-CD and the comparison algorithm. According to Table 16, MRIME-CD ranks first in 8 out of 10 constrained problems, with the worst ranking being 4th. In contrast, basic RIME's best ranking was 5th for all problems except C09. This significant difference illustrates the superior performance of

MRIME-CD. Compared to other algorithms, MRIME-CD consistently provides better design solutions. This reflects the fact that MRIME-CD maintains high solution accuracy when the design variables and constraints increase and has some engineering optimization capability. Therefore, the method can be effectively used for solving real engineering problems.

**Table 15.** Ten constrained engineering optimization problems

| Problem | Name | D |
|---|---|---|
| C01 | Tension/compression spring design problem | 3 |
| C02 | Pressure vessel design problem | 4 |
| C03 | Three-bar truss design problem | 2 |
| C04 | Welded beam design problem | 4 |
| C05 | Speed reducer design problem | 7 |
| C06 | Gear train design problem | 4 |
| C07 | Rolling element bearing design | 10 |
| C08 | Cantilever beam design problem | 5 |
| C09 | Multiple disk clutch brake design problem | 5 |
| C10 | Step-cone pulley problem | 5 |

**Table 16.** The statistics results of MRIME-CD and competitors solving constrained engineering optimization

| No. | Index | MRIME-CD | RIME | EO | SAO | ACGRIME | IRIME | TERIME | EOSMA | RDGMVO | MTVSCA |
|---|---|---|---|---|---|---|---|---|---|---|---|
| C01 | Best | 1.2686E-02 | 1.2806E-02 | 1.2761E-02 | 1.2753E-02 | 1.2667E-02 | 1.2798E-02 | 1.2841E-02 | 1.3387E-02 | 1.2686E-02 | 1.3540E-02 |
|  | Mean | 1.2898E-02 | 1.6084E-02 | 1.3094E-02 | 1.3224E-02 | 1.2947E-02 | 1.3193E-02 | 1.5628E-02 | 1.4791E-02 | 1.3363E-02 | 1.5338E-02 |
|  | Std | 3.3719E-04 | 2.3227E-03 | 3.1316E-04 | 5.6133E-04 | 2.9768E-04 | 5.4859E-04 | 3.0487E-03 | 8.9250E-04 | 6.8745E-04 | 1.2315E-03 |
|  | Rank | 1 | 10 | 3 | 5 | 2 | 4 | 9 | 7 | 6 | 8 |
| C02 | Best | 5.8842E+03 | 5.9673E+03 | 5.8716E+03 | 5.9601E+03 | 5.9113E+03 | 5.8812E+03 | 6.2803E+03 | 6.4373E+03 | 6.1766E+03 | 7.2526E+03 |
|  | Mean | 6.3527E+03 | 7.1599E+03 | 6.6048E+03 | 6.7324E+03 | 6.3867E+03 | 7.7121E+03 | 1.3082E+04 | 8.2614E+03 | 7.2371E+03 | 1.0186E+04 |
|  | Std | 4.0286E+02 | 1.1010E+03 | 4.7922E+02 | 5.9199E+02 | 3.8797E+02 | 1.0839E+03 | 7.4604E+03 | 1.0579E+03 | 7.0655E+02 | 1.6398E+03 |
|  | Rank | 1 | 5 | 3 | 4 | 2 | 7 | 10 | 8 | 6 | 9 |
| C03 | Best | 2.6389E+02 | 2.6389E+02 | 2.6389E+02 | 2.6389E+02 | 2.6389E+02 | 2.6389E+02 | 2.6389E+02 | 2.6390E+02 | 2.6389E+02 | 2.6395E+02 |
|  | Mean | 2.6389E+02 | 2.6430E+02 | 2.6394E+02 | 2.6408E+02 | 2.6395E+02 | 2.6397E+02 | 2.6412E+02 | 2.6402E+02 | 2.6403E+02 | 2.6418E+02 |
|  | Std | 1.9735E-03 | 6.5148E-01 | 9.1006E-02 | 1.7919E-01 | 1.0302E-01 | 1.0149E-01 | 3.1244E-01 | 7.6047E-02 | 1.6967E-01 | 2.1955E-01 |
|  | Rank | 1 | 10 | 2 | 7 | 3 | 4 | 8 | 5 | 6 | 9 |
| C04 | Best | 1.6939E+00 | 1.7058E+00 | 1.6975E+00 | 1.7204E+00 | 1.6953E+00 | 1.7159E+00 | 1.7563E+00 | 1.7590E+00 | 1.7119E+00 | 1.7364E+00 |
|  | Mean | 1.6998E+00 | 1.9304E+00 | 1.7641E+00 | 1.7698E+00 | 1.7507E+00 | 1.8112E+00 | 1.9837E+00 | 1.8287E+00 | 1.8226E+00 | 1.9557E+00 |
|  | Std | 5.1207E-03 | 2.2747E-01 | 6.2162E-02 | 2.1957E-02 | 6.0506E-02 | 7.4021E-02 | 1.9594E-01 | 4.8873E-02 | 8.3563E-02 | 9.6995E-02 |
|  | Rank | 1 | 8 | 3 | 4 | 2 | 5 | 10 | 7 | 6 | 9 |
| C05 | Best | 2.9936E+03 | 2.9947E+03 | 2.9939E+03 | 3.0015E+03 | 2.9936E+03 | 2.9939E+03 | 2.9958E+03 | 3.0087E+03 | 2.9953E+03 | 3.0231E+03 |
|  | Mean | 2.9936E+03 | 3.0080E+03 | 2.9971E+03 | 3.0091E+03 | 2.9943E+03 | 2.9958E+03 | 3.0228E+03 | 3.0243E+03 | 2.9984E+03 | 3.0485E+03 |
|  | Std | 3.4378E-07 | 1.0740E+01 | 4.1898E+00 | 5.1786E+00 | 1.0276E+00 | 1.4928E+00 | 2.2621E+01 | 8.1631E+00 | 2.9692E+00 | 1.2140E+01 |
|  | Rank | 1 | 6 | 4 | 7 | 2 | 3 | 8 | 9 | 5 | 10 |
| C06 | Best | 2.7009E-12 | 2.3078E-11 | 2.7009E-12 | 2.3078E-11 | 2.7009E-12 | 2.3078E-11 | 1.1661E-10 | 2.7009E-12 | 2.7009E-12 | 2.7009E-12 |
|  | Mean | 6.6817E-10 | 2.0376E-09 | 6.2012E-10 | 2.9446E-09 | 1.1767E-09 | 1.2449E-09 | 4.9188E-09 | 9.3329E-10 | 1.1702E-09 | 1.1446E-09 |
|  | Std | 7.5292E-10 | 1.9835E-09 | 6.3108E-10 | 4.5762E-09 | 1.2514E-09 | 1.5828E-09 | 7.9508E-09 | 1.3301E-09 | 1.2587E-09 | 8.5329E-10 |
|  | Rank | 2 | 8 | 1 | 9 | 6 | 7 | 10 | 3 | 5 | 4 |
| C07 | Best | -2.4358E+05 | -2.4358E+05 | -2.4358E+05 | -2.4358E+05 | -2.4358E+05 | -2.4358E+05 | -2.4358E+05 | -2.4306E+05 | -2.4358E+05 | -2.3943E+05 |
|  | Mean | -2.4358E+05 | -2.4296E+05 | -2.4358E+05 | -2.4358E+05 | -2.4358E+05 | -2.4358E+05 | -2.3333E+05 | -2.4158E+05 | -2.4358E+05 | -2.3146E+05 |
|  | Std | 8.9785E-11 | 1.5111E+03 | 1.0110E-04 | 2.7201E+00 | 8.5587E-08 | 2.6110E-09 | 8.5770E+03 | 8.2999E+02 | 1.8986E-04 | 3.6384E+03 |

|     |      | 1 | 7 | 4 | 6 | 3 | 2 | 9 | 8 | 5 | 10 |
|-----|------|---|---|---|---|---|---|---|---|---|----|
|     | Rank | 1 | 7 | 4 | 6 | 3 | 2 | 9 | 8 | 5 | 10 |
|     | Best | 1.3400E+00 | 1.3506E+00 | 1.3425E+00 | 1.3441E+00 | 1.3401E+00 | 1.3423E+00 | 1.3863E+00 | 1.3575E+00 | 1.3547E+00 | 1.3981E+00 |
| C08 | Mean | 1.3403E+00 | 1.3811E+00 | 1.3626E+00 | 1.3536E+00 | 1.3410E+00 | 1.3746E+00 | 1.6686E+00 | 1.4119E+00 | 1.4288E+00 | 1.4971E+00 |
|     | Std  | 3.0664E-04 | 4.1773E-02 | 1.7616E-02 | 6.0412E-03 | 1.0373E-03 | 3.1354E-02 | 1.9682E-01 | 3.2536E-02 | 5.2044E-02 | 4.7468E-02 |
|     | Rank | 1 | 6 | 4 | 3 | 2 | 5 | 10 | 7.0000E+00 | 8.0000E+00 | 9.0000E+00 |
|     | Best | 3.9247E+08 | 3.9247E+08 | 3.9247E+08 | 3.9247E+08 | 3.9247E+08 | 3.9247E+08 | 3.9247E+08 | 3.9247E+08 | 3.9247E+08 | 3.9247E+08 |
| C09 | Mean | 3.9247E+08 | 3.9247E+08 | 3.9247E+08 | 3.9247E+08 | 3.9247E+08 | 3.9247E+08 | 3.9247E+08 | 3.9247E+08 | 3.9247E+08 | 3.9247E+08 |
|     | Std  | 1.8187E-07 | 1.8187E-07 | 1.8187E-07 | 1.8187E-07 | 1.8187E-07 | 1.8187E-07 | 1.8187E-07 | 2.6961E-01 | 1.8187E-07 | 1.1921E+03 |
|     | Rank | 1 | 1 | 1 | 1 | 1 | 1 | 1 | 9 | 1 | 10 |
|     | Best | 1.6564E+01 | 1.6425E+01 | 1.6219E+01 | 1.8623E+01 | 1.6304E+01 | 1.6130E+01 | 1.6653E+01 | 1.6818E+01 | 1.6251E+01 | 1.7999E+01 |
| C10 | Mean | 1.7028E+01 | 1.7076E+01 | 1.6695E+01 | 2.9509E+01 | 1.6644E+01 | 1.6873E+01 | 2.1710E+01 | 1.7825E+01 | 1.7429E+01 | 6.7637E+01 |
|     | Std  | 1.4662E-01 | 2.9625E-01 | 2.6698E-01 | 1.0902E+01 | 1.8756E-01 | 3.6066E-01 | 8.4008E+00 | 7.2802E-01 | 4.7639E-01 | 4.1360E+01 |
|     | Rank | 4 | 5 | 2 | 9 | 1 | 3 | 8 | 7 | 6 | 10 |

## 6. Conclusions

This work proposed a modified RIME algorithm (MRIME-CD) with three efficient optimization approaches that mitigate the shortcomings of the basic RIME, such as the discrepancies between the global and local search phases and being prone to the local optimal areas. First, a covariance learning strategy is introduced in the soft-rime search stage to increase the population diversity and balance the over-exploitation ability of RIME through the bootstrapping effect of dominant populations. Second, in order to moderate the tendency of RIME population to approach the optimal individual in the early search stage, an average bootstrapping strategy is introduced into the hard-rime puncture mechanism, which guides the population search through the weighted position of the dominant populations, thus enhancing the global search ability of RIME in the early stage. Finally, a new stagnation indicator is proposed, and a stochastic covariance learning strategy is used to update the stagnant individuals in the population when the algorithm gets stagnant, thus enhancing the ability to jump out of the local optimal solution. To evaluate the performance of MRIME-CD, a total of 42 benchmark functions from the CEC2017 and CEC2022 test suites were used and 2 basic algorithms, 3 RIME variants and 3 superior algorithms were selected for comparison. A comprehensive analysis of the experimental results shows that the optimization persistence and convergence accuracy of MRIME-CD is significantly better than the other compared algorithms in dealing with complex and challenging problems, thus ensuring the feasibility and robustness of the proposed method. On the other hand, although MRIME-CD achieved the best overall performance, it still has room for improvement in some functions. Since the computation of the covariance matrix is more time-consuming, MRIME-CD is currently more suitable for optimization problems that do not require high computation time and it is necessary to consider parallel computing in the next research phase to speed up the program. In addition, the parameter sensitivity of MRIME-CD can be further investigated. In conclusion, MRIME-CD is a promising variant of the metaheuristic algorithm.

In the subsequent research, MRIME-CD can be considered to be applied to more complex engineering practical applications, such as engineering structure optimization, task planning and image segmentation problems, to further explore the superior performance of the metaheuristic algorithms for solving practical engineering problems. Although the proposed MRIME-CD method improves optimization accuracy, solution stability and convergence speed, there is still room for improvement in terms of computational cost and algorithmic complexity. In addition, we will develop multi-objective and binary versions of MRIME-CD to solve more practical problems.

**Competing Interest**

The authors declare that the authors have no competing interests as defined by Nature Research, or other interests that might be perceived to influence the results and/or discussion reported in this paper.

**Data Availability Statement**

The data is provided within the manuscript.

**Conflicts of Interest**

The authors declare no conflict of interest.

## Author contributions

**Shangqing Shi**: conceptualization, methodology, writing, data testing, reviewing, software. **Luoxiao Zhang:** methodology, writing, data testing, reviewing, software. **Yuchen Yin**: methodology, writing, reviewing. **Xiong Yang**: conceptualization, supervision, formal analysis. **Hoileong Lee:** reviewing, formal analysis.

## Appendix A

**Table A1.** Quantitative results of MRIME-CD and other competitors based on CEC2017 (D=10)

| No. | Metric | MRIME-CD | RIME | EO | SAO | ACGRIME | IRIME | TERIME | EOSMA | RDGMVO | MTVSCA |
|---|---|---|---|---|---|---|---|---|---|---|---|
| F1 | Best | 1.0000E+02 | 2.2296E+03 | 1.0282E+02 | 1.0137E+02 | 1.0009E+02 | 3.9764E+02 | 1.3846E+06 | 4.0425E+03 | 1.0039E+02 | 1.5439E+02 |
| | Ave | **1.0001E+02** | 1.2427E+04 | 3.5014E+03 | 3.1216E+03 | 3.7851E+03 | 7.6173E+03 | 6.7457E+06 | 2.6185E+04 | 4.4503E+03 | 1.3453E+03 |
| | Std | 7.7362E-03 | 6.9520E+03 | 3.0637E+03 | 3.1181E+03 | 3.9552E+03 | 7.4544E+03 | 3.6455E+06 | 1.9043E+04 | 5.2449E+03 | 1.0074E+03 |
| | Rank | 1 | 8 | 4 | 3 | 5 | 7 | 10 | 9 | 6 | 2 |
| F2 | Best | 3.0000E+02 | 3.0005E+02 | 3.0000E+02 | 3.0000E+02 | 3.0000E+02 | 3.0026E+02 | 3.5032E+02 | 3.0019E+02 | 3.0000E+02 | 3.0017E+02 |
| | Ave | 3.0000E+02 | 3.0034E+02 | 3.0001E+02 | **3.0000E+02** | 3.9898E+02 | 3.0634E+02 | 6.2869E+02 | 3.0324E+02 | 3.0000E+02 | 3.0548E+02 |
| | Std | 1.6265E-08 | 2.6731E-01 | 3.1326E-02 | 9.6334E-10 | 1.3251E+02 | 1.0393E+01 | 2.4226E+02 | 2.5528E+00 | 4.3576E-04 | 8.5137E+00 |
| | Rank | 2 | 5 | 4 | 1 | 9 | 8 | 10 | 6 | 3 | 7 |
| F3 | Best | 4.0000E+02 | 4.0001E+02 | 4.0017E+02 | 4.0019E+02 | 4.0023E+02 | 4.0181E+02 | 4.0370E+02 | 4.0009E+02 | 4.0001E+02 | 4.0023E+02 |
| | Ave | **4.0001E+02** | 4.0533E+02 | 4.0527E+02 | 4.0283E+02 | 4.0647E+02 | 4.0445E+02 | 4.1302E+02 | 4.0313E+02 | 4.1006E+02 | 4.0406E+02 |
| | Std | 2.3541E-02 | 2.2618E+00 | 8.1423E-01 | 8.0248E-01 | 1.8743E+00 | 7.6112E-01 | 1.9685E+01 | 1.2259E+00 | 2.0421E+01 | 1.2248E+00 |
| | Rank | 1 | 7 | 6 | 2 | 8 | 5 | 10 | 3 | 9 | 4 |
| F4 | Best | 5.0099E+02 | 5.0200E+02 | 5.0298E+02 | 5.0298E+02 | 5.0497E+02 | 5.0199E+02 | 5.1036E+02 | 5.0366E+02 | 5.0398E+02 | 5.1254E+02 |
| | Ave | **5.0564E+02** | 5.1271E+02 | 5.1067E+02 | 5.1021E+02 | 5.1195E+02 | 5.0962E+02 | 5.2159E+02 | 5.0827E+02 | 5.1448E+02 | 5.2142E+02 |
| | Std | 2.4181E+00 | 4.8798E+00 | 5.9762E+00 | 4.6662E+00 | 4.3528E+00 | 3.9932E+00 | 7.6898E+00 | 3.0709E+00 | 5.6540E+00 | 4.4677E+00 |
| | Rank | 1 | 7 | 5 | 4 | 6 | 3 | 10 | 2 | 8 | 9 |
| F5 | Best | 6.0000E+02 | 6.0005E+02 | 6.0000E+02 | 6.0000E+02 | 6.0000E+02 | 6.0000E+02 | 6.0045E+02 | 6.0000E+02 | 6.0000E+02 | 6.0001E+02 |
| | Ave | 6.0000E+02 | 6.0014E+02 | 6.0000E+02 | 6.0000E+02 | 6.0005E+02 | 6.0003E+02 | 6.0450E+02 | 6.0001E+02 | 6.0002E+02 | 6.0003E+02 |
| | Std | 5.2952E-05 | 6.7672E-02 | 1.8696E-04 | 7.7106E-14 | 8.6615E-02 | 1.5765E-02 | 2.7605E+00 | 8.5903E-03 | 1.7374E-02 | 1.8152E-02 |
| | Rank | 2 | 9 | 3 | 1 | 8 | 6 | 10 | 4 | 5 | 7 |
| F6 | Best | 7.0449E+02 | 7.0491E+02 | 7.1205E+02 | 7.1197E+02 | 7.1399E+02 | 7.1248E+02 | 7.2792E+02 | 7.1411E+02 | 7.1434E+02 | 7.2388E+02 |
| | Ave | **7.1526E+02** | 7.2298E+02 | 7.2069E+02 | 7.1830E+02 | 7.2586E+02 | 7.1988E+02 | 7.4922E+02 | 7.2047E+02 | 7.2514E+02 | 7.3470E+02 |
| | Std | 2.7373E+00 | 5.7474E+00 | 4.6727E+00 | 4.3390E+00 | 7.2713E+00 | 3.7315E+00 | 9.6415E+00 | 3.5529E+00 | 6.3560E+00 | 4.3935E+00 |
| | Rank | 1 | 6 | 5 | 2 | 8 | 3 | 10 | 4 | 7 | 9 |
| F7 | Best | 8.0099E+02 | 8.0498E+02 | 8.0398E+02 | 8.0298E+02 | 8.0398E+02 | 8.0200E+02 | 8.0578E+02 | 8.0110E+02 | 8.0398E+02 | 8.1164E+02 |
| | Ave | **8.0546E+02** | 8.1452E+02 | 8.1024E+02 | 8.1001E+02 | 8.1198E+02 | 8.0902E+02 | 8.2209E+02 | 8.0826E+02 | 8.1471E+02 | 8.2269E+02 |
| | Std | 2.2458E+00 | 5.0645E+00 | 4.4107E+00 | 4.6259E+00 | 4.1405E+00 | 3.9319E+00 | 7.8971E+00 | 2.7077E+00 | 5.4070E+00 | 4.3882E+00 |
| | Rank | 1 | 7 | 5 | 4 | 6 | 3 | 9 | 2 | 8 | 10 |
| F8 | Best | 9.0000E+02 | 9.0000E+02 | 9.0000E+02 | 9.0000E+02 | 9.0000E+02 | 9.0000E+02 | 9.0477E+02 | 9.0000E+02 | 9.0000E+02 | 9.0000E+02 |
| | Ave | 9.0000E+02 | 9.0010E+02 | 9.0005E+02 | **9.0000E+02** | 9.0006E+02 | 9.0000E+02 | 9.3906E+02 | 9.0001E+02 | 9.0004E+02 | 9.0000E+02 |
| | Std | 1.2536E-02 | 3.2365E-01 | 1.6390E-01 | 3.5951E-14 | 1.4376E-01 | 3.1652E-03 | 4.6403E+01 | 2.4312E-02 | 1.1634E-01 | 1.6566E-03 |
| | Rank | 3 | 9 | 7 | 1 | 8 | 4 | 10 | 5 | 6 | 2 |
| F9 | Best | 1.0118E+03 | 1.1227E+03 | 1.0120E+03 | 1.0035E+03 | 1.0102E+03 | 1.0319E+03 | 1.2612E+03 | 1.0346E+03 | 1.1220E+03 | 1.6920E+03 |
| | Ave | **1.3190E+03** | 1.4567E+03 | 1.4620E+03 | 1.4988E+03 | 1.4193E+03 | 1.3334E+03 | 1.7458E+03 | 1.6021E+03 | 1.5071E+03 | 2.0296E+03 |
| | Std | 2.4021E+02 | 2.4824E+02 | 2.4082E+02 | 2.7035E+02 | 1.9864E+02 | 1.5951E+02 | 2.4913E+02 | 2.3168E+02 | 2.3504E+02 | 1.4610E+02 |
| | Rank | 1 | 4 | 5 | 6 | 3 | 2 | 9 | 8 | 7 | 10 |
| F10 | Best | 1.1000E+03 | 1.1029E+03 | 1.1011E+03 | 1.1001E+03 | 1.1024E+03 | 1.1005E+03 | 1.1046E+03 | 1.1010E+03 | 1.1021E+03 | 1.1034E+03 |
| | Ave | **1.1034E+03** | 1.1107E+03 | 1.1052E+03 | 1.1051E+03 | 1.1106E+03 | 1.1055E+03 | 1.1335E+03 | 1.1038E+03 | 1.1168E+03 | 1.1073E+03 |
| | Std | 2.7351E+00 | 6.7096E+00 | 3.8663E+00 | 3.9185E+00 | 5.7203E+00 | 2.9203E+00 | 4.9079E+01 | 1.2449E+00 | 3.4874E+01 | 1.8638E+00 |
| | Rank | 1 | 8 | 4 | 3 | 7 | 5 | 10 | 2 | 9 | 6 |
| F11 | Best | 1.2002E+03 | 5.8867E+03 | 1.7521E+03 | 1.7961E+03 | 6.3079E+03 | 3.1128E+03 | 4.2830E+04 | 6.1770E+03 | 2.5164E+03 | 1.4954E+03 |
| | Ave | **1.4800E+03** | 8.5822E+04 | 1.2427E+04 | 1.2736E+04 | 1.2927E+05 | 2.3021E+04 | 3.3438E+06 | 1.9849E+04 | 2.1853E+04 | 6.1901E+03 |
| | Std | 1.5718E+02 | 2.0772E+05 | 9.2080E+03 | 1.2006E+04 | 1.5819E+05 | 1.9090E+04 | 3.6891E+06 | 8.9265E+03 | 2.4757E+04 | 4.5070E+03 |
| | Rank | 1 | 8 | 3 | 4 | 9 | 7 | 10 | 5 | 6 | 2 |
| F12 | Best | 1.3000E+03 | 1.3206E+03 | 1.3855E+03 | 1.3739E+03 | 1.3270E+03 | 1.3140E+03 | 1.3487E+03 | 1.3347E+03 | 1.3158E+03 | 1.3037E+03 |
| | Ave | **1.3051E+03** | 1.0938E+04 | 6.1471E+03 | 1.1369E+04 | 8.7038E+03 | 1.4274E+03 | 1.4536E+04 | 1.3733E+03 | 1.0520E+04 | 1.3170E+03 |
| | Std | 3.1423E+00 | 1.0714E+04 | 5.8584E+03 | 9.9014E+03 | 8.0886E+03 | 1.7736E+02 | 1.1671E+04 | 2.0220E+01 | 9.5558E+03 | 5.2323E+00 |
| | Rank | 1 | 8 | 5 | 9 | 6 | 4 | 10 | 3 | 7 | 2 |
| F13 | Best | 1.4000E+03 | 1.4067E+03 | 1.4253E+03 | 1.4033E+03 | 1.4041E+03 | 1.4021E+03 | 1.4222E+03 | 1.4115E+03 | 1.4053E+03 | 1.4067E+03 |
| | Ave | **1.4084E+03** | 2.7141E+03 | 1.4772E+03 | 3.8933E+03 | 1.7593E+03 | 1.4195E+03 | 1.5369E+03 | 1.4271E+03 | 2.0991E+03 | 1.4169E+03 |
| | Std | 9.5827E+00 | 2.4121E+03 | 2.7124E+01 | 2.4633E+03 | 4.7511E+02 | 8.9170E+00 | 1.6672E+02 | 4.3413E+00 | 1.1483E+03 | 5.4087E+00 |
| | Rank | 1 | 9 | 5 | 10 | 7 | 3 | 6 | 4 | 8 | 2 |
| F14 | Best | 1.5000E+03 | 1.5056E+03 | 1.5242E+03 | 1.5021E+03 | 1.5022E+03 | 1.5024E+03 | 1.5172E+03 | 1.5040E+03 | 1.5020E+03 | 1.5019E+03 |
| | Ave | **1.5016E+03** | 3.5305E+03 | 1.6351E+03 | 3.0047E+03 | 1.7814E+03 | 1.5073E+03 | 2.4642E+03 | 1.5172E+03 | 2.1574E+03 | 1.5043E+03 |
| | Std | 1.1692E+00 | 3.6588E+03 | 9.1230E+01 | 2.5091E+03 | 7.0479E+02 | 3.4015E+00 | 1.2667E+03 | 5.7661E+00 | 1.1270E+03 | 1.8295E+00 |
| | Rank | 1 | 10 | 5 | 9 | 6 | 3 | 8 | 4 | 7 | 2 |
| F15 | Best | 1.6000E+03 | 1.6017E+03 | 1.6007E+03 | 1.6003E+03 | 1.6005E+03 | 1.6003E+03 | 1.6033E+03 | 1.6016E+03 | 1.6003E+03 | 1.6057E+03 |
| | Ave | 1.6297E+03 | 1.6903E+03 | 1.6470E+03 | 1.6978E+03 | 1.6428E+03 | 1.6206E+03 | 1.6988E+03 | **1.6085E+03** | 1.6933E+03 | 1.6158E+03 |
| | Std | 5.0351E+01 | 8.0006E+01 | 5.9705E+01 | 1.1398E+02 | 5.9035E+01 | 3.6806E+01 | 1.0527E+02 | 7.1407E+00 | 1.0219E+02 | 8.8469E+00 |
| | Rank | 4 | 7 | 6 | 9 | 5 | 3 | 10 | 1 | 8 | 2 |
| F16 | Best | 1.7003E+03 | 1.7017E+03 | 1.7005E+03 | 1.7017E+03 | 1.7012E+03 | 1.7008E+03 | 1.7143E+03 | 1.7210E+03 | 1.7010E+03 | 1.7293E+03 |
| | Ave | 1.7159E+03 | 1.7433E+03 | 1.7373E+03 | 1.7316E+03 | 1.7226E+03 | **1.7071E+03** | 1.7485E+03 | 1.7345E+03 | 1.7391E+03 | 1.7403E+03 |
| | Std | 1.4042E+01 | 4.2034E+01 | 1.9365E+01 | 1.6438E+01 | 1.1237E+01 | 7.0989E+00 | 2.8371E+01 | 6.6619E+00 | 3.6793E+01 | 4.8408E+00 |
| | Rank | 2 | 9 | 6 | 4 | 3 | 1 | 10 | 5 | 7 | 8 |
| F17 | Best | 1.8000E+03 | 1.9188E+03 | 3.2653E+03 | 2.1405E+03 | 2.0427E+03 | 1.8441E+03 | 2.3320E+03 | 1.8518E+03 | 1.8499E+03 | 1.8101E+03 |
| | Ave | **1.8110E+03** | 9.4804E+03 | 1.7566E+04 | 1.9976E+04 | 1.0281E+04 | 2.2362E+03 | 1.9660E+04 | 1.9084E+03 | 1.5106E+04 | 1.8190E+03 |
| | Std | 1.0075E+01 | 8.1194E+03 | 1.0775E+04 | 1.4300E+04 | 7.7241E+03 | 5.2130E+02 | 1.3419E+04 | 3.2322E+01 | 1.1508E+04 | 5.4050E+00 |
| | Rank | 1 | 5 | 8 | 10 | 6 | 4 | 9 | 3 | 7 | 2 |
| F18 | Best | 1.9000E+03 | 1.9020E+03 | 1.9067E+03 | 1.9098E+03 | 1.9039E+03 | 1.9019E+03 | 1.9076E+03 | 1.9054E+03 | 1.9012E+03 | 1.9020E+03 |
| | Ave | **1.9006E+03** | 4.0541E+03 | 2.0145E+03 | 6.9722E+03 | 2.7326E+03 | 1.9045E+03 | 4.7967E+03 | 1.9092E+03 | 4.2695E+03 | 1.9033E+03 |
| | Std | 8.3542E-01 | 4.2828E+03 | 1.7737E+02 | 7.4284E+03 | 1.7456E+03 | 2.2473E+00 | 4.9732E+03 | 2.1880E+00 | 3.4058E+03 | 6.8311E-01 |
| | Rank | 1 | 7 | 5 | 10 | 6 | 3 | 9 | 4 | 8 | 2 |
| F19 | Best | 2.0000E+03 | 2.0009E+03 | 2.0003E+03 | 2.0000E+03 | 2.0007E+03 | 2.0000E+03 | 2.0130E+03 | 2.0020E+03 | 2.0000E+03 | 2.0118E+03 |
| | Ave | 2.0077E+03 | 2.0171E+03 | 2.0340E+03 | 2.0400E+03 | 2.0126E+03 | **2.0008E+03** | 2.0451E+03 | 2.0235E+03 | 2.0113E+03 | 2.0278E+03 |
| | Std | 2.0623E+01 | 2.0629E+01 | 3.9607E+01 | 4.6276E+01 | 1.2295E+01 | 9.3214E-01 | 1.3694E+01 | 8.7036E+00 | 1.1302E+01 | 7.5409E+00 |
| | Rank | 2 | 5 | 8 | 9 | 4 | 1 | 10 | 6 | 3 | 7 |
| F20 | Best | 2.2000E+03 | 2.2000E+03 | 2.2000E+03 | 2.2000E+03 | 2.2008E+03 | 2.2000E+03 | 2.2020E+03 | 2.2001E+03 | 2.2000E+03 | 2.2001E+03 |
| | Ave | 2.2931E+03 | 2.2625E+03 | 2.2990E+03 | 2.2731E+03 | 2.2402E+03 | 2.2569E+03 | **2.2098E+03** | 2.2308E+03 | 2.2827E+03 | 2.2662E+03 |
| | Std | 3.7236E+01 | 5.8780E+01 | 3.6786E+01 | 5.4183E+01 | 5.2920E+01 | 5.7158E+01 | 2.2043E+01 | 4.8807E+01 | 5.6682E+01 | 5.8323E+01 |

|  | Rank | 9 | 5 | 10 | 7 | 3 | 4 | 1 | 2 | 8 | 6 |
|---|---|---|---|---|---|---|---|---|---|---|---|
| F21 | Best | 2.2198E+03 | 2.2272E+03 | 2.2192E+03 | 2.2156E+03 | 2.2192E+03 | 2.2000E+03 | 2.2302E+03 | 2.2000E+03 | 2.2270E+03 | 2.2002E+03 |
|  | Ave | 2.2962E+03 | 2.3018E+03 | 2.2990E+03 | 2.2980E+03 | 2.3004E+03 | 2.2964E+03 | 2.2973E+03 | **2.2804E+03** | 2.3184E+03 | 2.2928E+03 |
|  | Std | 1.8290E+01 | 1.0765E+01 | 1.1410E+01 | 1.6816E+01 | 1.1649E+01 | 2.2457E+01 | 2.6771E+01 | 3.6256E+01 | 1.2436E+02 | 2.7928E+01 |
|  | Rank | 3 | 9 | 7 | 6 | 8 | 4 | 5 | 1 | 10 | 2 |
| F22 | Best | 2.6029E+03 | 2.6065E+03 | 2.6051E+03 | 2.6044E+03 | 2.6059E+03 | 2.6071E+03 | 2.3050E+03 | 2.6002E+03 | 2.6070E+03 | 2.3378E+03 |
|  | Ave | **2.6095E+03** | 2.6183E+03 | 2.6142E+03 | 2.6134E+03 | 2.6162E+03 | 2.6135E+03 | 2.6202E+03 | 2.6102E+03 | 2.6218E+03 | 2.6135E+03 |
|  | Std | 4.7283E+00 | 6.3379E+00 | 6.3038E+00 | 5.6027E+00 | 5.0488E+00 | 3.7018E+00 | 4.5780E+01 | 4.1213E+00 | 8.8675E+00 | 3.9569E+01 |
|  | Rank | 1 | 8 | 6 | 3 | 7 | 5 | 9 | 2 | 10 | 4 |
| F23 | Best | 2.5000E+03 | 2.5001E+03 | 2.5000E+03 | 2.5000E+03 | 2.5000E+03 | 2.5000E+03 | 2.5034E+03 | 2.5001E+03 | 2.5000E+03 | 2.4214E+03 |
|  | Ave | 2.7206E+03 | 2.7306E+03 | 2.7367E+03 | 2.7385E+03 | 2.7055E+03 | 2.7328E+03 | 2.6927E+03 | **2.6337E+03** | 2.7346E+03 | 2.7337E+03 |
|  | Std | 6.5295E+01 | 6.8370E+01 | 3.4481E+01 | 3.4673E+01 | 9.2096E+01 | 5.9025E+01 | 1.0855E+02 | 1.1434E+02 | 6.9904E+01 | 5.7894E+01 |
|  | Rank | 4 | 5 | 9 | 10 | 3 | 6 | 2 | 1 | 8 | 7 |
| F24 | Best | 2.8977E+03 | 2.8978E+03 | 2.8978E+03 | 2.8977E+03 | 2.8977E+03 | 2.8977E+03 | 2.8988E+03 | 2.8977E+03 | 2.8979E+03 | 2.8977E+03 |
|  | Ave | 2.9213E+03 | 2.9287E+03 | 2.9364E+03 | 2.9358E+03 | 2.9268E+03 | 2.9148E+03 | 2.9361E+03 | **2.8986E+03** | 2.9310E+03 | 2.9157E+03 |
|  | Std | 2.3168E+01 | 2.3330E+01 | 1.8907E+01 | 1.9773E+01 | 2.2698E+01 | 2.2510E+01 | 2.5844E+01 | 6.2381E-01 | 2.3056E+01 | 2.2508E+01 |
|  | Rank | 4 | 6 | 10 | 8 | 5 | 2 | 9 | 1 | 7 | 3 |
| F25 | Best | 2.6000E+03 | 2.8012E+03 | 2.9000E+03 | 2.6000E+03 | 2.8000E+03 | 2.9000E+03 | 2.9028E+03 | 2.6223E+03 | 2.6000E+03 | 2.9000E+03 |
|  | Ave | 2.9029E+03 | 2.9336E+03 | 2.9521E+03 | 2.9275E+03 | 2.9173E+03 | 2.9019E+03 | 2.9455E+03 | **2.8946E+03** | 2.9234E+03 | 2.9000E+03 |
|  | Std | 8.6103E+01 | 1.5725E+02 | 1.4498E+02 | 6.3318E+01 | 4.0435E+01 | 9.2308E+00 | 4.1706E+01 | 3.8889E+01 | 2.1142E+02 | 5.2212E-02 |
|  | Rank | 4 | 8 | 10 | 7 | 5 | 3 | 9 | 1 | 6 | 2 |
| F26 | Best | 3.0887E+03 | 3.0890E+03 | 3.0890E+03 | 3.0890E+03 | 3.0890E+03 | 3.0890E+03 | 3.0900E+03 | 3.0874E+03 | 3.0708E+03 | 3.0896E+03 |
|  | Ave | 3.0926E+03 | 3.0948E+03 | 3.0920E+03 | 3.0938E+03 | 3.0910E+03 | 3.0897E+03 | 3.0959E+03 | 3.0895E+03 | **3.0810E+03** | 3.0944E+03 |
|  | Std | 3.8027E+00 | 3.2079E+00 | 2.0989E+00 | 1.0294E+01 | 1.6182E+00 | 7.0890E-01 | 3.3172E+00 | 5.0982E-01 | 1.7231E+01 | 1.7269E+00 |
|  | Rank | 6 | 9 | 5 | 7 | 4 | 3 | 10 | 2 | 1 | 8 |
| F27 | Best | 3.1000E+03 | 3.1001E+03 | 3.1000E+03 | 3.1000E+03 | 3.1000E+03 | 3.1000E+03 | 3.1076E+03 | 2.8626E+03 | 3.1000E+03 | 3.1001E+03 |
|  | Ave | 3.2818E+03 | 3.2579E+03 | 3.3018E+03 | 3.2905E+03 | 3.2377E+03 | 3.1523E+03 | 3.2169E+03 | **3.1090E+03** | 3.2259E+03 | 3.1738E+03 |
|  | Std | 1.5495E+02 | 1.2949E+02 | 1.3766E+02 | 1.4370E+02 | 1.0839E+02 | 8.9441E+01 | 7.7548E+01 | 4.1800E+01 | 7.4150E+01 | 1.2368E+02 |
|  | Rank | 8 | 7 | 10 | 9 | 6 | 2 | 4 | 1 | 5 | 3 |
| F28 | Best | 3.1288E+03 | 3.1365E+03 | 3.1381E+03 | 3.1303E+03 | 3.1360E+03 | 3.1343E+03 | 3.1455E+03 | 3.1391E+03 | 3.1300E+03 | 3.1549E+03 |
|  | Ave | **3.1472E+03** | 3.1868E+03 | 3.1669E+03 | 3.1879E+03 | 3.1640E+03 | 3.1477E+03 | 3.2052E+03 | 3.1632E+03 | 3.2074E+03 | 3.1881E+03 |
|  | Std | 1.9850E+01 | 2.9729E+01 | 1.8461E+01 | 4.6056E+01 | 1.8075E+01 | 1.1041E+01 | 3.9697E+01 | 1.0078E+01 | 5.5473E+01 | 1.6509E+01 |
|  | Rank | 1 | 6 | 5 | 7 | 4 | 2 | 9 | 3 | 10 | 8 |
| F29 | Best | 3.3945E+03 | 3.9474E+03 | 4.1066E+03 | 3.8031E+03 | 3.6657E+03 | 3.6244E+03 | 4.0104E+03 | 4.2332E+03 | 3.2242E+03 | 3.6747E+03 |
|  | Ave | 3.0295E+05 | 2.4280E+05 | 4.6778E+05 | 3.9012E+05 | 9.4153E+04 | **7.8891E+03** | 2.9317E+05 | 1.0941E+04 | 1.2529E+04 | 6.3386E+04 |
|  | Std | 4.8322E+05 | 4.3619E+05 | 5.4234E+05 | 5.2545E+05 | 2.0215E+05 | 4.3694E+03 | 4.2992E+05 | 7.2485E+03 | 1.5381E+04 | 2.5760E+05 |
|  | Rank | 8 | 6 | 10 | 9 | 5 | 1 | 7 | 2 | 3 | 4 |

**Table A2.** Quantitative results of MRIME-CD and other competitors based on CEC2017 (D=30)

| No. | Metric | MRIME-CD | RIME | EO | SAO | ACGRIME | IRIME | TERIME | EOSMA | RDGMVO | MTVSCA |
|---|---|---|---|---|---|---|---|---|---|---|---|
| F1 | Best | 1.0010E+02 | 2.0064E+05 | 1.0349E+02 | 1.0515E+02 | 1.1902E+02 | 3.8289E+04 | 1.6466E+08 | 3.2494E+03 | 1.1104E+02 | 4.3550E+03 |
|  | Ave | **5.2321E+02** | 5.2338E+05 | 4.5407E+03 | 3.9519E+03 | 4.6691E+03 | 2.5086E+05 | 3.6175E+08 | 1.9535E+04 | 6.5473E+03 | 2.9694E+04 |
|  | Std | 1.2797E+03 | 2.3284E+05 | 6.0255E+03 | 4.2587E+03 | 5.2985E+03 | 1.5714E+05 | 1.5228E+08 | 1.3681E+04 | 6.2432E+03 | 1.7076E+04 |
|  | Rank | 1 | 9 | 3 | 2 | 4 | 8 | 10 | 6 | 5 | 7 |
| F2 | Best | 3.0000E+02 | 1.2438E+03 | 1.6770E+03 | 2.8682E+04 | 1.1360E+04 | 1.2224E+04 | 1.9307E+04 | 9.0249E+03 | 3.0006E+02 | 3.1969E+03 |
|  | Ave | **3.0000E+02** | 3.0961E+03 | 5.4917E+03 | 5.0930E+04 | 2.4622E+04 | 2.0881E+04 | 3.9592E+04 | 1.7020E+04 | 3.0076E+02 | 1.0088E+04 |
|  | Std | 9.3817E-07 | 1.1750E+03 | 2.3870E+03 | 1.2558E+04 | 6.5835E+03 | 5.8454E+03 | 1.0788E+04 | 3.8759E+03 | 7.7845E-01 | 3.6449E+03 |
|  | Rank | 1 | 3 | 4 | 10 | 8 | 7 | 9 | 6 | 2 | 5 |
| F3 | Best | 4.0001E+02 | 4.5996E+02 | 4.0418E+02 | 4.6957E+02 | 4.7364E+02 | 4.7183E+02 | 5.1943E+02 | 4.0740E+02 | 4.0714E+02 | 4.0146E+02 |
|  | Ave | 4.8160E+02 | 5.0294E+02 | 4.9100E+02 | 4.9079E+02 | 5.0968E+02 | 5.0126E+02 | 5.7611E+02 | 5.0315E+02 | **4.5483E+02** | 4.9893E+02 |
|  | Std | 3.2159E+01 | 1.8121E+01 | 2.7365E+01 | 1.2881E+01 | 1.4405E+01 | 1.6200E+01 | 3.7770E+01 | 1.9366E+01 | 3.6295E+01 | 2.6372E+01 |
|  | Rank | 2 | 7 | 4 | 3 | 9 | 6 | 10 | 8 | 1 | 5 |
| F4 | Best | 5.1393E+02 | 5.3603E+02 | 5.2898E+02 | 5.2089E+02 | 5.3681E+02 | 5.3205E+02 | 6.0499E+02 | 5.2271E+02 | 5.5174E+02 | 6.2448E+02 |
|  | Ave | **5.3301E+02** | 5.7023E+02 | 5.5884E+02 | 5.4127E+02 | 5.7139E+02 | 5.6273E+02 | 6.5597E+02 | 5.5615E+02 | 5.9503E+02 | 6.6037E+02 |
|  | Std | 1.0548E+01 | 1.6756E+01 | 1.8422E+01 | 1.1114E+01 | 1.8913E+01 | 1.5842E+01 | 2.7323E+01 | 1.5754E+01 | 2.5699E+01 | 1.1951E+01 |
|  | Rank | 1 | 6 | 4 | 2 | 7 | 5 | 9 | 3 | 8 | 10 |
| F5 | Best | 6.0000E+02 | 6.0072E+02 | 6.0000E+02 | 6.0000E+02 | 6.0027E+02 | 6.0013E+02 | 6.0672E+02 | 6.0016E+02 | 6.0024E+02 | 6.0013E+02 |
|  | Ave | 6.0014E+02 | 6.0181E+02 | 6.0003E+02 | **6.0001E+02** | 6.0096E+02 | 6.0027E+02 | 6.1193E+02 | 6.0050E+02 | 6.0101E+02 | 6.0028E+02 |
|  | Std | 2.3742E-01 | 7.5965E-01 | 1.1700E-01 | 4.5889E-02 | 7.1717E-01 | 9.7436E-02 | 2.8646E+00 | 2.6609E-01 | 7.6224E-01 | 1.1741E-01 |
|  | Rank | 3 | 9 | 2 | 1 | 7 | 4 | 10 | 6 | 8 | 5 |
| F6 | Best | 7.4106E+02 | 7.6878E+02 | 7.5958E+02 | 7.5053E+02 | 7.6987E+02 | 7.6531E+02 | 8.9335E+02 | 7.7193E+02 | 7.8624E+02 | 8.6140E+02 |
|  | Ave | **7.5922E+02** | 8.1511E+02 | 7.8528E+02 | 7.6452E+02 | 8.0823E+02 | 7.9351E+02 | 9.7453E+02 | 8.0433E+02 | 8.3328E+02 | 8.9736E+02 |
|  | Std | 9.4207E+00 | 2.1716E+01 | 1.6540E+01 | 1.0325E+01 | 2.3110E+01 | 1.3391E+01 | 3.6847E+01 | 1.8286E+01 | 2.5631E+01 | 1.5204E+01 |
|  | Rank | 1 | 7 | 3 | 2 | 6 | 4 | 10 | 5 | 8 | 9 |
| F7 | Best | 8.1691E+02 | 8.3447E+02 | 8.2289E+02 | 8.1791E+02 | 8.3980E+02 | 8.3484E+02 | 8.8790E+02 | 8.2773E+02 | 8.3482E+02 | 9.1796E+02 |
|  | Ave | **8.3154E+02** | 8.6976E+02 | 8.5722E+02 | 8.4275E+02 | 8.7427E+02 | 8.6254E+02 | 9.5224E+02 | 8.5468E+02 | 8.8469E+02 | 9.5256E+02 |
|  | Std | 7.6999E+00 | 1.5757E+01 | 1.6053E+01 | 1.2175E+01 | 1.8019E+01 | 1.3887E+01 | 2.5629E+01 | 1.3489E+01 | 2.3854E+01 | 1.1940E+01 |
|  | Rank | 1 | 6 | 4 | 2 | 7 | 5 | 9 | 3 | 8 | 10 |
| F8 | Best | 9.0027E+02 | 9.0535E+02 | 9.0000E+02 | 9.0000E+02 | 9.0459E+02 | 9.0111E+02 | 1.5761E+03 | 9.0129E+02 | 9.0481E+02 | 9.0043E+02 |
|  | Ave | 9.0609E+02 | 1.0875E+03 | 9.0541E+02 | **9.0274E+02** | 1.0538E+03 | 9.1397E+02 | 3.3249E+03 | 9.1079E+02 | 1.8461E+03 | 9.0334E+02 |
|  | Std | 5.1906E+00 | 2.1942E+02 | 1.1845E+01 | 5.8883E+00 | 2.1459E+02 | 1.9297E+01 | 1.5138E+03 | 6.1579E+00 | 8.0103E+02 | 5.9641E+00 |
|  | Rank | 4 | 8 | 3 | 1 | 7 | 6 | 10 | 5 | 9 | 2 |
| F9 | Best | 2.2539E+03 | 2.7464E+03 | 2.5502E+03 | 1.6641E+03 | 3.1336E+03 | 2.8132E+03 | 3.5097E+03 | 3.9834E+03 | 2.5149E+03 | 6.7009E+03 |
|  | Ave | 3.4969E+03 | 4.1949E+03 | 4.6265E+03 | **3.4492E+03** | 4.2035E+03 | 4.0020E+03 | 5.2683E+03 | 6.0789E+03 | 3.9264E+03 | 7.6255E+03 |
|  | Std | 6.1723E+02 | 6.2070E+02 | 8.9645E+02 | 6.7264E+02 | 6.8828E+02 | 4.9170E+02 | 5.5272E+02 | 5.7183E+02 | 5.3139E+02 | 3.5914E+02 |
|  | Rank | 2 | 5 | 7 | 1 | 6 | 4 | 8 | 9 | 3 | 10 |
| F10 | Best | 1.1179E+03 | 1.2018E+03 | 1.1154E+03 | 1.1054E+03 | 1.1722E+03 | 1.1329E+03 | 1.1901E+03 | 1.1417E+03 | 1.1272E+03 | 1.1466E+03 |
|  | Ave | 1.1617E+03 | 1.2782E+03 | 1.1569E+03 | **1.1406E+03** | 1.2730E+03 | 1.1961E+03 | 1.2761E+03 | 1.2040E+03 | 1.1674E+03 | 1.2173E+03 |
|  | Std | 3.1097E+01 | 4.7230E+01 | 2.8958E+01 | 2.8360E+01 | 4.9549E+01 | 2.9026E+01 | 4.3384E+01 | 2.8064E+01 | 2.2838E+01 | 3.2686E+01 |
|  | Rank | 3 | 10 | 2 | 1 | 8 | 5 | 9 | 6 | 4 | 7 |
| F11 | Best | 3.5843E+03 | 3.3268E+05 | 1.7935E+04 | 3.5935E+04 | 5.0690E+05 | 3.3586E+04 | 4.2907E+06 | 1.0712E+05 | 2.4394E+05 | 1.8434E+04 |
|  | Ave | **1.7951E+04** | 7.1057E+06 | 2.6002E+05 | 1.7609E+05 | 8.2844E+06 | 1.3768E+06 | 1.9435E+07 | 6.7642E+05 | 2.9066E+06 | 2.6285E+05 |
|  | Std | 1.2387E+04 | 6.5857E+06 | 2.6272E+05 | 1.1149E+05 | 8.3785E+06 | 1.4230E+06 | 1.1820E+07 | 5.1685E+05 | 2.4232E+06 | 2.8612E+05 |
|  | Rank | 1 | 8 | 3 | 2 | 9 | 6 | 10 | 5 | 7 | 4 |
| F12 | Best | 1.3187E+03 | 1.3851E+04 | 1.7794E+03 | 1.3323E+03 | 8.5182E+03 | 2.3193E+03 | 3.2901E+05 | 2.6322E+04 | 1.5448E+03 | 3.4016E+03 |
|  | Ave | **1.4084E+03** | 8.5497E+04 | 2.0858E+04 | 1.9085E+04 | 8.1345E+04 | 2.2839E+04 | 1.7124E+06 | 5.5892E+04 | 2.1690E+04 | 7.7119E+03 |
|  | Std | 3.2783E+02 | 5.4514E+04 | 1.9042E+04 | 1.7552E+04 | 4.4982E+04 | 2.1080E+04 | 1.4325E+06 | 1.8726E+04 | 2.5769E+04 | 4.5089E+03 |
|     | Rank | 1 | 9 | 4 | 3 | 8 | 6 | 10 | 7 | 5 | 2 |
|     | Best | 1.4080E+03 | 2.4408E+03 | 2.0777E+03 | 2.1577E+03 | 2.0973E+03 | 1.6385E+03 | 3.7116E+03 | 1.5679E+03 | 1.5732E+03 | 1.4632E+03 |
| F13 | Ave | **1.4279E+03** | 3.4527E+04 | 2.4153E+04 | 2.0568E+04 | 3.1812E+04 | 8.9119E+03 | 6.8317E+04 | 1.6573E+03 | 3.0740E+04 | 1.5032E+03 |
|     | Std | 4.3717E+00 | 2.8104E+04 | 2.2152E+04 | 1.4085E+04 | 3.3042E+04 | 8.3045E+03 | 5.3562E+04 | 4.6023E+01 | 2.7780E+04 | 2.0310E+01 |
|     | Rank | 1 | 9 | 6 | 5 | 8 | 4 | 10 | 3 | 7 | 2 |
|     | Best | 1.5131E+03 | 2.7229E+03 | 1.6478E+03 | 1.5227E+03 | 3.0038E+03 | 2.4228E+03 | 2.7930E+03 | 3.2798E+03 | 1.5456E+03 | 1.6424E+03 |
| F14 | Ave | **1.5565E+03** | 1.9722E+04 | 5.9462E+03 | 5.1157E+03 | 9.8548E+03 | 1.6706E+04 | 1.3645E+05 | 7.1623E+03 | 1.1160E+04 | 1.7383E+03 |
|     | Std | 4.1484E+01 | 1.2498E+04 | 6.6474E+03 | 5.6484E+03 | 6.1934E+03 | 1.2880E+04 | 1.2431E+05 | 2.7908E+03 | 1.4374E+04 | 6.1789E+01 |
|     | Rank | 1 | 9 | 4 | 3 | 6 | 8 | 10 | 5 | 7 | 2 |
|     | Best | 1.6033E+03 | 1.8481E+03 | 1.6316E+03 | 1.7656E+03 | 1.7711E+03 | 1.7892E+03 | 1.9537E+03 | 1.7550E+03 | 2.1083E+03 | 2.3679E+03 |
| F15 | Ave | **1.8503E+03** | 2.4530E+03 | 2.2892E+03 | 2.2069E+03 | 2.3661E+03 | 2.3266E+03 | 2.5941E+03 | 2.2298E+03 | 2.5484E+03 | 2.8860E+03 |
|     | Std | 2.2370E+02 | 2.8969E+02 | 2.9743E+02 | 2.6847E+02 | 2.4973E+02 | 2.1509E+02 | 2.8394E+02 | 2.2599E+02 | 2.5187E+02 | 1.8843E+02 |
|     | Rank | 1 | 7 | 4 | 2 | 6 | 5 | 9 | 3 | 8 | 10 |
|     | Best | 1.7255E+03 | 1.7641E+03 | 1.7341E+03 | 1.7273E+03 | 1.7548E+03 | 1.7414E+03 | 1.7951E+03 | 1.7759E+03 | 1.7391E+03 | 1.9001E+03 |
| F16 | Ave | **1.7739E+03** | 2.0206E+03 | 1.9089E+03 | 1.9047E+03 | 1.9313E+03 | 1.9362E+03 | 2.1342E+03 | 1.8600E+03 | 2.0942E+03 | 2.0391E+03 |
|     | Std | 5.2317E+01 | 1.4022E+02 | 1.4820E+02 | 1.5041E+02 | 1.3223E+02 | 1.3340E+02 | 1.8994E+02 | 5.3909E+01 | 1.9135E+02 | 9.9623E+01 |
|     | Rank | 1 | 7 | 4 | 3 | 5 | 6 | 10 | 2 | 9 | 8 |
|     | Best | 1.8226E+03 | 2.7378E+04 | 3.5406E+04 | 2.0775E+04 | 3.3429E+04 | 3.3228E+04 | 1.6829E+05 | 2.2674E+04 | 4.9932E+04 | 2.1758E+03 |
| F17 | Ave | **1.8550E+03** | 3.4221E+05 | 2.4256E+05 | 2.7369E+05 | 3.2434E+05 | 3.1347E+05 | 1.3001E+06 | 4.2717E+04 | 2.6979E+05 | 4.7617E+03 |
|     | Std | 4.3168E+01 | 2.6527E+05 | 2.2444E+05 | 2.1026E+05 | 2.6213E+05 | 2.6118E+05 | 1.3255E+06 | 1.2515E+04 | 2.0557E+05 | 3.5512E+03 |
|     | Rank | 1 | 9 | 4 | 6 | 8 | 7 | 10 | 3 | 5 | 2 |
|     | Best | 1.9077E+03 | 2.7867E+03 | 1.9291E+03 | 1.9138E+03 | 2.9823E+03 | 1.9892E+03 | 4.3215E+04 | 2.9974E+03 | 1.9985E+03 | 1.9427E+03 |
| F18 | Ave | **1.9181E+03** | 1.5941E+04 | 7.9870E+03 | 8.8116E+03 | 1.0754E+04 | 1.2344E+04 | 2.8167E+05 | 9.2782E+03 | 1.2855E+04 | 1.9818E+03 |
|     | Std | 1.9977E+01 | 1.5057E+04 | 9.8760E+03 | 1.1536E+04 | 2.0829E+04 | 1.5351E+04 | 2.4384E+05 | 4.8829E+03 | 1.4397E+04 | 2.6171E+01 |
|     | Rank | 1 | 9 | 3 | 4 | 6 | 7 | 10 | 5 | 8 | 2 |
|     | Best | 2.0223E+03 | 2.0545E+03 | 2.0303E+03 | 2.0493E+03 | 2.0651E+03 | 2.0408E+03 | 2.1096E+03 | 2.0976E+03 | 2.0433E+03 | 2.2377E+03 |
| F19 | Ave | **2.0802E+03** | 2.3253E+03 | 2.2233E+03 | 2.2697E+03 | 2.2621E+03 | 2.2091E+03 | 2.3766E+03 | 2.2626E+03 | 2.3898E+03 | 2.4750E+03 |
|     | Std | 6.5124E+01 | 1.4765E+02 | 1.5414E+02 | 1.5616E+02 | 1.0413E+02 | 1.1005E+02 | 1.5268E+02 | 8.1236E+01 | 1.8727E+02 | 1.0532E+02 |
|     | Rank | 1 | 7 | 3 | 6 | 4 | 2 | 8 | 5 | 9 | 10 |
|     | Best | 2.3091E+03 | 2.3270E+03 | 2.3189E+03 | 2.3197E+03 | 2.3323E+03 | 2.3312E+03 | 2.4097E+03 | 2.3265E+03 | 2.3431E+03 | 2.4189E+03 |
| F20 | Ave | **2.3304E+03** | 2.3818E+03 | 2.3523E+03 | 2.3405E+03 | 2.3642E+03 | 2.3664E+03 | 2.4507E+03 | 2.3519E+03 | 2.3890E+03 | 2.4484E+03 |
|     | Std | 1.0960E+01 | 2.4525E+01 | 1.4608E+01 | 1.1143E+01 | 1.8704E+01 | 1.7106E+01 | 2.3449E+01 | 1.2695E+01 | 2.5559E+01 | 1.1988E+01 |
|     | Rank | 1 | 7 | 4 | 2 | 5 | 6 | 10 | 3 | 8 | 9 |
|     | Best | 2.3000E+03 | 2.3032E+03 | 2.3000E+03 | 2.3000E+03 | 2.3000E+03 | 2.3014E+03 | 2.3744E+03 | 2.3002E+03 | 2.3000E+03 | 2.3005E+03 |
| F21 | Ave | 2.5528E+03 | 3.4571E+03 | 3.0477E+03 | **2.3001E+03** | 2.3008E+03 | 2.6269E+03 | 4.2970E+03 | 2.3013E+03 | 2.9570E+03 | 2.3029E+03 |
|     | Std | 8.2335E+02 | 1.6052E+03 | 1.5449E+03 | 5.8238E-01 | 1.2246E+00 | 9.8924E+02 | 2.3001E+03 | 1.3492E+00 | 1.2274E+03 | 2.3942E+00 |
|     | Rank | 5 | 9 | 8 | 1 | 2 | 6 | 10 | 3 | 7 | 4 |
|     | Best | 2.6533E+03 | 2.6951E+03 | 2.6772E+03 | 2.6590E+03 | 2.6837E+03 | 2.6832E+03 | 2.7586E+03 | 2.6624E+03 | 2.6931E+03 | 2.7811E+03 |
| F22 | Ave | **2.6765E+03** | 2.7309E+03 | 2.7041E+03 | 2.6934E+03 | 2.7154E+03 | 2.7209E+03 | 2.8130E+03 | 2.7000E+03 | 2.7408E+03 | 2.8073E+03 |
|     | Std | 1.1506E+01 | 2.0732E+01 | 1.7964E+01 | 1.5217E+01 | 1.8543E+01 | 1.6121E+01 | 2.8231E+01 | 1.6027E+01 | 2.8173E+01 | 1.3848E+01 |
|     | Rank | 1 | 7 | 4 | 2 | 5 | 6 | 10 | 3 | 8 | 9 |
|     | Best | 2.8225E+03 | 2.8646E+03 | 2.8473E+03 | 2.8445E+03 | 2.8600E+03 | 2.8541E+03 | 2.9435E+03 | 2.8319E+03 | 2.8695E+03 | 2.9368E+03 |
| F23 | Ave | **2.8452E+03** | 2.9024E+03 | 2.8719E+03 | 2.8681E+03 | 2.8948E+03 | 2.9006E+03 | 3.0091E+03 | 2.8610E+03 | 2.9138E+03 | 2.9694E+03 |

|  |  |  |  |  |  |  |  |  |  |  |
|---|---|---|---|---|---|---|---|---|---|---|
|  | Std | 9.6612E+00 | 2.2733E+01 | 1.3978E+01 | 1.6152E+01 | 1.8811E+01 | 2.0280E+01 | 3.3480E+01 | 1.3492E+01 | 2.0015E+01 | 1.5241E+01 |
|  | Rank | 1 | 7 | 4 | 3 | 5 | 6 | 10 | 2 | 8 | 9 |
| F24 | Best | 2.8837E+03 | 2.8871E+03 | 2.8835E+03 | 2.8834E+03 | 2.8838E+03 | 2.8835E+03 | 2.9027E+03 | 2.8836E+03 | 2.8768E+03 | 2.8840E+03 |
|  | Ave | 2.8957E+03 | 2.8926E+03 | 2.8889E+03 | 2.8896E+03 | 2.9146E+03 | 2.8883E+03 | 2.9542E+03 | 2.8886E+03 | **2.8821E+03** | 2.8913E+03 |
|  | Std | 1.5479E+01 | 1.0020E+01 | 1.1289E+01 | 1.0259E+01 | 1.8421E+01 | 4.6262E+00 | 3.4199E+01 | 4.3311E+00 | 1.2261E+01 | 9.6588E+00 |
|  | Rank | 8 | 7 | 4 | 5 | 9 | 2 | 10 | 3 | 1 | 6 |
| F25 | Best | 2.8000E+03 | 2.9023E+03 | 2.8000E+03 | 3.7045E+03 | 2.8001E+03 | 2.9013E+03 | 3.1875E+03 | 2.8935E+03 | 2.8002E+03 | 4.6483E+03 |
|  | Ave | 3.7881E+03 | 4.3876E+03 | 4.0576E+03 | 4.0494E+03 | **3.6283E+03** | 4.3297E+03 | 5.3335E+03 | 3.7757E+03 | 4.3131E+03 | 5.0386E+03 |
|  | Std | 3.6921E+02 | 4.2397E+02 | 3.3750E+02 | 1.4863E+02 | 7.6307E+02 | 4.0486E+02 | 6.6647E+02 | 4.1998E+02 | 8.1907E+02 | 1.9564E+02 |
|  | Rank | 3 | 8 | 5 | 4 | 1 | 7 | 10 | 2 | 6 | 9 |
| F26 | Best | 3.1866E+03 | 3.1981E+03 | 3.1919E+03 | 3.1938E+03 | 3.2006E+03 | 3.1880E+03 | 3.2130E+03 | 3.1957E+03 | 3.1428E+03 | 3.2065E+03 |
|  | Ave | 3.2138E+03 | 3.2217E+03 | 3.2143E+03 | 3.2128E+03 | 3.2172E+03 | 3.2082E+03 | 3.2507E+03 | 3.2140E+03 | **3.1677E+03** | 3.2307E+03 |
|  | Std | 1.4102E+01 | 1.1186E+01 | 8.2745E+00 | 1.1102E+01 | 7.8804E+00 | 6.3238E+00 | 1.4744E+01 | 7.6663E+00 | 1.2381E+01 | 1.1687E+01 |
|  | Rank | 4 | 8 | 6 | 3 | 7 | 2 | 10 | 5 | 1 | 9 |
| F27 | Best | 3.1000E+03 | 3.2111E+03 | 3.1902E+03 | 3.1000E+03 | 3.2234E+03 | 3.2095E+03 | 3.2643E+03 | 3.1977E+03 | 3.1041E+03 | 3.2011E+03 |
|  | Ave | **3.1952E+03** | 3.2614E+03 | 3.2139E+03 | 3.2097E+03 | 3.2747E+03 | 3.2364E+03 | 3.3346E+03 | 3.2187E+03 | 3.2286E+03 | 3.2313E+03 |
|  | Std | 4.8500E+01 | 3.7410E+01 | 1.7385E+01 | 3.7554E+01 | 2.6616E+01 | 2.1804E+01 | 3.4951E+01 | 1.6243E+01 | 3.8159E+01 | 1.9836E+01 |
|  | Rank | 1 | 8 | 3 | 2 | 9 | 7 | 10 | 4 | 5 | 6 |
| F28 | Best | 3.2368E+03 | 3.4447E+03 | 3.2932E+03 | 3.3327E+03 | 3.4047E+03 | 3.2946E+03 | 3.4504E+03 | 3.4333E+03 | 3.3163E+03 | 3.6194E+03 |
|  | Ave | **3.3937E+03** | 3.7487E+03 | 3.5262E+03 | 3.5086E+03 | 3.6566E+03 | 3.5652E+03 | 3.8276E+03 | 3.6092E+03 | 3.6964E+03 | 3.8967E+03 |
|  | Std | 9.2297E+01 | 1.6506E+02 | 1.5486E+02 | 1.4758E+02 | 1.3941E+02 | 1.4872E+02 | 1.7917E+02 | 9.9355E+01 | 1.8100E+02 | 1.0591E+02 |
|  | Rank | 1 | 8 | 3 | 2 | 6 | 4 | 9 | 5 | 7 | 10 |
| F29 | Best | 4.9435E+03 | 3.8593E+04 | 5.2114E+03 | 5.1065E+03 | 5.2884E+04 | 8.0737E+03 | 3.9231E+04 | 2.4229E+04 | 3.5935E+03 | 1.0421E+04 |
|  | Ave | **5.1677E+03** | 3.1346E+05 | 9.7758E+03 | 8.4106E+03 | 3.3199E+05 | 1.7391E+04 | 4.4560E+05 | 1.0060E+05 | 2.0534E+04 | 1.7610E+04 |
|  | Std | 2.5735E+02 | 2.6259E+05 | 6.2910E+03 | 3.1780E+03 | 2.7168E+05 | 7.9617E+03 | 3.4266E+05 | 5.5698E+04 | 3.0555E+04 | 5.2917E+03 |
|  | Rank | 1 | 8 | 3 | 2 | 9 | 4 | 10 | 7 | 6 | 5 |

**Table A3.** Quantitative results of MRIME-CD and other competitors based on CEC2017 (D=50)

| No. | Metric | MRIME-CD | RIME | EO | SAO | ACGRIME | IRIME | TERIME | EOSMA | RDGMVO | MTVSCA |
|---|---|---|---|---|---|---|---|---|---|---|---|
| F1 | Best | 1.0143E+02 | 1.1713E+06 | 1.0626E+02 | 1.0000E+02 | 7.4603E+03 | 6.2797E+05 | 1.1079E+09 | 1.8409E+05 | 1.7869E+02 | 3.2812E+05 |
|  | Ave | 3.2294E+03 | 3.2988E+06 | **2.2836E+03** | 2.5571E+03 | 6.0743E+04 | 2.0874E+06 | 2.5689E+09 | 8.5983E+05 | 6.5542E+03 | 1.1396E+06 |
|  | Std | 3.3943E+03 | 1.3513E+06 | 2.4540E+03 | 2.9134E+03 | 9.0164E+03 | 9.5971E+05 | 8.8583E+08 | 4.3942E+05 | 1.2001E+04 | 5.3791E+05 |
|  | Rank | 3 | 9 | 1 | 2 | 5 | 8 | 10 | 6 | 4 | 7 |
| F2 | Best | 3.0000E+02 | 1.6966E+04 | 1.6630E+04 | 9.0193E+04 | 5.6470E+04 | 4.1801E+04 | 7.3699E+04 | 4.4429E+04 | 3.3233E+02 | 2.4931E+04 |
|  | Ave | **3.0001E+02** | 3.0232E+04 | 2.8187E+04 | 1.2786E+05 | 7.7959E+04 | 8.6256E+04 | 1.2234E+05 | 7.0116E+04 | 5.3861E+02 | 4.3101E+04 |
|  | Std | 8.8253E-03 | 6.7421E+03 | 7.0261E+03 | 1.8982E+04 | 1.3038E+04 | 2.0184E+04 | 2.3697E+04 | 1.1318E+04 | 1.6091E+02 | 8.8974E+03 |
|  | Rank | 1 | 4 | 3 | 10 | 7 | 8 | 9 | 6 | 2 | 5 |
| F3 | Best | 4.0966E+02 | 5.2741E+02 | 4.2856E+02 | 4.2854E+02 | 5.3238E+02 | 4.4867E+02 | 6.7942E+02 | 4.7324E+02 | 4.2705E+02 | 5.1953E+02 |
|  | Ave | 5.4603E+02 | 6.0104E+02 | 5.1521E+02 | 5.3083E+02 | 6.0414E+02 | 5.6283E+02 | 8.0283E+02 | 5.5787E+02 | **4.7364E+02** | 5.9361E+02 |
|  | Std | 5.9848E+01 | 3.9516E+01 | 5.1285E+01 | 4.6325E+01 | 3.9080E+01 | 4.3158E+01 | 6.4517E+01 | 4.6276E+01 | 3.7363E+01 | 3.7949E+01 |
|  | Rank | 4 | 8 | 2 | 3 | 9 | 6 | 10 | 5 | 1 | 7 |
| F4 | Best | 5.3681E+02 | 5.8452E+02 | 5.8581E+02 | 5.4378E+02 | 5.9553E+02 | 6.0134E+02 | 7.5839E+02 | 5.7564E+02 | 6.3730E+02 | 7.9023E+02 |
|  | Ave | **5.7122E+02** | 6.4338E+02 | 6.3618E+02 | 5.7771E+02 | 6.5918E+02 | 6.3471E+02 | 8.2961E+02 | 6.1874E+02 | 7.1534E+02 | 8.3033E+02 |
|  | Std | 1.7163E+01 | 2.9027E+01 | 2.6946E+01 | 1.9306E+01 | 3.4314E+01 | 2.1743E+01 | 3.4761E+01 | 2.9423E+01 | 3.7307E+01 | 1.8675E+01 |

|  | Rank | 1 | 6 | 5 | 2 | 7 | 4 | 9 | 3 | 8 | 10 |
|---|---|---|---|---|---|---|---|---|---|---|---|
| F5 | Best | 6.0002E+02 | 6.0294E+02 | 6.0001E+02 | 6.0000E+02 | 6.0112E+02 | 6.0034E+02 | 6.1230E+02 | 6.0072E+02 | 6.0072E+02 | 6.0050E+02 |
|  | Ave | 6.0044E+02 | 6.0619E+02 | 6.0008E+02 | **6.0001E+02** | 6.0356E+02 | 6.0066E+02 | 6.1795E+02 | 6.0191E+02 | 6.0379E+02 | 6.0094E+02 |
|  | Std | 3.7175E-01 | 2.2615E+00 | 8.8321E-02 | 4.4036E-02 | 1.5203E+00 | 2.0450E-01 | 3.1922E+00 | 7.0822E-01 | 2.3181E+00 | 3.4137E-01 |
|  | Rank | 3 | 9 | 2 | 1 | 7 | 4 | 10 | 6 | 8 | 5 |
| F6 | Best | 7.8773E+02 | 8.7074E+02 | 8.1830E+02 | 7.9386E+02 | 8.4033E+02 | 8.4574E+02 | 1.1741E+03 | 8.4827E+02 | 8.5329E+02 | 1.0668E+03 |
|  | Ave | **8.1105E+02** | 9.3086E+02 | 8.8102E+02 | 8.3450E+02 | 8.9978E+02 | 9.0927E+02 | 1.2898E+03 | 8.9904E+02 | 9.6922E+02 | 1.1135E+03 |
|  | Std | 1.4325E+01 | 3.8059E+01 | 3.6524E+01 | 5.0526E+01 | 3.0286E+01 | 3.8055E+01 | 6.4186E+01 | 2.9098E+01 | 5.1582E+01 | 1.9788E+01 |
|  | Rank | 1 | 7 | 3 | 2 | 5 | 6 | 10 | 4 | 8 | 9 |
| F7 | Best | 8.3084E+02 | 8.8616E+02 | 8.9751E+02 | 8.3880E+02 | 9.1117E+02 | 8.8539E+02 | 1.0614E+03 | 8.6708E+02 | 8.9552E+02 | 1.0818E+03 |
|  | Ave | **8.6788E+02** | 9.4619E+02 | 9.3588E+02 | 8.8332E+02 | 9.6549E+02 | 9.3443E+02 | 1.1374E+03 | 9.1897E+02 | 9.8349E+02 | 1.1296E+03 |
|  | Std | 1.4126E+01 | 3.9788E+01 | 2.7899E+01 | 5.4956E+01 | 3.4043E+01 | 2.7981E+01 | 3.5694E+01 | 2.2818E+01 | 3.6207E+01 | 2.0573E+01 |
|  | Rank | 1 | 6 | 5 | 2 | 7 | 4 | 10 | 3 | 8 | 9 |
| F8 | Best | 9.0971E+02 | 1.0360E+03 | 9.0518E+02 | 9.0000E+02 | 1.1563E+03 | 9.4731E+02 | 3.3798E+03 | 9.2735E+02 | 2.3299E+03 | 9.1577E+02 |
|  | Ave | 9.4130E+02 | 2.2642E+03 | 9.7911E+02 | **9.0976E+02** | 2.4992E+03 | 1.1132E+03 | 8.4758E+03 | 1.0346E+03 | 7.3350E+03 | 9.9874E+02 |
|  | Std | 3.5956E+01 | 1.1789E+03 | 1.3538E+02 | 1.5999E+01 | 1.3720E+03 | 1.2472E+02 | 3.1925E+03 | 9.1579E+01 | 2.5509E+03 | 6.2582E+01 |
|  | Rank | 2 | 7 | 3 | 1 | 8 | 6 | 10 | 5 | 9 | 4 |
| F9 | Best | 4.6266E+03 | 5.3943E+03 | 4.5843E+03 | 3.5560E+03 | 5.5893E+03 | 5.1371E+03 | 7.6380E+03 | 8.5334E+03 | 4.5176E+03 | 1.1809E+04 |
|  | Ave | 6.2769E+03 | 7.0817E+03 | 7.2816E+03 | **5.1408E+03** | 7.0757E+03 | 6.9543E+03 | 9.4782E+03 | 1.0596E+04 | 6.3042E+03 | 1.3375E+04 |
|  | Std | 7.9564E+02 | 8.0655E+02 | 1.1618E+03 | 8.3877E+02 | 7.6753E+02 | 8.4356E+02 | 8.4737E+02 | 8.6803E+02 | 8.9858E+02 | 5.0054E+02 |
|  | Rank | 2 | 6 | 7 | 1 | 5 | 4 | 8 | 9 | 3 | 10 |
| F10 | Best | 1.1797E+03 | 1.3081E+03 | 1.1574E+03 | 1.1368E+03 | 1.3195E+03 | 1.2131E+03 | 1.3856E+03 | 1.2799E+03 | 1.1750E+03 | 1.2497E+03 |
|  | Ave | 1.2688E+03 | 1.4964E+03 | 1.2283E+03 | **1.1634E+03** | 1.5075E+03 | 1.2860E+03 | 1.6046E+03 | 1.3697E+03 | 1.2458E+03 | 1.3530E+03 |
|  | Std | 4.4445E+01 | 8.6959E+01 | 4.2849E+01 | 2.3259E+01 | 9.5110E+01 | 4.1465E+01 | 1.4224E+02 | 3.9915E+01 | 4.3978E+01 | 6.1207E+01 |
|  | Rank | 4 | 8 | 2 | 1 | 9 | 5 | 10 | 7 | 3 | 6 |
| F11 | Best | 1.6088E+04 | 7.9885E+06 | 2.8091E+05 | 2.7579E+05 | 1.2991E+07 | 1.9668E+06 | 4.4410E+07 | 1.0700E+06 | 3.2512E+06 | 2.3865E+05 |
|  | Ave | **2.0308E+05** | 6.4237E+07 | 2.1052E+06 | 1.4976E+06 | 4.1768E+07 | 1.1406E+07 | 1.6119E+08 | 4.4703E+06 | 1.3493E+07 | 2.5389E+06 |
|  | Std | 1.9804E+05 | 3.7113E+07 | 1.5652E+06 | 1.0075E+06 | 2.5266E+07 | 5.5933E+06 | 5.4781E+07 | 2.3856E+06 | 7.2419E+06 | 1.2579E+06 |
|  | Rank | 1 | 9 | 3 | 2 | 8 | 6 | 10 | 5 | 7 | 4 |
| F12 | Best | 1.5850E+03 | 6.9448E+04 | 1.9708E+03 | 1.3640E+03 | 2.8553E+04 | 5.1789E+03 | 2.3959E+06 | 1.0121E+04 | 4.3408E+03 | 5.9341E+03 |
|  | Ave | **4.4876E+03** | 1.8994E+05 | 8.6542E+03 | 5.2059E+03 | 1.4824E+05 | 2.4286E+04 | 1.3449E+07 | 4.6902E+04 | 1.8844E+04 | 1.3983E+04 |
|  | Std | 3.3893E+03 | 8.9622E+04 | 7.0484E+03 | 6.9814E+03 | 1.1812E+05 | 1.2240E+04 | 8.2867E+06 | 2.8956E+04 | 1.5128E+04 | 6.2976E+03 |
|  | Rank | 1 | 9 | 3 | 2 | 8 | 6 | 10 | 7 | 5 | 4 |
| F13 | Best | 1.4334E+03 | 1.3943E+04 | 4.6787E+03 | 6.9629E+03 | 3.3380E+04 | 3.9589E+03 | 9.0955E+04 | 2.7834E+03 | 1.1651E+04 | 1.5894E+03 |
|  | Ave | **1.4568E+03** | 1.6308E+05 | 9.3115E+04 | 6.6926E+04 | 2.9450E+05 | 1.1806E+05 | 5.5057E+05 | 6.4949E+03 | 1.0672E+05 | 1.7043E+03 |
|  | Std | 2.1662E+01 | 1.0402E+05 | 7.2116E+04 | 5.7262E+04 | 1.8118E+05 | 8.6403E+04 | 3.1374E+05 | 2.7487E+03 | 8.7406E+04 | 6.4946E+01 |
|  | Rank | 1 | 8 | 5 | 4 | 9 | 7 | 10 | 3 | 6 | 2 |
| F14 | Best | 1.5846E+03 | 1.3781E+04 | 1.7281E+03 | 1.6229E+03 | 2.3158E+04 | 3.0882E+03 | 1.9553E+05 | 3.1466E+03 | 1.7563E+03 | 2.2513E+03 |
|  | Ave | **1.8511E+03** | 5.1509E+04 | 1.1938E+04 | 1.3445E+04 | 4.9045E+04 | 1.6444E+04 | 1.5821E+06 | 2.4594E+04 | 1.7020E+04 | 3.1029E+03 |
|  | Std | 1.7730E+02 | 2.5947E+04 | 7.8315E+03 | 5.5026E+03 | 2.1888E+04 | 6.3564E+03 | 1.3399E+06 | 1.1797E+04 | 1.9523E+04 | 1.1781E+03 |
|  | Rank | 1 | 9 | 3 | 4 | 8 | 5 | 10 | 7 | 6 | 2 |
| F15 | Best | 1.7330E+03 | 2.5444E+03 | 2.0770E+03 | 2.0347E+03 | 2.3079E+03 | 2.3414E+03 | 2.4792E+03 | 1.7927E+03 | 2.4088E+03 | 3.1485E+03 |
|  | Ave | **2.0044E+03** | 3.3614E+03 | 2.8290E+03 | 2.8004E+03 | 3.1066E+03 | 3.1097E+03 | 3.5465E+03 | 2.7647E+03 | 3.2175E+03 | 4.0360E+03 |
|     | Std  | 2.4101E+02 | 3.4437E+02 | 3.9044E+02 | 3.4066E+02 | 3.9599E+02 | 3.7135E+02 | 4.6734E+02 | 3.7226E+02 | 3.6997E+02 | 3.0672E+02 |
|     | Rank | 1 | 8 | 4 | 3 | 5 | 6 | 9 | 2 | 7 | 10 |
|     | Best | 1.7613E+03 | 2.3517E+03 | 1.9004E+03 | 1.9293E+03 | 2.1108E+03 | 2.0265E+03 | 2.2588E+03 | 2.0498E+03 | 2.1835E+03 | 2.7559E+03 |
| F16 | Ave  | **2.2728E+03** | 3.0210E+03 | 2.7998E+03 | 2.5440E+03 | 2.7804E+03 | 2.7741E+03 | 3.0594E+03 | 2.5776E+03 | 2.8473E+03 | 3.2929E+03 |
|     | Std  | 2.2600E+02 | 3.0688E+02 | 2.8257E+02 | 2.9828E+02 | 2.7191E+02 | 2.7777E+02 | 3.3499E+02 | 2.4157E+02 | 3.2821E+02 | 2.1694E+02 |
|     | Rank | 1 | 8 | 6 | 2 | 5 | 4 | 9 | 3 | 7 | 10 |
|     | Best | 1.8673E+03 | 2.6756E+05 | 1.4516E+05 | 4.8870E+04 | 3.0769E+05 | 1.3917E+05 | 4.4317E+05 | 4.2629E+04 | 1.0616E+05 | 9.5040E+03 |
| F17 | Ave  | **2.3332E+03** | 1.8241E+06 | 8.8357E+05 | 8.0623E+05 | 1.8533E+06 | 9.4044E+05 | 4.5609E+06 | 1.7833E+05 | 1.2582E+06 | 2.8550E+04 |
|     | Std  | 4.3348E+02 | 1.2570E+06 | 5.0655E+05 | 5.5418E+05 | 1.3932E+06 | 6.9803E+05 | 3.0635E+06 | 7.6872E+04 | 8.7099E+05 | 1.7432E+04 |
|     | Rank | 1 | 8 | 5 | 4 | 9 | 6 | 10 | 3 | 7 | 2 |
|     | Best | 1.9401E+03 | 1.0150E+04 | 2.0522E+03 | 1.9361E+03 | 1.0266E+04 | 2.3994E+03 | 1.6017E+05 | 3.6003E+03 | 2.2373E+03 | 2.0675E+03 |
| F18 | Ave  | **2.8110E+03** | 8.5626E+04 | 1.8287E+04 | 2.0849E+04 | 1.1289E+05 | 2.2694E+04 | 6.9329E+05 | 1.9707E+04 | 1.9543E+04 | 3.3841E+03 |
|     | Std  | 2.2471E+03 | 6.7627E+04 | 1.1557E+04 | 1.0803E+04 | 1.0749E+05 | 1.4448E+04 | 4.4730E+05 | 9.7776E+03 | 1.7387E+04 | 2.2664E+03 |
|     | Rank | 1 | 8 | 3 | 6 | 9 | 7 | 10 | 5 | 4 | 2 |
|     | Best | 2.0654E+03 | 2.3220E+03 | 2.1251E+03 | 2.0410E+03 | 2.2979E+03 | 2.2380E+03 | 2.3104E+03 | 2.2855E+03 | 2.4680E+03 | 2.9966E+03 |
| F19 | Ave  | **2.3977E+03** | 2.9469E+03 | 2.7415E+03 | 2.5775E+03 | 2.8186E+03 | 2.7776E+03 | 3.0467E+03 | 2.7769E+03 | 2.9506E+03 | 3.5019E+03 |
|     | Std  | 2.0306E+02 | 2.5405E+02 | 3.1751E+02 | 2.9357E+02 | 2.5844E+02 | 2.5724E+02 | 2.6614E+02 | 2.5569E+02 | 2.4797E+02 | 1.6965E+02 |
|     | Rank | 1 | 7 | 3 | 2 | 6 | 5 | 9 | 4 | 8 | 10 |
|     | Best | 2.3341E+03 | 2.3886E+03 | 2.3555E+03 | 2.3456E+03 | 2.3677E+03 | 2.3896E+03 | 2.5514E+03 | 2.3671E+03 | 2.4234E+03 | 2.5838E+03 |
| F20 | Ave  | **2.3582E+03** | 2.4498E+03 | 2.4142E+03 | 2.3836E+03 | 2.4161E+03 | 2.4359E+03 | 2.6334E+03 | 2.4094E+03 | 2.4893E+03 | 2.6234E+03 |
|     | Std  | 1.3791E+01 | 3.6941E+01 | 2.4586E+01 | 1.8292E+01 | 2.3204E+01 | 2.7665E+01 | 3.6513E+01 | 2.6607E+01 | 3.5637E+01 | 1.7845E+01 |
|     | Rank | 1 | 7 | 4 | 2 | 5 | 6 | 10 | 3 | 8 | 9 |
|     | Best | 2.3000E+03 | 2.3164E+03 | 2.3000E+03 | 2.3000E+03 | 2.3007E+03 | 6.4790E+03 | 9.0899E+03 | 2.3058E+03 | 2.3000E+03 | 2.3491E+03 |
| F21 | Ave  | 7.3802E+03 | 8.7034E+03 | 9.0461E+03 | **5.4241E+03** | 7.1097E+03 | 8.5162E+03 | 1.1189E+04 | 1.0671E+04 | 7.8473E+03 | 1.4849E+04 |
|     | Std  | 1.5864E+03 | 1.2063E+03 | 1.5247E+03 | 2.8073E+03 | 2.8332E+03 | 7.8250E+02 | 6.3344E+02 | 3.3715E+03 | 1.1861E+03 | 1.8335E+03 |
|     | Rank | 3 | 6 | 7 | 1 | 2 | 5 | 9 | 8 | 4 | 10 |
|     | Best | 2.7469E+03 | 2.8130E+03 | 2.7857E+03 | 2.7586E+03 | 2.7927E+03 | 2.8018E+03 | 3.0329E+03 | 2.7873E+03 | 2.8557E+03 | 2.9920E+03 |
| F22 | Ave  | **2.7827E+03** | 2.8928E+03 | 2.8345E+03 | 2.8028E+03 | 2.8590E+03 | 2.8710E+03 | 3.1104E+03 | 2.8357E+03 | 2.9259E+03 | 3.0634E+03 |
|     | Std  | 2.1227E+01 | 3.4263E+01 | 3.2233E+01 | 1.9557E+01 | 3.2865E+01 | 2.9671E+01 | 3.7385E+01 | 2.6485E+01 | 4.2952E+01 | 2.3908E+01 |
|     | Rank | 1 | 7 | 3 | 2 | 5 | 6 | 10 | 4 | 8 | 9 |
|     | Best | 2.9261E+03 | 3.0017E+03 | 2.9505E+03 | 2.9530E+03 | 2.9640E+03 | 2.9740E+03 | 3.2295E+03 | 2.9392E+03 | 3.0208E+03 | 3.1466E+03 |
| F23 | Ave  | **2.9497E+03** | 3.0539E+03 | 3.0042E+03 | 2.9882E+03 | 3.0250E+03 | 3.0431E+03 | 3.3594E+03 | 2.9850E+03 | 3.0768E+03 | 3.2152E+03 |
|     | Std  | 1.2543E+01 | 3.5664E+01 | 3.0737E+01 | 2.2904E+01 | 3.2317E+01 | 2.8009E+01 | 7.5084E+01 | 2.0482E+01 | 4.4515E+01 | 2.4970E+01 |
|     | Rank | 1 | 7 | 4 | 3 | 5 | 6 | 10 | 2 | 8 | 9 |
|     | Best | 2.9648E+03 | 3.0232E+03 | 2.9815E+03 | 2.9605E+03 | 3.0288E+03 | 2.9898E+03 | 3.1576E+03 | 3.0102E+03 | 2.9340E+03 | 3.0358E+03 |
| F24 | Ave  | 3.0604E+03 | 3.0739E+03 | 3.0680E+03 | 3.0321E+03 | 3.1048E+03 | 3.0660E+03 | 3.3055E+03 | 3.0862E+03 | **2.9711E+03** | 3.0893E+03 |
|     | Std  | 3.8853E+01 | 3.3236E+01 | 2.9285E+01 | 3.7253E+01 | 3.9796E+01 | 3.1386E+01 | 9.5975E+01 | 2.1304E+01 | 3.0097E+01 | 2.9124E+01 |
|     | Rank | 3 | 6 | 5 | 2 | 9 | 4 | 10 | 7 | 1 | 8 |
|     | Best | 2.9000E+03 | 4.8891E+03 | 4.1510E+03 | 4.0353E+03 | 2.9133E+03 | 4.8337E+03 | 6.7482E+03 | 4.3156E+03 | 2.9000E+03 | 5.6435E+03 |
| F25 | Ave  | 4.4603E+03 | 5.4316E+03 | 4.9158E+03 | 4.4557E+03 | **3.3459E+03** | 5.2705E+03 | 7.5011E+03 | 4.7812E+03 | 4.3990E+03 | 6.8581E+03 |
|     | Std  | 3.9836E+02 | 3.5771E+02 | 3.3839E+02 | 2.0073E+02 | 6.6105E+02 | 2.6805E+02 | 4.4611E+02 | 2.5513E+02 | 1.6050E+03 | 3.3180E+02 |
|     | Rank | 4 | 8 | 6 | 3 | 1 | 7 | 10 | 5 | 2 | 9 |
| F26 | Best | 3.2321E+03 | 3.3042E+03 | 3.2384E+03 | 3.2376E+03 | 3.2915E+03 | 3.2594E+03 | 3.3243E+03 | 3.2592E+03 | 3.1656E+03 | 3.2916E+03 |

|  | Metric | MRIME-CD | RIME | EO | SAO | ACGRIME | IRIME | TERIME | EOSMA | RDGMVO | MTVSCA |
|---|---|---|---|---|---|---|---|---|---|---|---|
|  | Ave | 3.3707E+03 | 3.3984E+03 | 3.3278E+03 | 3.3074E+03 | 3.4463E+03 | 3.3475E+03 | 3.5064E+03 | 3.3615E+03 | **3.1980E+03** | 3.3893E+03 |
|  | Std | 1.0610E+02 | 5.8013E+01 | 5.9546E+01 | 5.3489E+01 | 7.6802E+01 | 3.9789E+01 | 1.0031E+02 | 5.0268E+01 | 1.1463E+01 | 5.5710E+01 |
|  | Rank | 6 | 8 | 3 | 2 | 9 | 4 | 10 | 5 | 1 | 7 |
|  | Best | 3.2599E+03 | 3.2956E+03 | 3.2685E+03 | 3.2589E+03 | 3.3114E+03 | 3.2735E+03 | 3.4492E+03 | 3.2833E+03 | 3.2437E+03 | 3.2813E+03 |
| F27 | Ave | 3.3152E+03 | 3.3439E+03 | 3.3125E+03 | 3.2880E+03 | 3.3694E+03 | 3.3359E+03 | 3.6220E+03 | 3.3398E+03 | **3.2834E+03** | 3.3703E+03 |
|  | Std | 3.4070E+01 | 2.4502E+01 | 2.0419E+01 | 2.4463E+01 | 3.7782E+01 | 3.1436E+01 | 1.1468E+02 | 2.9994E+01 | 1.6714E+01 | 4.1303E+01 |
|  | Rank | 4 | 7 | 3 | 2 | 8 | 5 | 10 | 6 | 1 | 9 |
|  | Best | 3.3196E+03 | 3.5524E+03 | 3.3773E+03 | 3.2314E+03 | 3.5540E+03 | 3.5602E+03 | 3.5499E+03 | 3.4900E+03 | 3.4946E+03 | 4.1399E+03 |
| F28 | Ave | **3.4506E+03** | 4.3664E+03 | 3.7501E+03 | 3.6955E+03 | 4.1032E+03 | 3.9068E+03 | 4.2169E+03 | 3.9821E+03 | 4.1435E+03 | 4.6788E+03 |
|  | Std | 1.0716E+02 | 3.4910E+02 | 2.3664E+02 | 3.2406E+02 | 2.3098E+02 | 1.9641E+02 | 2.8490E+02 | 2.3876E+02 | 2.9282E+02 | 2.3615E+02 |
|  | Rank | 1 | 9 | 3 | 2 | 6 | 4 | 8 | 5 | 7 | 10 |
|  | Best | 5.8243E+05 | 8.3169E+06 | 6.4033E+05 | 6.5809E+05 | 7.2090E+06 | 7.6318E+05 | 3.1320E+06 | 4.9055E+06 | 4.3120E+03 | 1.6768E+06 |
| F29 | Ave | 7.0782E+05 | 2.0096E+07 | 1.0702E+06 | 9.2521E+05 | 2.0843E+07 | 1.1848E+06 | 6.6224E+06 | 1.2093E+07 | **5.6915E+05** | 3.4051E+06 |
|  | Std | 9.6834E+04 | 7.2820E+06 | 2.5840E+05 | 1.6511E+05 | 9.1634E+06 | 2.7265E+05 | 2.3966E+06 | 3.8231E+06 | 3.6101E+05 | 7.6473E+05 |
|  | Rank | 2 | 9 | 4 | 3 | 10 | 5 | 7 | 8 | 1 | 6 |

Table A4. Quantitative results of MRIME-CD and other competitors based on CEC2017 (D=100)

| No. | Metric | MRIME-CD | RIME | EO | SAO | ACGRIME | IRIME | TERIME | EOSMA | RDGMVO | MTVSCA |
|---|---|---|---|---|---|---|---|---|---|---|---|
|  | Best | 1.2712E+02 | 1.5196E+07 | 1.4091E+03 | 1.5048E+02 | 2.7867E+06 | 1.3154E+07 | 1.0787E+10 | 2.5904E+07 | 6.7173E+02 | 2.1297E+07 |
| F1 | Ave | 6.0299E+03 | 2.8874E+07 | 8.9705E+03 | **5.0679E+03** | 1.5772E+07 | 4.0195E+07 | 2.3980E+10 | 5.6972E+07 | 1.0066E+04 | 5.2496E+07 |
|  | Std | 6.1014E+03 | 6.3999E+06 | 7.9242E+03 | 5.6387E+03 | 1.0944E+07 | 1.2506E+07 | 5.2659E+09 | 2.4894E+07 | 1.2591E+04 | 1.8912E+07 |
|  | Rank | 2 | 6 | 3 | 1 | 5 | 7 | 10 | 9 | 4 | 8 |
|  | Best | 3.2083E+02 | 1.5108E+05 | 1.2179E+05 | 3.0280E+05 | 1.7637E+05 | 1.9276E+05 | 2.9095E+05 | 2.1092E+05 | 9.8406E+03 | 1.1688E+05 |
| F2 | Ave | **3.6217E+02** | 2.0772E+05 | 1.5245E+05 | 3.5672E+05 | 2.2350E+05 | 2.6532E+05 | 3.8665E+05 | 2.6160E+05 | 2.0166E+04 | 1.4928E+05 |
|  | Std | 2.8564E+01 | 3.1733E+04 | 1.5475E+04 | 3.1534E+04 | 1.7651E+04 | 3.0286E+04 | 5.2402E+04 | 2.3746E+04 | 5.4684E+03 | 1.7039E+04 |
|  | Rank | 1 | 5 | 4 | 9 | 6 | 8 | 10 | 7 | 2 | 3 |
|  | Best | 6.2205E+02 | 6.8559E+02 | 5.9857E+02 | 6.0451E+02 | 7.2017E+02 | 6.6683E+02 | 1.3733E+03 | 7.0703E+02 | 4.9446E+02 | 7.3701E+02 |
| F3 | Ave | 7.0010E+02 | 7.8608E+02 | 7.1012E+02 | 6.8042E+02 | 8.6351E+02 | 7.8042E+02 | 1.7575E+03 | 8.2888E+02 | **5.7997E+02** | 8.4530E+02 |
|  | Std | 4.9687E+01 | 5.3599E+01 | 3.7739E+01 | 4.1893E+01 | 6.8324E+01 | 5.4122E+01 | 2.1932E+02 | 6.0585E+01 | 5.2635E+01 | 5.7312E+01 |
|  | Rank | 3 | 6 | 4 | 2 | 9 | 5 | 10 | 7 | 1 | 8 |
|  | Best | 6.1442E+02 | 7.7057E+02 | 7.3315E+02 | 6.3631E+02 | 7.8021E+02 | 7.5829E+02 | 1.2600E+03 | 7.7261E+02 | 8.7908E+02 | 1.2912E+03 |
| F4 | Ave | **6.7344E+02** | 8.6459E+02 | 9.0313E+02 | 6.8624E+02 | 9.5313E+02 | 8.6365E+02 | 1.4452E+03 | 8.6409E+02 | 1.0190E+03 | 1.3553E+03 |
|  | Std | 2.7125E+01 | 5.8698E+01 | 6.5034E+01 | 2.7148E+01 | 9.2095E+01 | 6.1636E+01 | 7.7181E+01 | 4.7835E+01 | 7.2582E+01 | 3.5115E+01 |
|  | Rank | 1 | 5 | 6 | 2 | 7 | 3 | 10 | 4 | 8 | 9 |
|  | Best | 6.0072E+02 | 6.0929E+02 | 6.0029E+02 | 6.0002E+02 | 6.0678E+02 | 6.0187E+02 | 6.2334E+02 | 6.0459E+02 | 6.0927E+02 | 6.0243E+02 |
| F5 | Ave | 6.0276E+02 | 6.1566E+02 | 6.0166E+02 | **6.0009E+02** | 6.1346E+02 | 6.0327E+02 | 6.3273E+02 | 6.0853E+02 | 6.1654E+02 | 6.0442E+02 |
|  | Std | 9.4770E-01 | 3.5619E+00 | 7.7425E-01 | 9.5991E-02 | 3.7962E+00 | 6.6620E-01 | 4.4944E+00 | 2.1603E+00 | 5.2689E+00 | 1.0258E+00 |
|  | Rank | 3 | 8 | 2 | 1 | 7 | 4 | 10 | 6 | 9 | 5 |
|  | Best | 9.1253E+02 | 1.1620E+03 | 1.0751E+03 | 9.3074E+02 | 1.0986E+03 | 1.1804E+03 | 2.2480E+03 | 1.1143E+03 | 1.1920E+03 | 1.6647E+03 |
| F6 | Ave | **9.7235E+02** | 1.3370E+03 | 1.2326E+03 | 1.0557E+03 | 1.2335E+03 | 1.2654E+03 | 2.5556E+03 | 1.3233E+03 | 1.4693E+03 | 1.7513E+03 |
|  | Std | 3.3729E+01 | 9.0708E+01 | 1.0536E+02 | 1.6514E+02 | 7.2523E+01 | 5.4582E+01 | 1.6001E+02 | 8.1559E+01 | 1.3059E+02 | 3.7855E+01 |
|  | Rank | 1 | 7 | 3 | 2 | 4 | 5 | 10 | 6 | 8 | 9 |
|  | Best | 9.1641E+02 | 1.0576E+03 | 1.0555E+03 | 9.3830E+02 | 1.0890E+03 | 1.0595E+03 | 1.6066E+03 | 1.0521E+03 | 1.1701E+03 | 1.5464E+03 |
| F7 | Ave | **9.7593E+02** | 1.1805E+03 | 1.1564E+03 | 9.8689E+02 | 1.1854E+03 | 1.1614E+03 | 1.7641E+03 | 1.1585E+03 | 1.3164E+03 | 1.6605E+03 |

|  |  |  |  |  |  |  |  |  |  |  |
|---|---|---|---|---|---|---|---|---|---|---|
|  | Std | 2.3967E+01 | 7.5883E+01 | 5.9895E+01 | 3.0059E+01 | 5.2084E+01 | 5.8054E+01 | 6.5191E+01 | 5.4981E+01 | 7.4153E+01 | 3.9193E+01 |
|  | Rank | 1 | 6 | 3 | 2 | 7 | 5 | 10 | 4 | 8 | 9 |
|  | Best | 1.0095E+03 | 3.0717E+03 | 1.4481E+03 | 9.0484E+02 | 8.1378E+03 | 2.1015E+03 | 2.4711E+04 | 2.4929E+03 | 1.0851E+04 | 1.6470E+03 |
| F8 | Ave | 1.1822E+03 | 1.0643E+04 | 8.9521E+03 | **9.4485E+02** | 2.0812E+04 | 4.9837E+03 | 4.1737E+04 | 5.8106E+03 | 1.9624E+04 | 3.4381E+03 |
|  | Std | 1.0375E+02 | 5.3858E+03 | 4.3659E+03 | 3.4290E+01 | 5.4654E+03 | 1.9453E+03 | 1.1154E+04 | 2.0848E+03 | 3.3578E+03 | 1.3541E+03 |
|  | Rank | 2 | 7 | 6 | 1 | 9 | 4 | 10 | 5 | 8 | 3 |
|  | Best | 1.0610E+04 | 1.2396E+04 | 1.1714E+04 | 7.7046E+03 | 1.3017E+04 | 1.1907E+04 | 2.1030E+04 | 1.8413E+04 | 1.0515E+04 | 2.7709E+04 |
| F9 | Ave | **1.3418E+04** | 1.6180E+04 | 1.5695E+04 | 1.4092E+04 | 1.5415E+04 | 1.5544E+04 | 2.2985E+04 | 2.3069E+04 | 1.3481E+04 | 2.9568E+04 |
|  | Std | 1.4338E+03 | 1.3859E+03 | 1.6761E+03 | 6.5153E+03 | 1.2865E+03 | 1.3764E+03 | 1.0222E+03 | 1.9075E+03 | 1.3220E+03 | 7.0580E+02 |
|  | Rank | 1 | 7 | 6 | 3 | 4 | 5 | 8 | 9 | 2 | 10 |
|  | Best | 1.8206E+03 | 2.9042E+03 | 2.6296E+03 | 2.1274E+03 | 1.2486E+04 | 4.5182E+03 | 7.8833E+03 | 6.1573E+03 | 1.7646E+03 | 2.7954E+03 |
| F10 | Ave | **2.2878E+03** | 3.6570E+03 | 3.3652E+03 | 2.9024E+03 | 2.0689E+04 | 6.5941E+03 | 2.2721E+04 | 8.5899E+03 | 2.3267E+03 | 3.5408E+03 |
|  | Std | 1.8353E+02 | 3.7314E+02 | 4.6826E+02 | 4.6556E+02 | 5.2149E+03 | 1.3871E+03 | 7.3003E+03 | 1.4544E+03 | 2.1273E+02 | 4.0413E+02 |
|  | Rank | 1 | 6 | 4 | 3 | 9 | 7 | 10 | 8 | 2 | 5 |
|  | Best | 9.7202E+05 | 7.6500E+07 | 2.7173E+06 | 2.6874E+06 | 1.1783E+08 | 2.2673E+07 | 8.1444E+08 | 7.1800E+06 | 1.2738E+07 | 7.4515E+06 |
| F11 | Ave | **6.3937E+06** | 4.8748E+08 | 9.7050E+06 | 9.1975E+06 | 3.6125E+08 | 9.6198E+07 | 1.7390E+09 | 3.4435E+07 | 4.8755E+07 | 2.6617E+07 |
|  | Std | 4.2479E+06 | 1.8259E+08 | 4.8201E+06 | 4.0384E+06 | 1.5327E+08 | 4.5257E+07 | 5.2431E+08 | 1.4443E+07 | 1.8297E+07 | 1.0503E+07 |
|  | Rank | 1 | 9 | 3 | 2 | 8 | 7 | 10 | 5 | 6 | 4 |
|  | Best | 5.5259E+03 | 8.2062E+04 | 3.5880E+03 | 1.3808E+03 | 2.9575E+04 | 2.2953E+04 | 1.4367E+07 | 1.2866E+04 | 7.3833E+03 | 5.3692E+03 |
| F12 | Ave | 1.2479E+04 | 2.4270E+05 | 1.0897E+04 | **5.6761E+03** | 6.7025E+04 | 4.8518E+04 | 2.9730E+07 | 2.4479E+04 | 2.7572E+04 | 9.4301E+03 |
|  | Std | 5.1705E+03 | 6.1163E+05 | 6.0279E+03 | 4.8001E+03 | 2.4689E+04 | 1.8556E+04 | 1.0636E+07 | 7.7211E+03 | 1.2614E+04 | 2.9029E+03 |
|  | Rank | 4 | 9 | 3 | 1 | 8 | 7 | 10 | 5 | 6 | 2 |
|  | Best | 1.5338E+03 | 6.2363E+05 | 2.5074E+05 | 1.0225E+05 | 3.5795E+05 | 2.7268E+05 | 1.6410E+06 | 9.4632E+04 | 1.0052E+05 | 2.5477E+04 |
| F13 | Ave | **1.9220E+03** | 2.1625E+06 | 7.1030E+05 | 2.9259E+05 | 1.9331E+06 | 8.1745E+05 | 9.2958E+06 | 2.1936E+05 | 7.2106E+05 | 2.6420E+04 |
|  | Std | 3.5965E+02 | 7.8840E+05 | 2.8237E+05 | 1.1710E+05 | 7.4967E+05 | 3.8794E+05 | 3.6416E+06 | 7.2746E+04 | 4.1759E+05 | 2.3887E+04 |
|  | Rank | 1 | 9 | 5 | 4 | 8 | 7 | 10 | 3 | 6 | 2 |
|  | Best | 1.9880E+03 | 3.6487E+04 | 1.9895E+03 | 1.6464E+03 | 1.6558E+04 | 4.2719E+03 | 1.2097E+06 | 2.8531E+03 | 3.7974E+03 | 2.2078E+03 |
| F14 | Ave | 3.3950E+03 | 1.1517E+05 | 4.3453E+03 | **3.2745E+03** | 6.2763E+04 | 1.1536E+04 | 5.5564E+06 | 5.2553E+03 | 1.2139E+04 | 4.3973E+03 |
|  | Std | 1.8168E+03 | 1.8768E+05 | 3.2892E+03 | 2.8920E+03 | 3.7122E+04 | 7.3466E+03 | 2.4877E+06 | 2.2063E+03 | 6.7516E+03 | 1.9096E+03 |
|  | Rank | 2 | 9 | 3 | 1 | 8 | 6 | 10 | 5 | 7 | 4 |
|  | Best | 2.0696E+03 | 4.5818E+03 | 3.1029E+03 | 3.2545E+03 | 3.6599E+03 | 4.1907E+03 | 5.5431E+03 | 4.1001E+03 | 3.4697E+03 | 7.2259E+03 |
| F15 | Ave | **2.5702E+03** | 5.9877E+03 | 5.0319E+03 | 4.2895E+03 | 5.3684E+03 | 5.5044E+03 | 6.8867E+03 | 5.2453E+03 | 5.2465E+03 | 8.4937E+03 |
|  | Std | 3.4060E+02 | 6.3825E+02 | 7.4675E+02 | 6.6255E+02 | 6.7107E+02 | 6.8832E+02 | 6.4090E+02 | 6.4685E+02 | 6.0359E+02 | 4.3106E+02 |
|  | Rank | 1 | 8 | 3 | 2 | 6 | 7 | 9 | 4 | 5 | 10 |
|  | Best | 2.1374E+03 | 3.8195E+03 | 2.6702E+03 | 2.6250E+03 | 3.8567E+03 | 3.8584E+03 | 4.4540E+03 | 3.4631E+03 | 3.4582E+03 | 5.3235E+03 |
| F16 | Ave | **3.4943E+03** | 4.9563E+03 | 4.5517E+03 | 3.7746E+03 | 4.6545E+03 | 4.6915E+03 | 5.4163E+03 | 4.2128E+03 | 4.7626E+03 | 5.9778E+03 |
|  | Std | 5.9444E+02 | 5.5121E+02 | 5.9857E+02 | 5.6302E+02 | 4.6256E+02 | 4.2821E+02 | 4.5976E+02 | 3.5929E+02 | 5.7471E+02 | 2.8975E+02 |
|  | Rank | 1 | 8 | 4 | 2 | 5 | 6 | 9 | 3 | 7 | 10 |
|  | Best | 4.7480E+03 | 9.2356E+05 | 3.0552E+05 | 4.3724E+05 | 6.1482E+05 | 4.8779E+05 | 4.1255E+06 | 1.4285E+05 | 3.8152E+05 | 3.6150E+04 |
| F17 | Ave | **1.1557E+04** | 3.4209E+06 | 1.2402E+06 | 1.4238E+06 | 2.6747E+06 | 1.8894E+06 | 1.6725E+07 | 4.7848E+05 | 1.1192E+06 | 1.4332E+05 |
|  | Std | 5.7958E+03 | 1.5649E+06 | 5.2924E+05 | 5.3768E+05 | 1.1552E+06 | 8.2723E+05 | 8.0093E+06 | 1.7640E+05 | 5.3467E+05 | 6.3608E+04 |
|  | Rank | 1 | 9 | 5 | 6 | 8 | 7 | 10 | 3 | 4 | 2 |
| F18 | Best | 2.1154E+03 | 3.0134E+05 | 2.1315E+03 | 1.9514E+03 | 9.6695E+04 | 2.9843E+03 | 1.9392E+06 | 2.1900E+03 | 3.7688E+03 | 2.1292E+03 |
|     | Ave  | **3.9018E+03** | 2.8190E+06 | 5.1714E+03 | 5.6485E+03 | 1.5310E+06 | 1.1340E+04 | 6.2171E+06 | 4.3222E+03 | 1.7076E+04 | 4.4911E+03 |
|     | Std  | 2.5073E+03 | 1.9366E+06 | 4.1557E+03 | 5.3690E+03 | 1.1155E+06 | 7.5894E+03 | 3.3453E+06 | 2.4473E+03 | 1.1364E+04 | 2.2594E+03 |
|     | Rank | 1 | 9 | 4 | 5 | 8 | 6 | 10 | 2 | 7 | 3 |
|     | Best | 2.6131E+03 | 4.0096E+03 | 2.9659E+03 | 2.5618E+03 | 3.7128E+03 | 3.4548E+03 | 3.9075E+03 | 3.6420E+03 | 3.7684E+03 | 6.3552E+03 |
|     | Ave  | 4.0847E+03 | 4.9831E+03 | 4.5535E+03 | **3.8647E+03** | 4.6799E+03 | 4.6415E+03 | 5.4581E+03 | 4.6517E+03 | 4.7689E+03 | 6.7004E+03 |
| F19 | Std  | 4.7790E+02 | 4.8187E+02 | 6.1730E+02 | 5.6559E+02 | 4.4519E+02 | 4.3521E+02 | 5.0285E+02 | 4.7938E+02 | 5.3639E+02 | 1.8920E+02 |
|     | Rank | 2 | 8 | 3 | 1 | 6 | 4 | 9 | 5 | 7 | 10 |
|     | Best | 2.4121E+03 | 2.5830E+03 | 2.5172E+03 | 2.4359E+03 | 2.5424E+03 | 2.5787E+03 | 3.0729E+03 | 2.5449E+03 | 2.6143E+03 | 2.9542E+03 |
|     | Ave  | **2.4755E+03** | 2.7248E+03 | 2.6079E+03 | 2.5168E+03 | 2.6451E+03 | 2.6982E+03 | 3.2488E+03 | 2.6587E+03 | 2.8061E+03 | 3.1415E+03 |
| F20 | Std  | 2.7429E+01 | 7.8252E+01 | 5.2007E+01 | 3.9044E+01 | 5.8273E+01 | 5.8121E+01 | 7.9008E+01 | 5.5475E+01 | 7.1232E+01 | 4.4147E+01 |
|     | Rank | 1 | 7 | 3 | 2 | 4 | 6 | 10 | 5 | 8 | 9 |
|     | Best | 1.2034E+04 | 1.5303E+04 | 1.5434E+04 | 2.3000E+03 | 2.3155E+03 | 1.4730E+04 | 2.1981E+04 | 2.1464E+04 | 2.3001E+03 | 3.0737E+04 |
|     | Ave  | 1.5299E+04 | 1.8111E+04 | 1.8216E+04 | **1.5124E+04** | 1.8942E+04 | 1.7957E+04 | 2.5320E+04 | 2.5641E+04 | 1.5253E+04 | 3.2036E+04 |
| F21 | Std  | 1.5212E+03 | 1.6640E+03 | 2.0364E+03 | 7.3635E+03 | 2.8129E+03 | 1.3627E+03 | 1.1600E+03 | 2.0020E+03 | 3.5586E+03 | 4.6569E+02 |
|     | Rank | 3 | 5 | 6 | 1 | 7 | 4 | 8 | 9 | 2 | 10 |
|     | Best | 2.9224E+03 | 3.0837E+03 | 3.0292E+03 | 2.9378E+03 | 3.0051E+03 | 3.0443E+03 | 3.5069E+03 | 3.0139E+03 | 3.1000E+03 | 3.2600E+03 |
|     | Ave  | 3.0132E+03 | 3.2167E+03 | 3.1212E+03 | **3.0103E+03** | 3.1264E+03 | 3.1229E+03 | 3.6375E+03 | 3.1209E+03 | 3.2220E+03 | 3.6209E+03 |
| F22 | Std  | 3.6272E+01 | 6.2286E+01 | 3.8739E+01 | 3.2775E+01 | 4.3961E+01 | 3.9037E+01 | 5.9212E+01 | 5.1564E+01 | 5.9221E+01 | 1.1201E+02 |
|     | Rank | 2 | 7 | 4 | 1 | 6 | 5 | 10 | 3 | 8 | 9 |
|     | Best | 3.3671E+03 | 3.5475E+03 | 3.4516E+03 | 3.3790E+03 | 3.4915E+03 | 3.5218E+03 | 4.1910E+03 | 3.4749E+03 | 3.5864E+03 | 3.6513E+03 |
|     | Ave  | **3.4317E+03** | 3.7134E+03 | 3.5567E+03 | 3.4446E+03 | 3.6034E+03 | 3.6348E+03 | 4.3551E+03 | 3.5692E+03 | 3.8010E+03 | 4.0444E+03 |
| F23 | Std  | 2.4886E+01 | 8.2009E+01 | 6.0678E+01 | 3.2850E+01 | 5.8658E+01 | 5.7338E+01 | 8.4984E+01 | 5.0290E+01 | 9.7413E+01 | 1.3235E+02 |
|     | Rank | 1 | 7 | 3 | 2 | 5 | 6 | 10 | 4 | 8 | 9 |
|     | Best | 3.2290E+03 | 3.3071E+03 | 3.2430E+03 | 3.1915E+03 | 3.3861E+03 | 3.3533E+03 | 4.0789E+03 | 3.3637E+03 | 3.1105E+03 | 3.4165E+03 |
|     | Ave  | 3.3352E+03 | 3.5106E+03 | 3.3491E+03 | 3.3174E+03 | 3.6205E+03 | 3.4671E+03 | 4.7977E+03 | 3.5495E+03 | **3.2500E+03** | 3.5507E+03 |
| F24 | Std  | 5.4825E+01 | 5.3679E+01 | 4.6956E+01 | 4.7613E+01 | 7.2449E+01 | 5.2661E+01 | 4.1976E+02 | 5.5550E+01 | 4.5972E+01 | 5.8204E+01 |
|     | Rank | 3 | 6 | 4 | 2 | 9 | 5 | 10 | 7 | 1 | 8 |
|     | Best | 6.9767E+03 | 9.3310E+03 | 7.3984E+03 | 6.6454E+03 | 4.6155E+03 | 8.8533E+03 | 1.3936E+04 | 7.4924E+03 | 2.9001E+03 | 8.9738E+03 |
|     | Ave  | 7.9618E+03 | 1.0461E+04 | 9.0379E+03 | **7.4174E+03** | 9.2698E+03 | 9.8533E+03 | 1.6184E+04 | 8.7198E+03 | 9.9217E+03 | 1.3190E+04 |
| F25 | Std  | 4.3001E+02 | 6.2345E+02 | 6.6665E+02 | 3.7226E+02 | 9.2789E+02 | 4.8946E+02 | 1.0022E+03 | 5.5764E+02 | 3.4851E+03 | 1.5826E+03 |
|     | Rank | 2 | 8 | 4 | 1 | 5 | 6 | 10 | 3 | 7 | 9 |
|     | Best | 3.3562E+03 | 3.4574E+03 | 3.3797E+03 | 3.3164E+03 | 3.4252E+03 | 3.3745E+03 | 3.6919E+03 | 3.4533E+03 | 3.2000E+03 | 3.4785E+03 |
|     | Ave  | 3.4714E+03 | 3.5979E+03 | 3.4303E+03 | 3.3977E+03 | 3.5824E+03 | 3.4751E+03 | 3.9032E+03 | 3.5836E+03 | **3.2529E+03** | 3.5767E+03 |
| F26 | Std  | 5.7276E+01 | 6.9928E+01 | 3.3914E+01 | 3.7262E+01 | 7.1754E+01 | 4.6553E+01 | 8.9889E+01 | 8.5434E+01 | 8.1202E+01 | 5.6012E+01 |
|     | Rank | 4 | 9 | 3 | 2 | 7 | 5 | 10 | 8 | 1 | 6 |
|     | Best | 3.3589E+03 | 3.5279E+03 | 3.3762E+03 | 3.3337E+03 | 3.5194E+03 | 3.4652E+03 | 4.5813E+03 | 3.5317E+03 | 3.3000E+03 | 3.5814E+03 |
|     | Ave  | 3.4295E+03 | 3.5940E+03 | 3.4659E+03 | 3.4233E+03 | 3.6836E+03 | 3.5569E+03 | 6.4812E+03 | 3.6248E+03 | **3.3080E+03** | 3.7140E+03 |
| F27 | Std  | 4.3281E+01 | 3.6144E+01 | 3.9942E+01 | 3.4255E+01 | 7.7338E+01 | 4.5902E+01 | 1.0747E+03 | 4.8338E+01 | 1.9659E+01 | 7.7918E+01 |
|     | Rank | 3 | 6 | 4 | 2 | 8 | 5 | 10 | 7 | 1 | 9 |
|     | Best | 3.9486E+03 | 5.8345E+03 | 4.7697E+03 | 4.2830E+03 | 5.7253E+03 | 4.9291E+03 | 6.1093E+03 | 5.1434E+03 | 5.1240E+03 | 7.1518E+03 |
|     | Ave  | **4.6490E+03** | 7.1215E+03 | 5.7926E+03 | 5.1669E+03 | 6.9448E+03 | 6.1174E+03 | 6.9854E+03 | 6.2903E+03 | 6.3373E+03 | 8.4413E+03 |
| F28 | Std  | 2.9594E+02 | 4.9539E+02 | 5.4117E+02 | 4.0607E+02 | 5.1350E+02 | 4.9434E+02 | 5.2098E+02 | 4.5008E+02 | 5.1020E+02 | 5.0217E+02 |
|     | Rank | 1 | 9 | 3 | 2 | 7 | 4 | 8 | 5 | 6 | 10 |

| | | | | | | | | | | |
|---|---|---|---|---|---|---|---|---|---|---|
| F29 | Best | 6.9610E+03 | 7.3265E+06 | 7.7427E+03 | 5.9660E+03 | 7.7490E+06 | 6.2983E+04 | 8.4612E+06 | 4.3849E+04 | 5.6331E+04 | 5.9596E+04 |
| | Ave | 1.1888E+04 | 3.7941E+07 | 2.3307E+04 | **9.3292E+03** | 3.3914E+07 | 2.3365E+05 | 2.1113E+07 | 1.2129E+05 | 3.6752E+05 | 1.3083E+05 |
| | Std | 5.5962E+03 | 1.8545E+07 | 1.6972E+04 | 3.1614E+03 | 1.7823E+07 | 1.3242E+05 | 8.0128E+06 | 4.9844E+04 | 3.1151E+05 | 5.4985E+04 |
| | Rank | 2 | 10 | 3 | 1 | 9 | 6 | 8 | 4 | 7 | 5 |

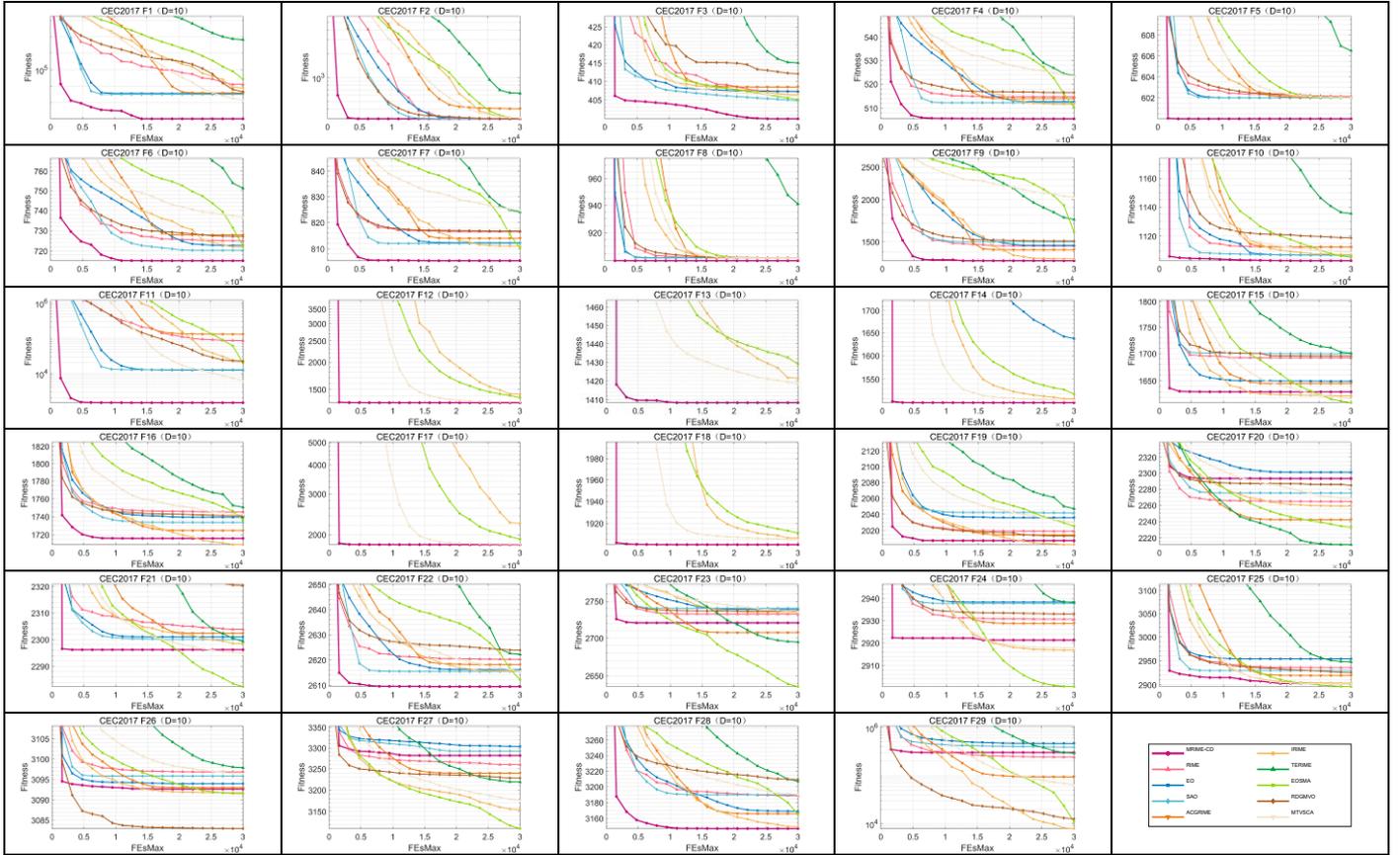

**Figure A1.** Convergence curves of MRIME-CD and other competitors based on CEC2017 (D=10)

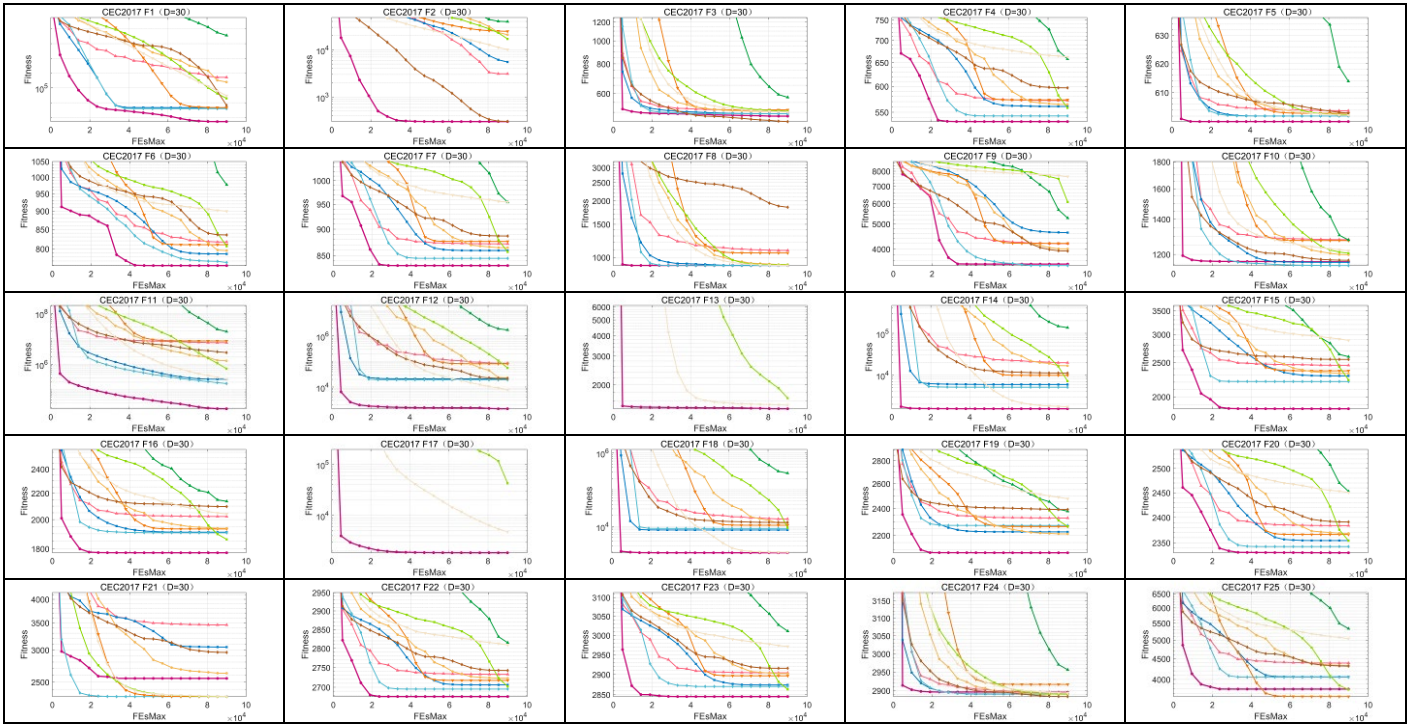

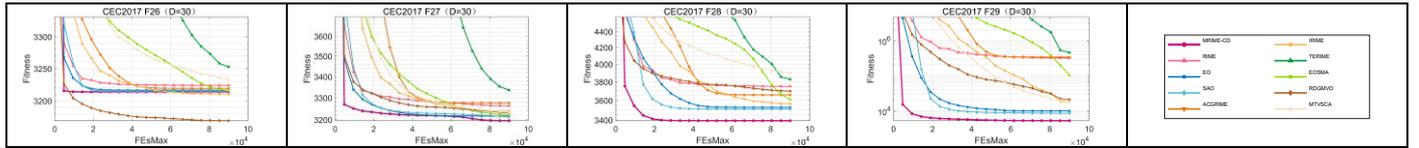
**Figure A2.** Convergence curves of MRIME-CD and other competitors based on CEC2017 (D=30)

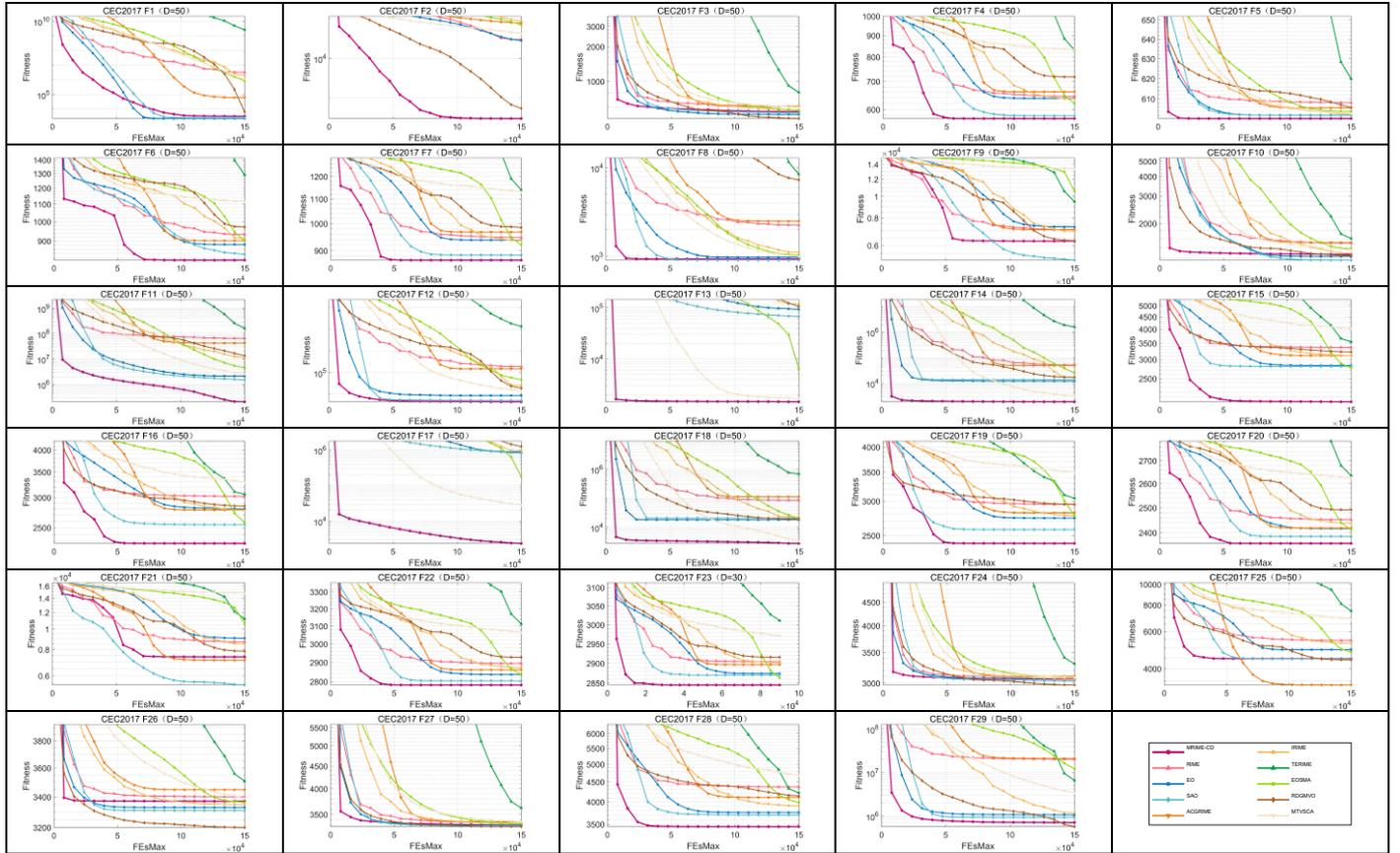
**Figure A3.** Convergence curves of MRIME-CD and other competitors based on CEC2017 (D=50)

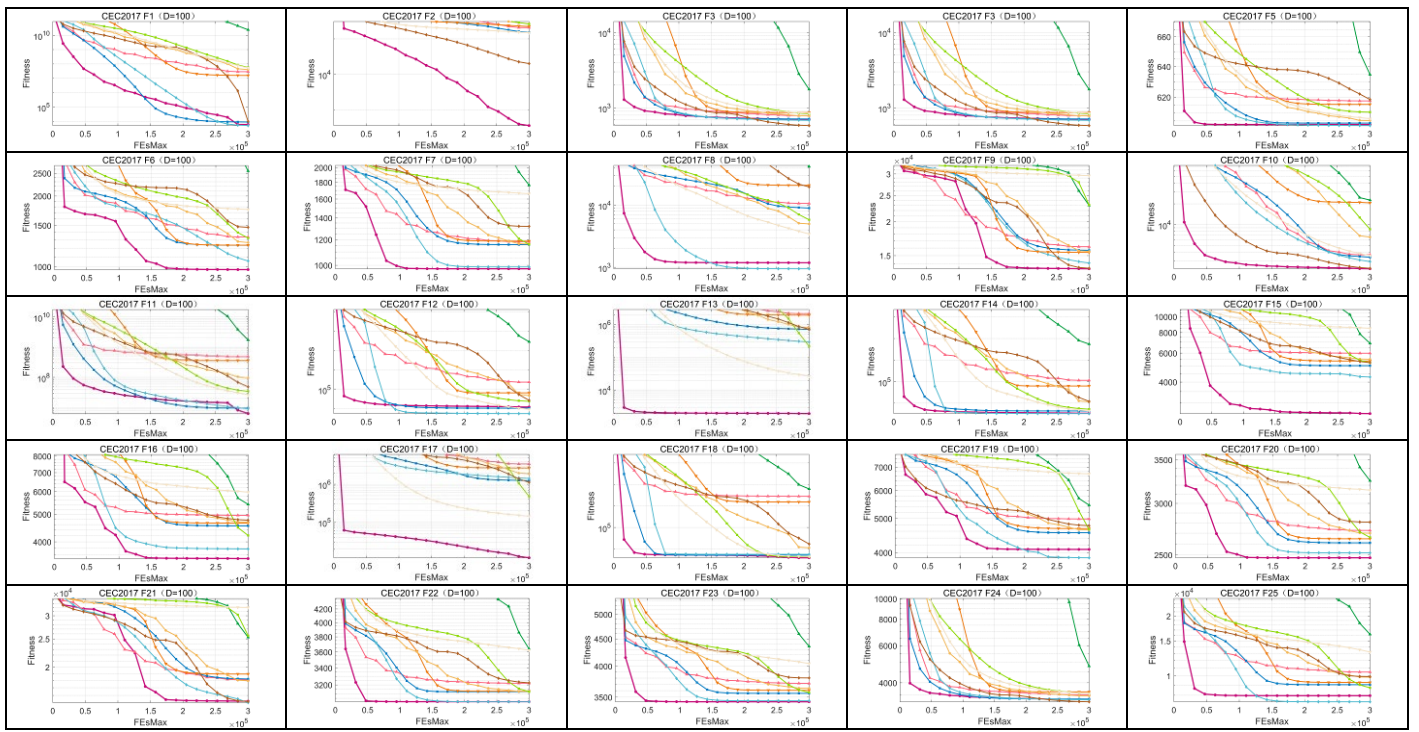

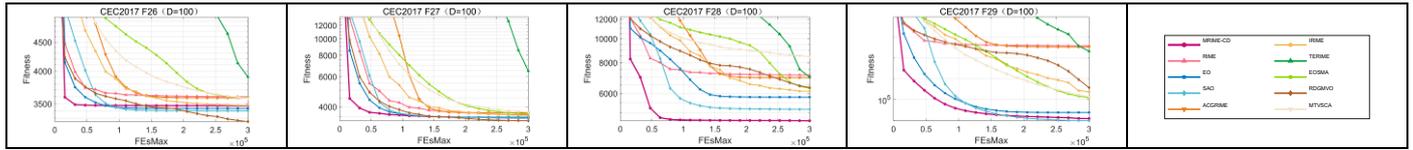

**Figure A4.** Convergence curves of MRIME-CD and other competitors based on CEC2017 (D=100)

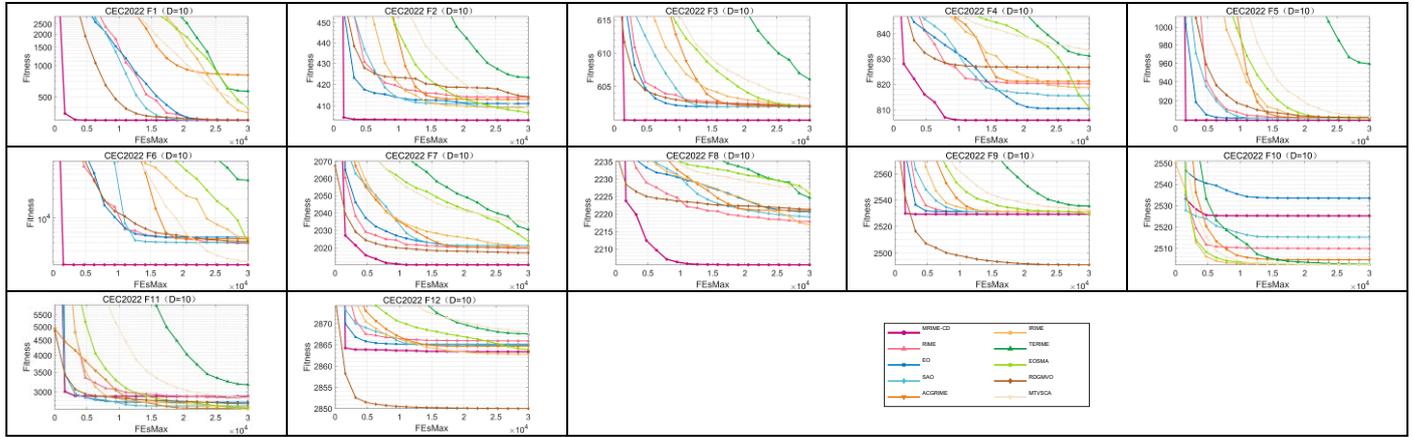

**Figure A5.** Convergence curves of MRIME-CD and other competitors based on CEC2022 (D=10)

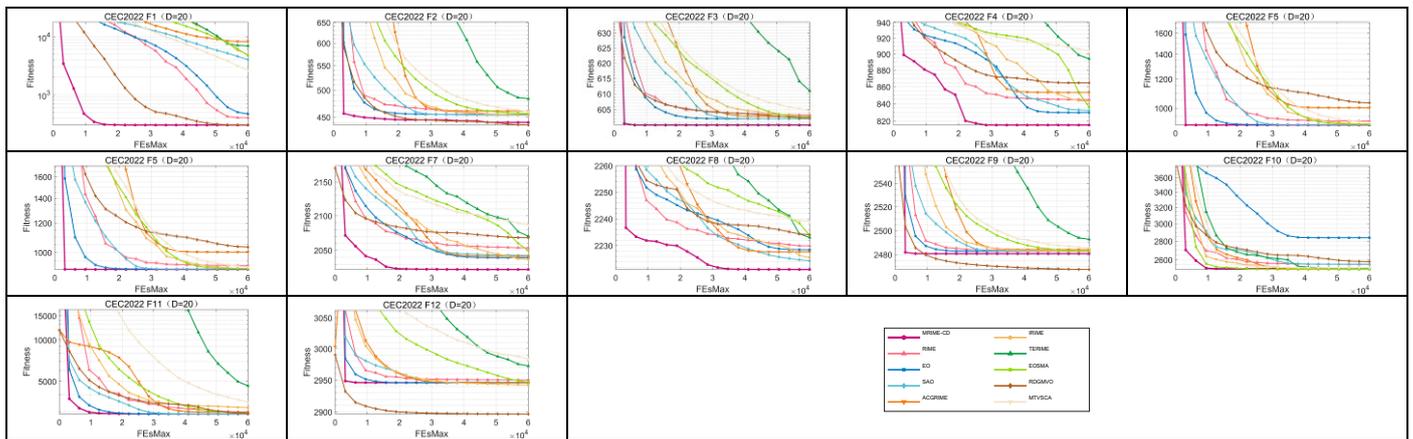

**Figure A6.** Convergence curves of MRIME-CD and other competitors based on CEC2022 (D=20)